\documentclass[10pt,twocolumn,letterpaper]{article}

\usepackage{cvpr}              
\usepackage{multirow}      
\usepackage[table]{xcolor}
\usepackage[most]{tcolorbox}
\usepackage{tikz}
\usepackage{fancyvrb}
\usepackage{booktabs}
\usepackage{fancyvrb}
\usepackage{listings}
\usepackage{threeparttable}
\usepackage{stfloats}

\definecolor{cvprblue}{rgb}{0.21,0.49,0.74}
\definecolor{cityblue}{RGB}{128, 159, 225}
\definecolor{citypink}{RGB}{227, 108, 194}
\usepackage[pagebackref,breaklinks,colorlinks,allcolors=cvprblue]{hyperref}
\newcommand{\imageref}{%
  \raisebox{-0.2em}{\includegraphics[width=0.022\linewidth]{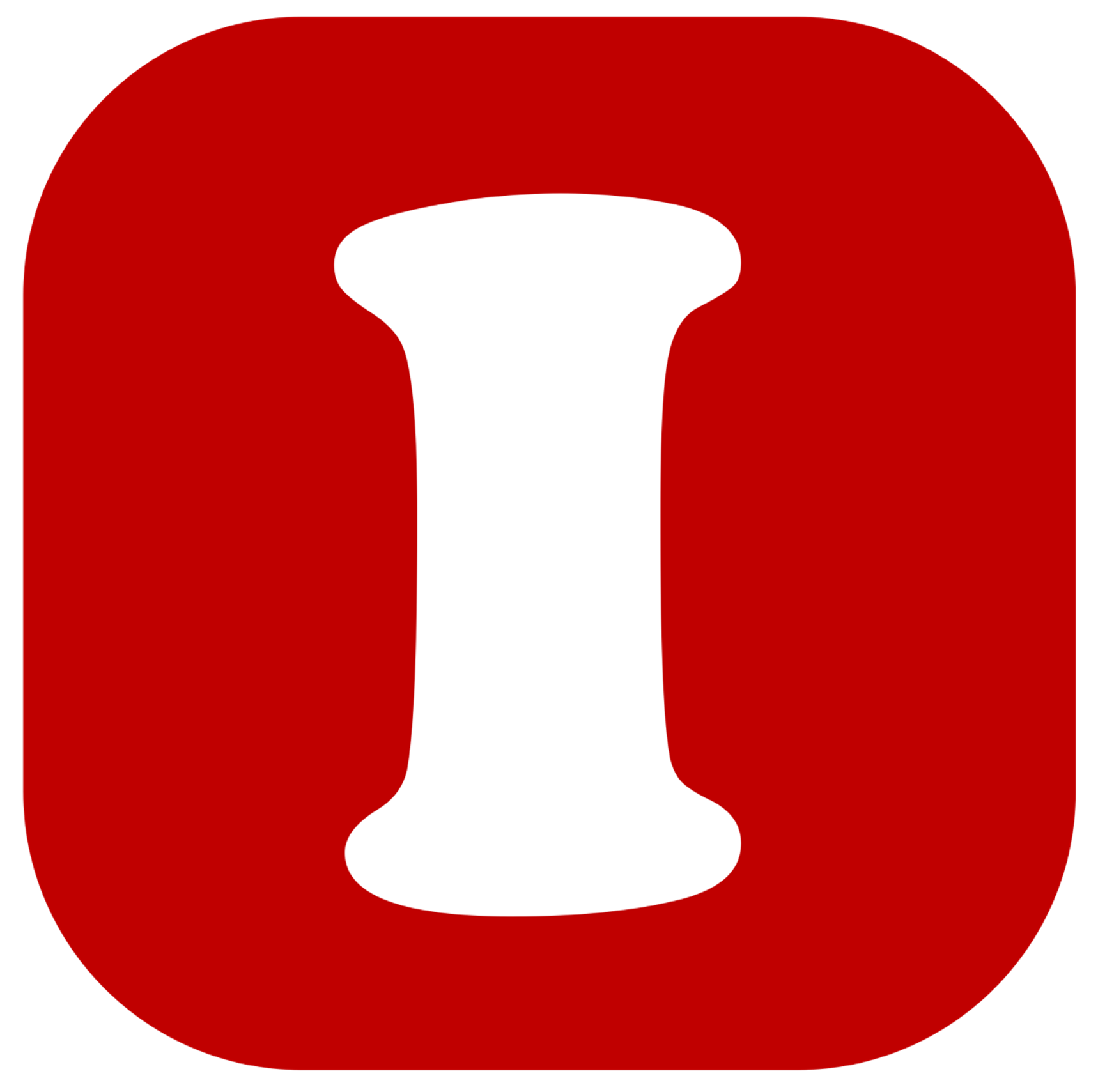}}%
}
\newcommand{\textref}{%
  \raisebox{-0.2em}{\includegraphics[width=0.022\linewidth]{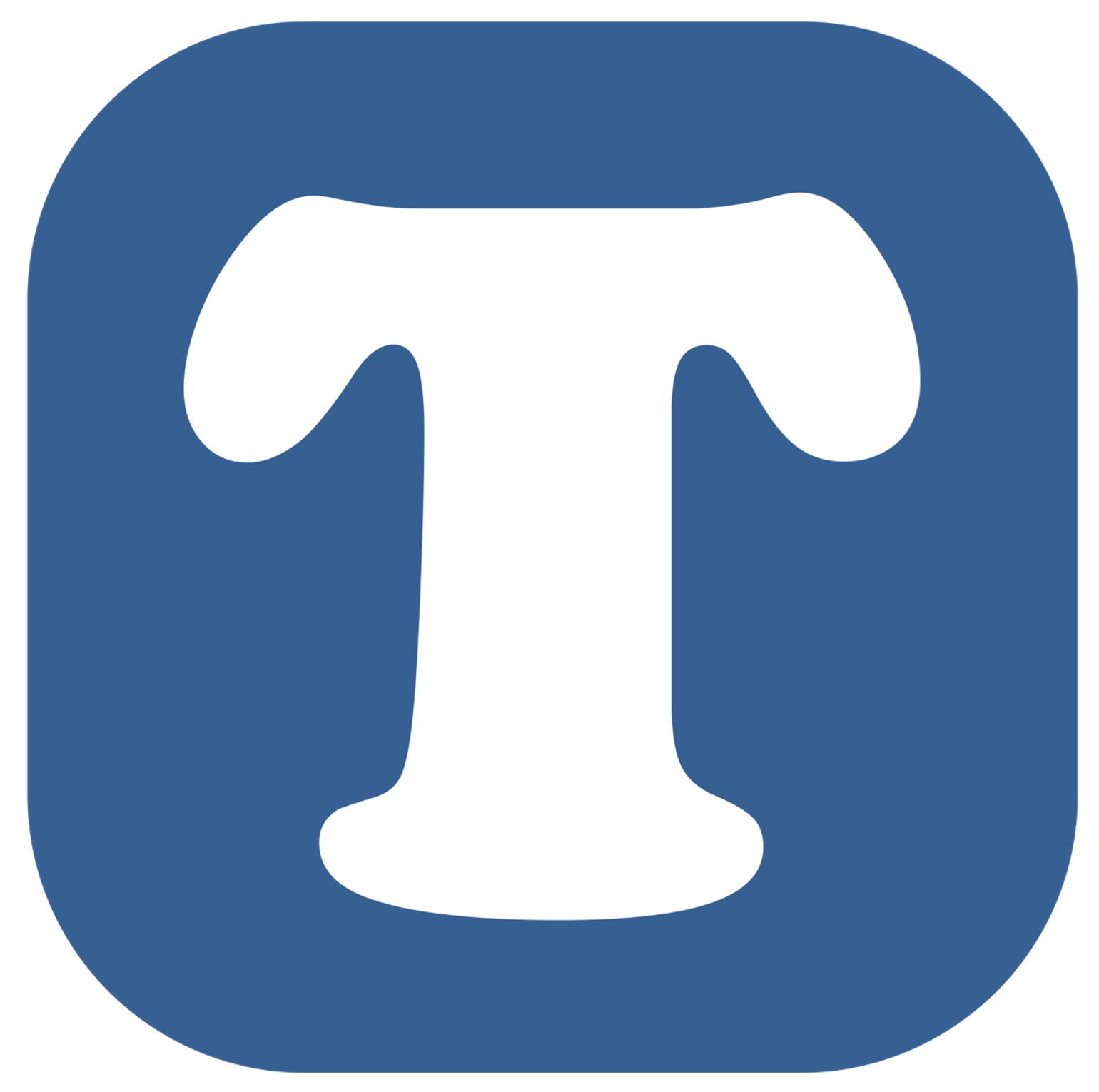}}%
}
\newcommand{\autoregressive}{%
  \raisebox{-0.2em}{\includegraphics[width=0.022\linewidth]{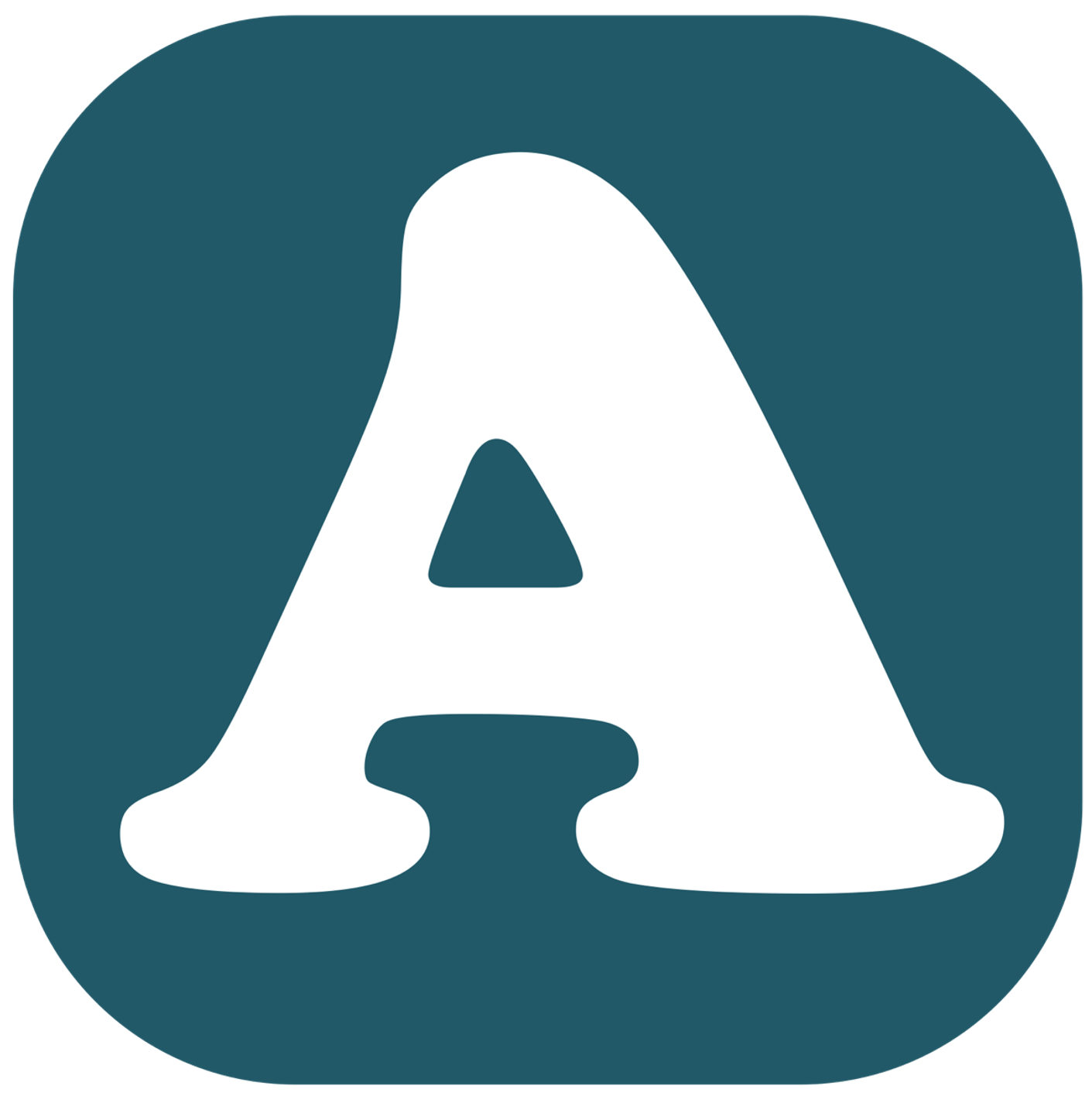}}%
}
\newcommand{\tablespacing}{
    \multicolumn{15}{c}{} \vspace{-0.3cm}
}
\newcommand{\myfont}{\fontsize{11pt}{13pt}\selectfont}

\def\paperID{} 
\def\confName{CVPR}
\def\confYear{2026}

\title{ViStoryBench: Comprehensive Benchmark Suite for Story Visualization}

\author{
	Cailin Zhuang$^{1,2,3,*}$\quad
    Ailin Huang$^{2,*,\dagger}$\quad
    Yaoqi Hu$^{3,*}$\quad
    Jingwei Wu$^{2}$\quad
    Wei Cheng$^{2,\dagger}$
    \\
    Jiaqi Liao$^{2,4}$\quad
    Hongyuan Wang$^{2}$\quad
    Xinyao Liao$^{2}$\quad
    Weiwei Cai$^{2}$\quad
    Hengyuan Xu$^{2}$
     \\
    Xuanyang Zhang$^{2}$\quad
    Xianfang Zeng$^{2}$\quad
    Zhewei Huang$^{2,\ddagger}$\quad
    Gang Yu$^{2,\ddagger}$\quad
    Chi Zhang$^{4,\ddagger}$
	\\
    [0.4em]\normalsize
    $^{1}$~ShanghaiTech University\quad
    $^{2}$~StepFun\quad
    $^{3}$~AIGC Research\quad
    $^{4}$~AGI Lab, Westlake University \\    
    [0.4em]
    \textcolor{citypink}{
    \raisebox{-0.2\height}{\includegraphics[height=0.5cm]{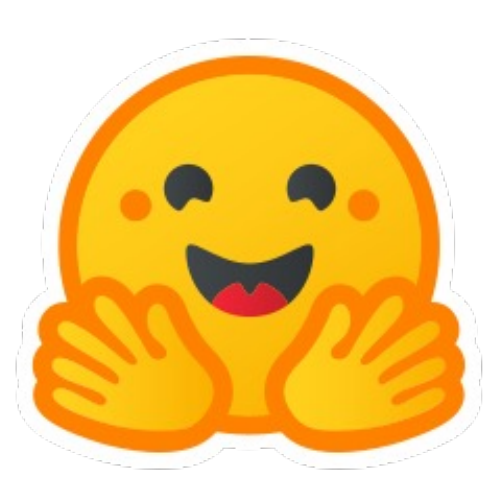}}~{\href{https://huggingface.co/datasets/ViStoryBench/ViStoryBench}{\textbf{Dataset}}}
    \quad
    \raisebox{-0.2\height}{\includegraphics[height=0.5cm]{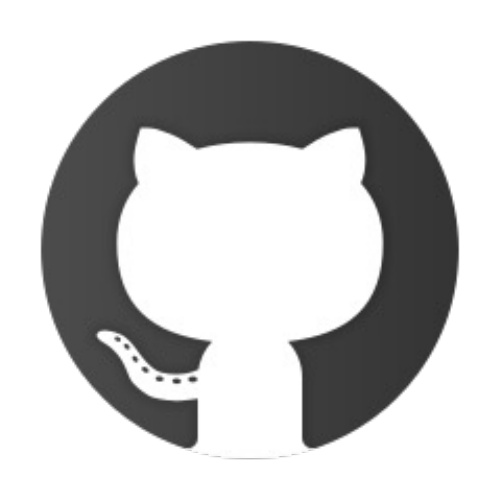}}~{\href{https://github.com/vistorybench/vistorybench}{\textbf{Code}}}
    \quad
    \raisebox{-0.2\height}{\includegraphics[height=0.5cm]{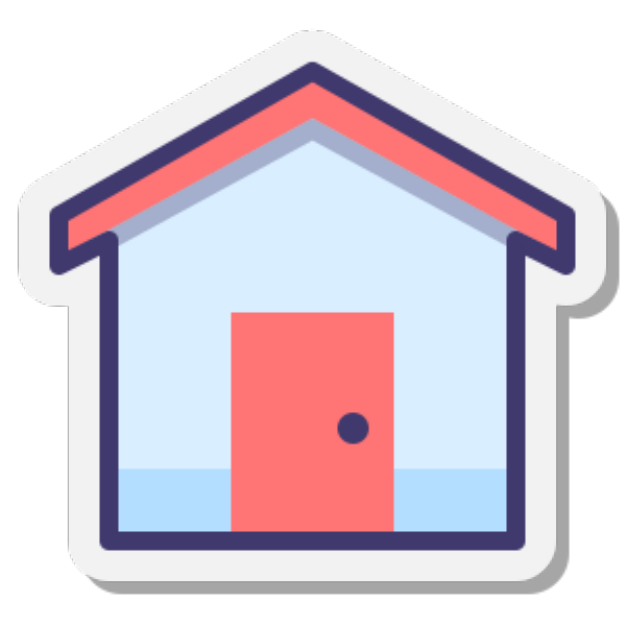}}~{\href{https://vistorybench.github.io}{\textbf{Project Page}}}
    \quad
    \raisebox{-0.2\height}{\includegraphics[height=0.5cm]{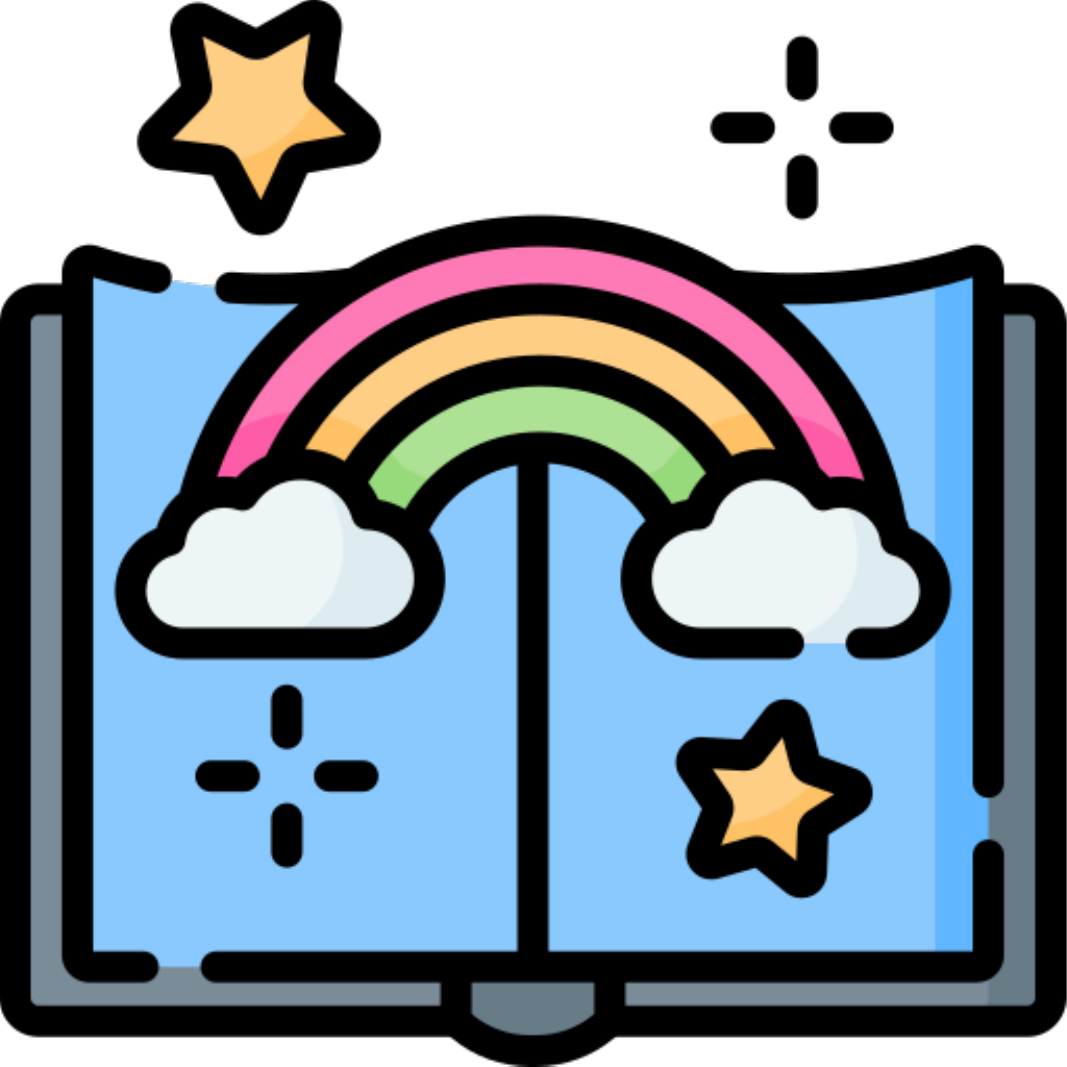}}~{\href{https://vistorybench.github.io/story_detail}{\textbf{Story Explorer}}}}
    }

\definecolor{mygray}{gray}{.9}
\definecolor{best}{rgb}{1, 0.7, 0.8}
\definecolor{best2}{rgb}{1, 0.8, 0.9}

\definecolor{colorfirst}{RGB}{252,151,100}
\definecolor{colorsecond}{RGB}{253,187,132}
\definecolor{colorthird}{RGB}{253,212,158}
\definecolor{colorfourth}{RGB}{254,232,200}
\definecolor{colorfifth}{RGB}{255,247,236}
\newcommand{\rankfirst}[1]{\cellcolor{colorfirst}#1}
\newcommand{\ranksecond}[0]{\cellcolor{colorsecond}}
\newcommand{\rankthird}[0]{\cellcolor{colorthird}}
\newcommand{\rankfourth}[0]{\cellcolor{colorfourth}}
\newcommand{\rankfifth}[0]{\cellcolor{colorfifth}}

\DeclareRobustCommand{\legendsquare}[1]{%
  \textcolor{#1}{\rule{2ex}{2ex}}%
}
\DeclareRobustCommand{\legendsquarebox}[1]{%
  \tikz[] \draw[black, fill=#1, line width=0.4pt] (0,0) rectangle (1.5ex,1.5ex);%
}
\newtcbox{\highlighttext}[3]{%
  arc=4pt, %
  colback=#2!80, %
  boxrule=0pt, %
  left=0.3pt, right=0pt, top=-1.5pt, bottom=-1.5pt,
  nobeforeafter,
  tcbox raise base,
  enhanced,
  fontupper=\textcolor{#1}{#3}%
}

\definecolor{teaserdatasetback}{RGB}{240,248,253}
\definecolor{teaserdatasetback_shallow}{RGB}{240,248,253}
\definecolor{teaserdatasettext}{RGB}{220,168,0}
\definecolor{benchback}{RGB}{255,251,219}
\definecolor{benchtext}{RGB}{192,79,21}

\definecolor{datasetback}{RGB}{254,232,207}
\definecolor{datasettext}{RGB}{191,144,1}
\definecolor{castback}{RGB}{255,255,229}
\definecolor{casttext}{RGB}{65,171,93}
\definecolor{statsback}{RGB}{255,245,240}
\definecolor{statstext}{RGB}{239,101,78}

\definecolor{PAback}{RGB}{240,248,253}
\definecolor{PAtext}{RGB}{220,168,0}
\definecolor{CPback}{RGB}{255,255,217}
\definecolor{CPtext}{RGB}{220,168,0}
\definecolor{OCCMback}{RGB}{254,232,207}
\definecolor{OCCMtext}{RGB}{192,79,21}
\definecolor{SIMback}{RGB}{253,248,247}
\definecolor{SIMtext}{RGB}{192,79,21}

\lstdefinelanguage{json}{
    basicstyle=\ttfamily,
    showstringspaces=false,
    commentstyle=\color{gray},
    stringstyle=\color{purple},
    keywordstyle=\color{blue},
    morestring=[b]",
    morecomment=[l]{//},
}

\begin{document}

\twocolumn[{
  \renewcommand\twocolumn[1][]{#1}
  \maketitle
  \begin{center}
  \vspace{-6ex}
  \includegraphics[width=0.95\textwidth]{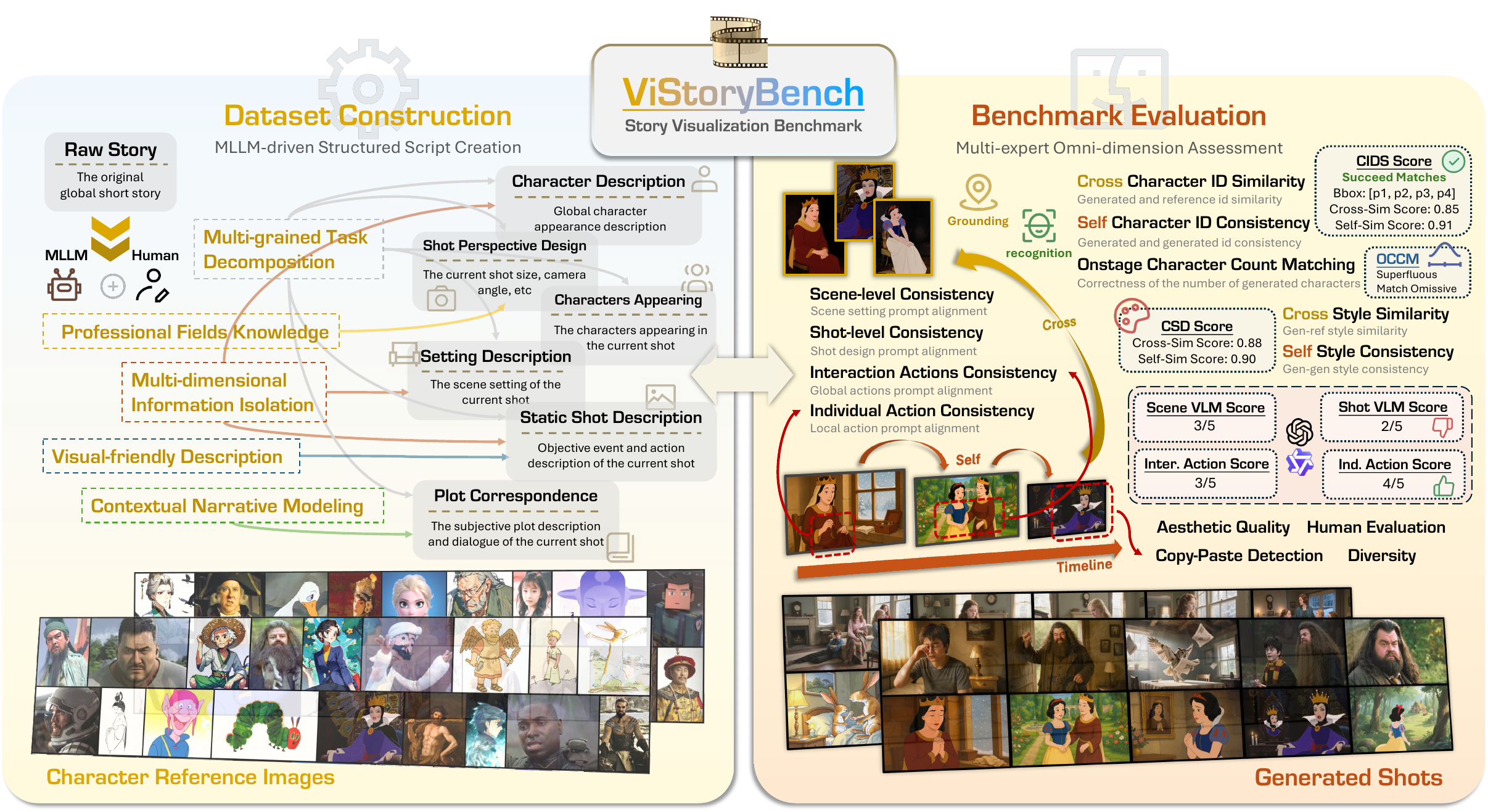}
  \vspace{-1ex}
    \captionof{figure}{\small \textbf{Overview of ViStoryBench}.  
    \highlighttext{teaserdatasettext}{teaserdatasetback}{\small \textbf{Dataset Construction}}~: We employ structured prompt engineering with 5 strategies to convert an LLM into a controllable visual narrative script generator, with human proofreading for reasonableness and manual collection of character reference images.  
    \highlighttext{benchtext}{benchback}{\small \textbf{Benchmark Evaluation}}~: Based on this dataset, we develop metrics for character/style similarity and multi-grained prompt alignment, evaluated through a hybrid framework combining expert models and VLMs.}
  \label{fig:teaser}
  \end{center}
}]

\let\thefootnote\relax\footnotetext{* Equal contribution. \dag~Project leads. \ddag~Corresponding authors. }


\begin{abstract}

Story visualization aims to generate coherent image sequences that faithfully represent a narrative and match given character references. Despite progress in generative models, existing benchmarks remain narrow in scope, often limited to short prompts, lacking character references, or single-image cases, failing to reflect real-world narrative complexity and obscuring true model performance.
We introduce \textbf{ViStoryBench}, a comprehensive benchmark designed to evaluate story visualization models across varied narrative structures, visual styles, and character settings. It features richly annotated multi-shot scripts derived from curated stories spanning literature, film, and folklore. Large language models assist in story summarization and script generation, with all outputs verified by humans for coherence and fidelity. Character references are carefully curated to maintain consistency across different artistic styles.  
ViStoryBench proposes a suite of multi-dimensional automated metrics to evaluate character consistency, style similarity, prompt alignment, aesthetic quality, and artifacts like copy-paste behavior. These metrics are validated through human studies and used to assess a broad range of open-source and commercial models, enabling systematic analysis and encouraging advances in visual storytelling.

\end{abstract}

	\section{Introduction}

    In recent years, story visualization~\cite{xu2024mmstoryagent,cheng2024theatergen,zhou2024storydiffusion} has emerged as a rapidly evolving research field, aiming to generate sequences of visually consistent images that faithfully convey a narrative while adhering to character references, thereby creating an engaging storytelling experience. This capability holds broad potential in areas such as entertainment and education, especially with recent advances in generative models~\cite{xing2023survey,liu2024sora,ming2024survey,ma2025styletailor,li2025uniface,xu2025cyc3d,su2025safe,wang2025safevar} enabling open-domain visual creation.

    Compared to general image or video generation evaluation~\cite{heusel2017gans,hessel2021clipscore,wang2023exploring,hartwig2024survey,tu2025favor}, story visualization evaluation involves more dimensions, including not only visual quality and diversity, but also character consistency and narrative alignment. However, recent studies~\cite{wu2025less,xu2024mmstoryagent,wu2025automated} often adopt limited evaluation metrics without a unified standard, and their constrained test scenarios fail to reflect real-world applicability. There is thus a pressing need for a comprehensive benchmark to systematically evaluate and advance story visualization methods under in the wild.
    In this paper, we present \textbf{ViStoryBench}, a comprehensive benchmark for story visualization with the following key features:

    \textbf{Multifaceted Dataset Creation}: We construct a diverse dataset spanning multiple story genres (e.g., comedy, horror) and visual styles (e.g., anime, 3D rendering) to enable comprehensive model evaluation across narrative and aesthetic dimensions. 
    We carefully curate 80 story segments with 344 characters to balance narrative structures and visual elements. ViStoryBench includes stories with both single and multiple protagonists, testing the models' ability to maintain character consistency. Additionally, it features complex plots and intricate world-building, challenging models to generate accurate visual content.

    \textbf{Comprehensive Evaluation Metrics}: In addition to conventional image generation metrics, such as image quality, diversity, and prompt alignment, we also quantify several attributes particularly vital to story visualization. These encompass stylistic consistency across the generated sequence, alignment of character actions and interactions with textual descriptions, visual novelty (beyond mere copy-pasting of references), and correctness of character sets in each scene.     
    Our benchmark incorporates 12 automated evaluation metrics. This structured and multifaceted framework enables researchers to thoroughly identify both the strengths and weaknesses of different models, thus fostering targeted improvements.

    \begin{table*}[b]
    \centering
    \vspace{-1ex}
    \caption{\small \textbf{Comparison with Related Benchmarks}. ViStoryBench offers the most comprehensive coverage in terms of styles, evaluation metrics, and the number of methods evaluated, highlighting its suitability for diverse storytelling assessments.}
    \vspace{-1ex}
    \resizebox{0.8\textwidth}{!}{
        \begin{tabular}{lcccccc}
        \toprule
        \textbf{Benchmark} & \textbf{\#Stories} & \textbf{\#Prompts / Shots} & \textbf{\#Ref Images} & \textbf{\#Styles} & \textbf{\#Metrics} & \textbf{\#Methods Evaluated} \\ 
        \midrule
        DreamBench++~\cite{peng2024dreambench++} & - & \textbf{1,350} & 150 & 3 & 2 & 7 \\ 
        OmniContext~\cite{wu2025omnigen2} & - & 400 & \textbf{790} & 2 & 2 & 6\\ 
        VinaBench~\cite{gao2025vinabench} & \textbf{85} & 1,181 & - & 6 & \textbf{12} & 3 \\ 
        \midrule
        \cellcolor{teaserdatasetback}\textbf{ViStoryBench} & \cellcolor{teaserdatasetback}80 & \cellcolor{teaserdatasetback}1,317 & \cellcolor{teaserdatasetback}509 & \cellcolor{teaserdatasetback}\textbf{10} & \cellcolor{teaserdatasetback}\textbf{12} & \cellcolor{teaserdatasetback}\textbf{25} \\ 
        \bottomrule
    \end{tabular}
    }
    \label{tab:benchmark_comparison}
    \vspace{-2ex}
\end{table*}

    \textbf{Extensive Model Evaluation}: We evaluate over 30 methods (including 25 base methods and their variants), analyzing the alignment between automated metrics and human evaluations to provide reliable model insights.

	We will release the complete benchmark, detailed prompts for data construction, evaluation results for each model, and code to ensure reproducibility, aiming to facilitate future advancements in story visualization.

	\section{Related Work}
	
	\subsection{Story Visualization}

    Recent advances in diffusion and auto-regressive models have significantly advanced multimodal long-story generation, with growing emphasis on visual consistency and cross-modal coherence.
    
    For story image generation~\cite{storyweaver,shen2024boosting,pan2024synthesizing,li2022word,chen2022charactercentric,maharana2021integrating,DreamStory,tao2024storyimager,sun2026muse,zhang2026persistent,meng2026logistory}, methods have evolved from GAN-based methods~\cite{li2019storygan} to diffusion-based techniques. Training-free methods like StoryDiffusion~\cite{zhou2024storydiffusion} leverage consistent self-attention, while UNO~\cite{wu2025uno} and USO~\cite{wu2025uso} promote content and style customization. Recent works integrate LLMs/VLMs for character and scene planning~\cite{cheng2024theatergen,wang2025storyanchors,song2025scenedecorator,wang2025characonsist}, or use CoT for scene understanding~\cite{oliveira2025storyreasoningdatasetusingchainofthought}, with extensions to story continuation~\cite{yang2024seedstory,tttvideo}, next-shot generation~\cite{he2025cut2next}, and comic creation~\cite{cheng2024autostudio,wu2024diffsensei,chen2024manga}. Unified multimodal models~\cite{wu2025omnigen2,qwen2025qwenimage,gong2025seedream} further support diverse generation tasks.
    
    In story video generation~\cite{kang2025text2story,wang2024aesopagent,hu2024storyagent,zhang2024vast,zhao2024moviedreamer,guo2025long,huang2025step,zhang2025storymem,zhang2025bridging}, systems like MMStoryAgent~\cite{xu2024mmstoryagent}, AnimDirector~\cite{li2024anim}, MovieAgent~\cite{wu2025movieagent} and FilmMaster~\cite{2025filmaster} orchestrate multi-component pipelines. Recent trends shift toward holistic scene-level synthesis, with Captain Cinema~\cite{xiao2025captain}, ShotAdapter~\cite{kara2025shotadapter}, HoloCine~\cite{meng2025holocine}, Sora2~\cite{2025sora2} and Seedance~\cite{gao2025Seedance} generating multi-shot narratives in single passes. We focus on the image-level evaluation of these systems, by extracting video keyframes as storyboards.
    
    The field expands into 3D storytelling~\cite{huang2024story3dagent,wang2024sims} while commercial platforms~\cite{morphic2024studio,moki2024,gong2025seedream} and closed-source MLLMs~\cite{hurst2024gpt4o,gemini2.02025,comanici2025gemini2.5,google2025nanobanana} bridge research with applications. Despite progress, challenges persist in multi-image coherence, long-range dependency modeling, fine-grained control, and complex narrative alignment. ViStoryBench addresses these gaps through standardized evaluation, enabling systematic analysis and driving future advancements in visual storytelling.

	\subsection{Datasets and Benchmarks}

    We summarize story visualization datasets~\cite{huang2016visual,gupta2018imagine,li2019storygan,yang2024seedstory,ye2024openstory,Liu_2024_CVPR_Storygen,yin20252kchar10kstory} in \textbf{Appendix}~\ref{sec:dataset}, showing significant advances in scale, resolution, and diversity.
    
    Several benchmarks have been proposed for video generation and visual storytelling. StoryBench~\cite{bugliarello2023storybench} evaluates text-to-video models across multiple tasks, while VBench~\cite{zheng2025vbench} and Video-Bench~\cite{han2025videobench} focus on single-shot videos. ShotBench~\cite{liu2025shotbench} introduces camera-focused QA evaluation, and DreamSim~\cite{fu2023dreamsim} provides perceptual similarity metrics. Recent benchmarks address specialized aspects: SFD~\cite{ghermi2024short} targets long-form genre-diverse stories, StoryEval measures story completion using vision-language models, MovieBench~\cite{wu2024moviebench} supports long-video analysis, RISEBench~\cite{zhao2025envisioning} covers reasoning-informed editing, and VinaBench~\cite{gao2025vinabench} enhances visual narrative fidelity with commonsense and discourse annotations.
    
    Compared to existing benchmarks (Table~\ref{tab:benchmark_comparison}), ViStoryBench differs significantly from DreamBench++~\cite{peng2024dreambench++} and OmniContext~\cite{wu2025omnigen2}, which focus on single-image generation with limited style variation. While VinaBench also targets storytelling, it lacks character references and supports fewer styles. ViStoryBench introduces a reference-based, multi-image benchmark with broader style coverage and stronger support for character consistency for comprehensive story visualization evaluation.


\begin{figure*}[t]
  \centering
  
  \vspace{-1ex}
  \begin{subfigure}{0.42\textwidth} 
    \includegraphics[width=\linewidth]{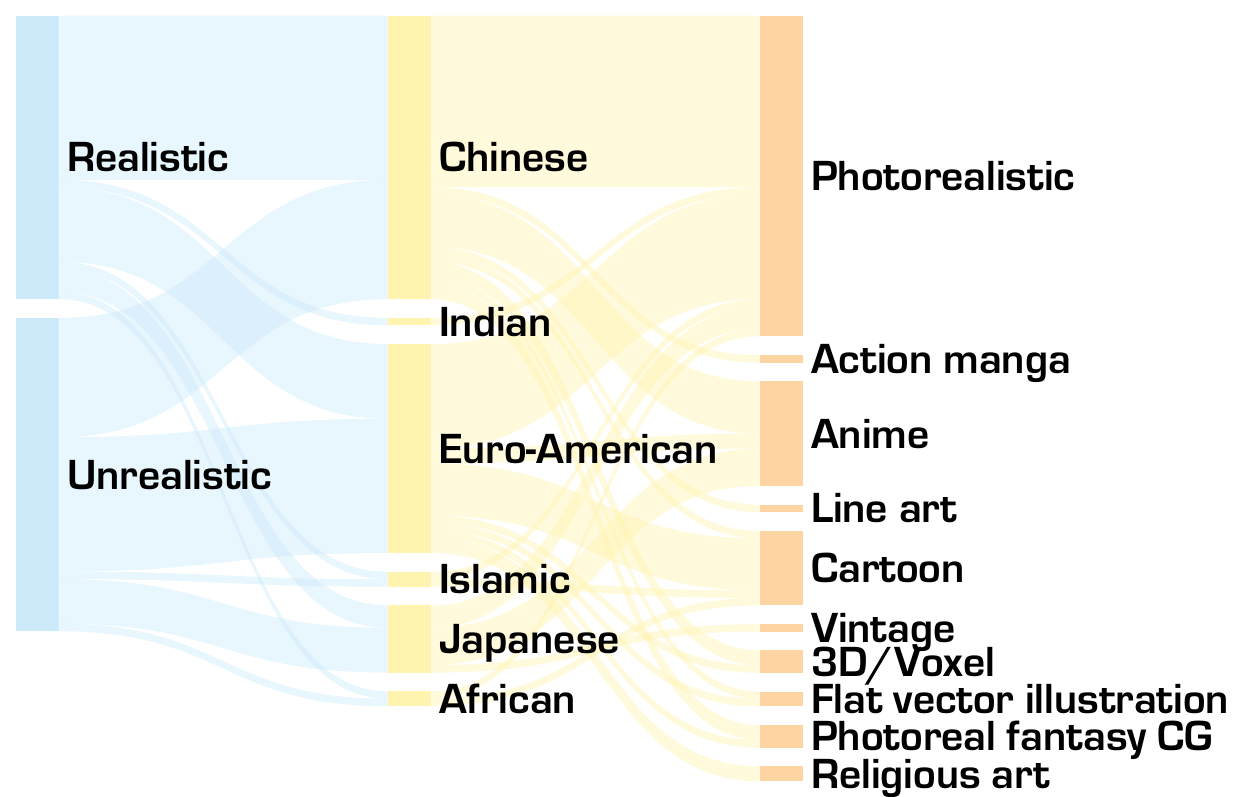}
    \caption{\textbf{Cultural and Style Distribution.}}
    \label{fig:main_sankey}
  \end{subfigure}
  \hfill 
  \begin{subfigure}{0.28\textwidth} 
    \includegraphics[width=\linewidth]{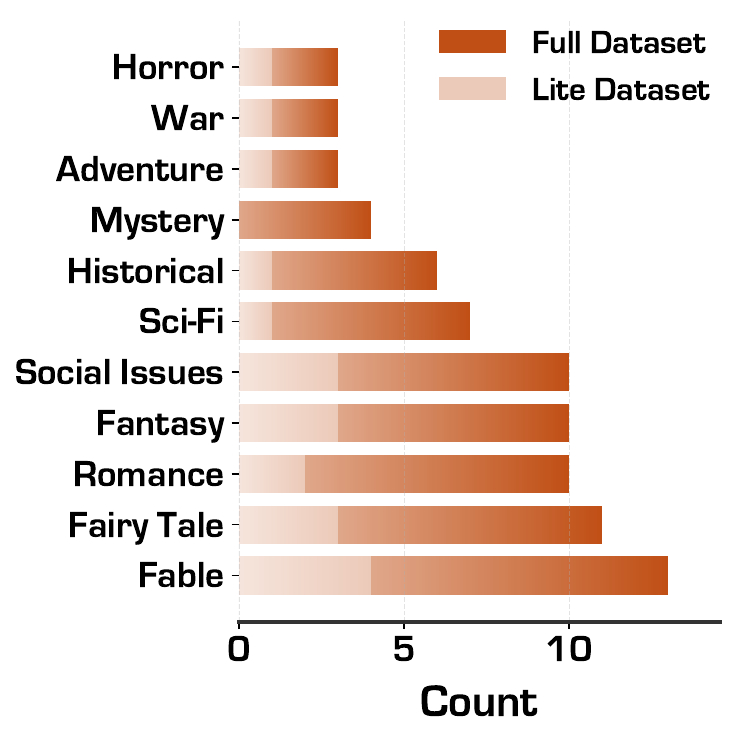}
    \caption{\textbf{Theme Distribution (Lite vs. Full)}.}
    \label{fig:main_char_dist}
  \end{subfigure}%
  \hfill 
  \begin{subfigure}{0.28\textwidth}
    \includegraphics[width=\linewidth]{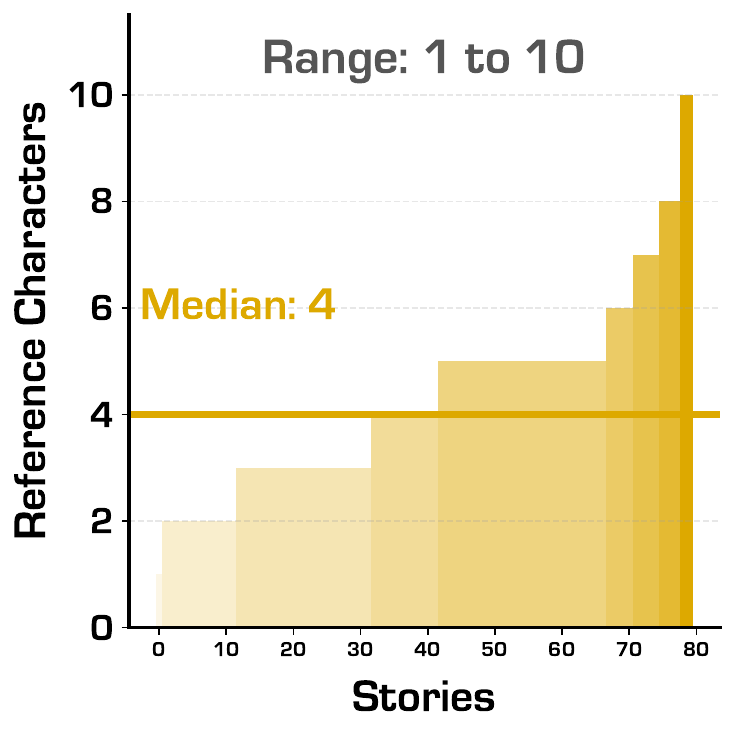}
    \caption{\textbf{Casts per Story.}}
    \label{fig:main_lite_full}
  \end{subfigure}
  \vspace{-1ex}
  \caption{
    \textbf{ViStoryBench Dataset Statistics Overview.} 
    (a) Distribution of story style and cultural origin of our dataset. 
    (b) Theme distribution comparison of Full and Lite subset, showing high statistical similarity. 
    (c) Distribution of reference characters number per story. 
    \textbf{See Appendix~\ref{sec:dataset}, \ref{sec:lite} for details.}
  }
  \label{fig:dataset_stats_main}
  \vspace{-2ex}
\end{figure*}


	\begin{figure*}[h]
		\centering
		\includegraphics[width=0.97\linewidth]{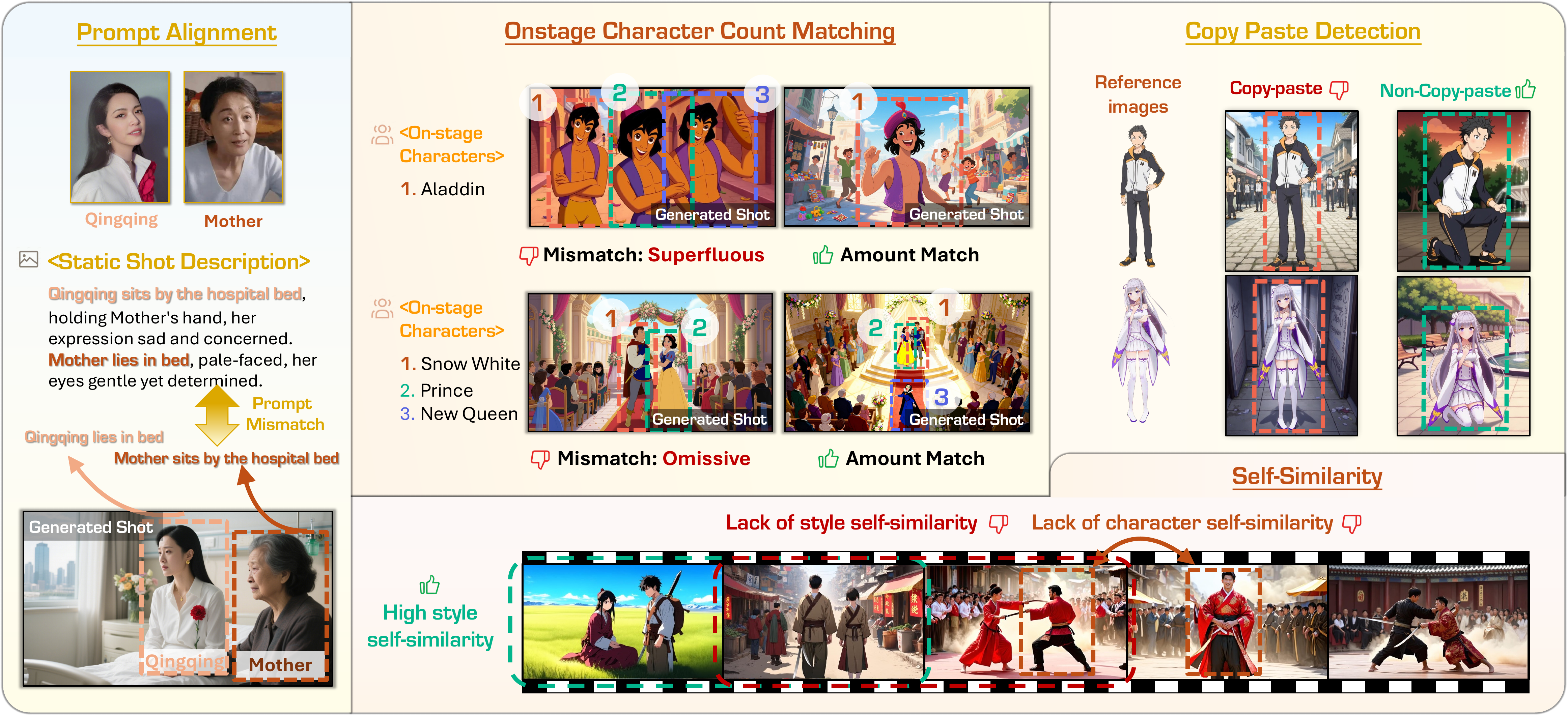}
		\caption{
        \small \textbf{Case Study of Failure in Story Visualization.}
\highlighttext{PAtext}{PAback}{\textbf{Prompt Alignment}}~: Alignment of character interaction, individual actions, camera design, and scene setting descriptions. 
\highlighttext{OCCMtext}{OCCMback}{\textbf{Onstage Character Count Matching}}~: Whether the number of characters in the generated shots matches the script. 
\highlighttext{CPtext}{CPback}{\textbf{Copy-Paste Detection}}~: Quantifies the tendency to replicate reference images rather than generating novel instances.
\highlighttext{SIMtext}{SIMback}{\textbf{Self \& Cross Similarity}}~: Style and identity consistency of generated characters with reference images and across shots.
}
		\label{fig:fail}
        \vspace{-1em}
	\end{figure*}

	\section{ViStoryBench}
	\label{gen_inst}
	
	\subsection{Problem Definition}
	
	Based on some prior work~\cite{li2019storygan,lin2022collaborative,yang2024seedstory}, we design a story generation task with comprehensive input conditions. Given a story script, we first provide the appearance descriptions $T_1$, $T_2$, $\dots$, $T_n$ and corresponding images $S_1$, $S_2$, $\dots$, $S_n$ for $n$ characters $C_1$, $C_2$, $\dots$, $C_n$, where $T_i$ and $S_i$ are consistent with each other, and $C_i = (T_i, S_i)$. Next, we provide $m$ storyboard shot descriptions: $\text{Shot}_1$, $\text{Shot}_2$, $\dots$, $\text{Shot}_m$. Each $\text{Shot}_i$ includes a text description that contains the following components:	
    \begin{enumerate}[label=\textcircled{\scriptsize\arabic*}, leftmargin=1.4em, parsep=-0.05em]
		\item \textbf{Setting Description}: A description of the current scene's setting
		\item \textbf{Plot Correspondence}: The story segment from the original narrative corresponding to this shot.
		\item \textbf{Onstage Characters}: A list of characters present in the current shot.
		\item \textbf{Static Shot Description}: A description of the static actions or positions of characters and objects in the frame, representing a fixed visual state.
		\item \textbf{Shot Perspective Design}: Cinematographic information, including shot size, shot type, and camera angle.
    \end{enumerate}
	
	The objective of the story visualization task is to generate a sequence of images $I_1...I_m$ that faithfully represent the described storyboard shots, adhering to the provided prompts and character information. This involves accurately depicting the characters, their actions, the scene settings, and the specified camera perspectives. The specific quantitative evaluation methods are described later.

    \subsection{Source Data}

    \paragraph{Story and Script.} To ensure narrative diversity, we manually curate 80 story segments from a wide range of sources, including film and television scripts, literary classics, world folklore, novels, and picture books. For lengthy stories, we employ LLMs to assist in summarization, producing concise versions of several hundred words. Each story is then adapted into a script containing character descriptions and storyboards, also with LLM assistance. All LLM-generated content is manually reviewed to ensure narrative coherence and logical consistency. 
    \paragraph{Character Reference Images.} For each character, we manually collect reference images from the Internet that align with the textual descriptions, ensuring a consistent visual style within each story. A small subset of images is generated using SDXL~\cite{podell2023sdxl}. 
    The final dataset comprises 344 characters and 509 reference images, categorized the 80 stories into 10 distinct style genres for stylistic diversity. Details on data collection, style distribution, and statistics are provided in \textbf{Appendix}~\ref{sec:dataset} and visualized in Figure~\ref{fig:dataset_stats_main}.

	\subsection{Evaluation Metrics}
	
	\paragraph{Overview.} We present a comprehensive suite of metrics to assess story visualization models across multiple key dimensions. To provide intuitive motivation for our evaluation metrics, Figure~\ref{fig:fail} illustrates typical failure cases in story visualization. The metrics cover the following aspects:

    \begin{enumerate}[label=\textcircled{\scriptsize\arabic*}, leftmargin=1.4em, parsep=-0.05em]
        \item \textbf{Cross- and Self-Similarity}: Assessing the resemblance between generated and reference images, as well as the consistency within the generated images themselves.
        
        \item \textbf{Prompt Alignment}: Measuring how well the generated images align with the storyboard descriptions provided in the prompts. 

        \item \textbf{Character Matching}: We calculate the accuracy of the number of characters in each generated image.
        
        \item \textbf{Aesthetic}: We evaluate the aesthetic metric, generation quality and diversity of the generated results.

        \item \textbf{Copy-Paste}: We specifically design a Copy-Paste Detection metric to check if the model excessively replicates the character reference images.
    \end{enumerate}

    \paragraph{Preliminaries.} 
    We briefly introduce the core models and tools used in our evaluation pipeline. Grounding DINO~\cite{zhang2022dino,liu2024grounding} serves as our open-set object detector for cropping character regions. For feature extraction, we use ArcFace~\cite{deng2019arcface}, AdaFace~\cite{kim2022adaface}, and FaceNet~\cite{schroff2015facenet} for realistic characters, and CLIP~\cite{radford2021learning} otherwise. The CLIP-based model from CSD~\cite{somepalli2024measuring} is employed for style feature extraction. We use the Inception Score for diversity and the Aesthetic Predictor V2.5 for quality evaluation.

    Next, we briefly describe the calculation procedure for each metric, with additional details provided in \textbf{Appendix}~\ref{sec:prompt_adherance}, \ref{sec:cids}, \ref{sec:csd}. For clarity in presentation, both the cosine similarity scores and the original $0-4$ rating scale are converted to a $100$-point scale.

    \begin{figure*}[h]
    	\centering
        \vspace{-2ex}
    	\includegraphics[width=0.9\linewidth]{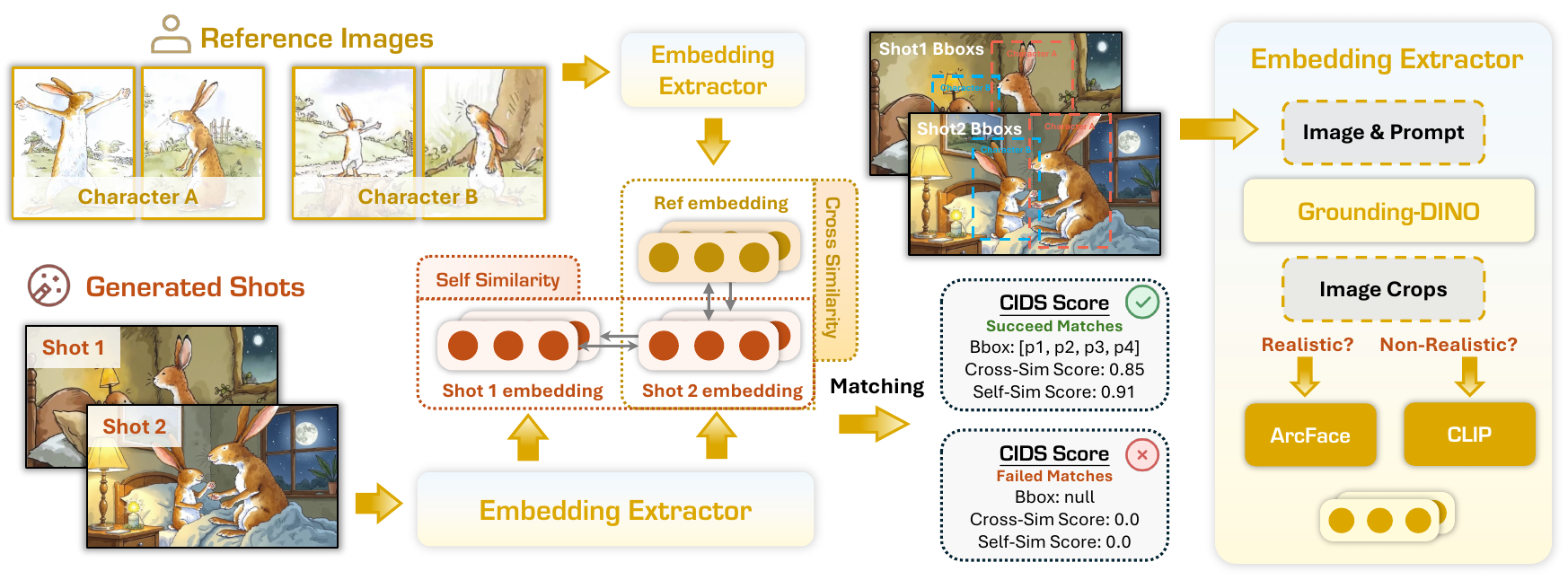}
    	\caption{\small \textbf{Character Identification Similarity (CIDS) Metric}. Evaluating both cross-similarity and self-consistency by detecting and cropping character regions from reference and generated images, then computing cosine similarity between matched character features.} 
        \vspace{-2ex}
        \label{fig:CIDS}
    \end{figure*}

    \paragraph{Character Identification Similarity (CIDS).}
    The CIDS computation pipeline comprises four stages: 
    \begin{enumerate}[label=\textcircled{\scriptsize\arabic*}, leftmargin=1.4em, parsep=-0.05em]
        \item \textbf{Character Detection}: Grounding DINO crops character regions from reference and generated images. For character reference, Grounding DINO typically trims the edges. For generated images, however, it may fail entirely, returning an empty result; 
        \item \textbf{Feature Extraction}: CLIP or face models convert crops to 512-d feature vectors; 
        \item \textbf{Character Matching}: Similarity computation and bipartite matching to find optimal character correspondences; 
        \item \textbf{Scoring}: Average cosine similarity of matched pairs yields the final score.
    \end{enumerate}

    \paragraph{Style Similarity.}
    The CSD computation pipeline comprises four stages: 
    \begin{enumerate}[label=\textcircled{\scriptsize\arabic*}, leftmargin=1.4em, parsep=-0.05em]
        \item \textbf{Image Encoding}: Encoding each image using a style-trained CLIP vision encoder; 
        \item \textbf{Feature Extraction}: Extracting style features through CSD layers; 
        \item \textbf{Similarity Computing}: Computing pairwise cosine similarity between all style embeddings; 
        \item \textbf{Scoring}: Averaging the scores over all valid pairs to produce the final score.
    \end{enumerate}

    \paragraph{Prompt Alignment.} Regarding the similarity between the generated images and the provided storyboard descriptions, we categorize them as follows:

    \begin{enumerate}[label=\textcircled{\scriptsize\arabic*}, leftmargin=1.4em, parsep=-0.05em]

        \item \textbf{Scene Score}: Overall correspondence between the generated scene and the narrative details provided in the storyboard's static shot description, including setting, mood, and layout.

        \item \textbf{Shot Score}: Consistency between the depicted camera perspective (e.g., close-up, wide shot, over-the-shoulder) in the generated image and the specified shot design in the storyboard.

        \item \textbf{Character Interaction (CI)}: Alignment between the group-level interactions of characters in the generated image and the intended interactions described in the storyboard's static shot description.

        \item \textbf{Individual Actions (IA)}: Accuracy of the gestures, expressions, and poses of each character in the generated image relative to their described behavior in the static shot description.

    \end{enumerate}

    \begin{figure*}
    \centering
    \includegraphics[width=0.97\linewidth]{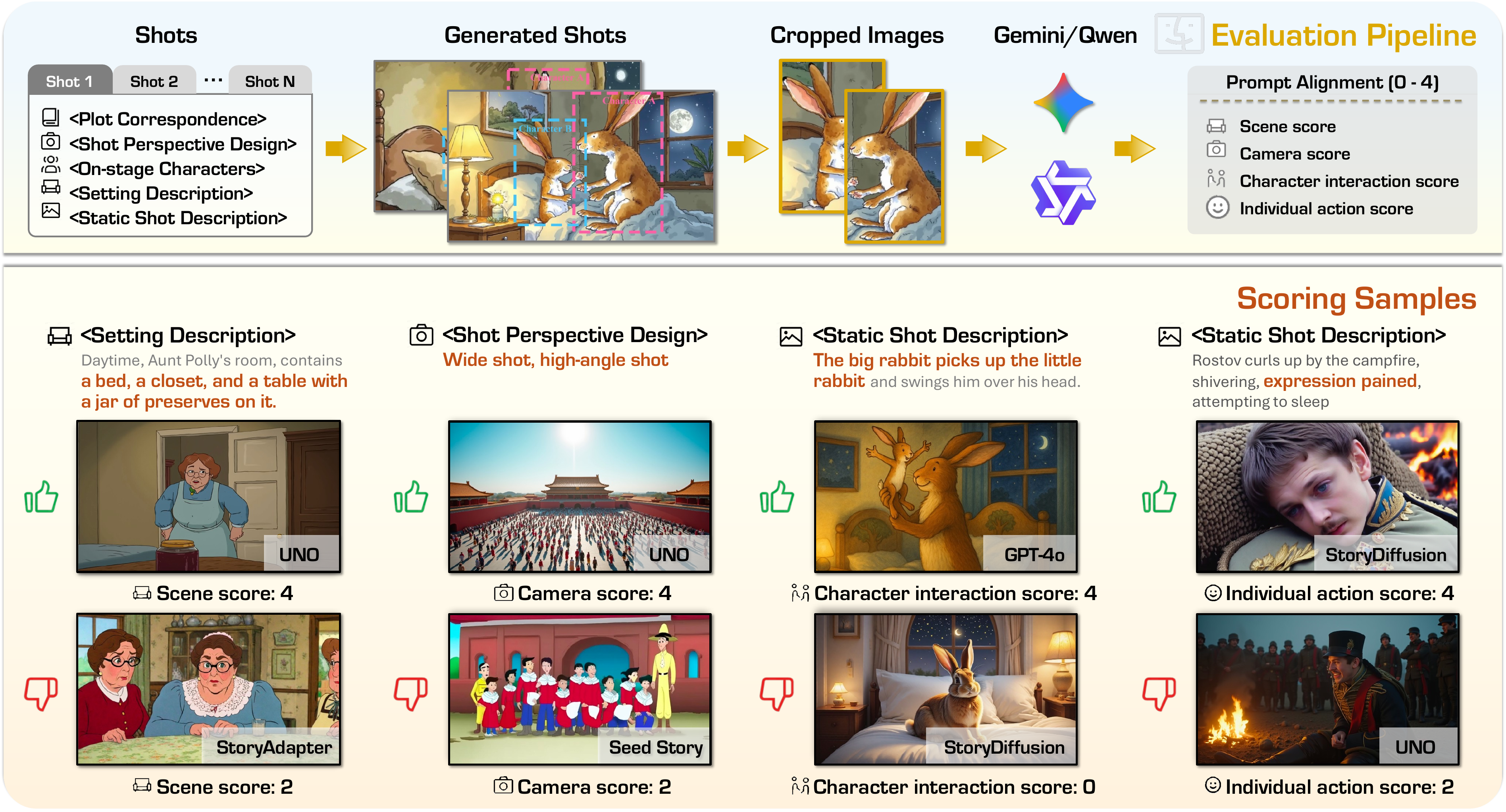}
    \caption{\small \textbf{Prompt Alignment Evaluation}. Based on the descriptions, \textit{Scene Score}, \textit{Shot Score}, \textit{Character Interaction} and \textit{Individual Action} are evaluated via Gemini-3-Pro for best evaluation accuracy, Qwen3-VL~\cite{yang2025qwen3} for reproducibility.}
    \label{fig:gpt}        
\vspace{-1em}
\end{figure*}
    
    We primarily use Gemini-3-Pro for automated evaluation. As shown in Figure~\ref{fig:gpt}, we have established \textit{Likert-scale questionnaires} to evaluate the consistency of each generated image on a scale from $0$ to $4$. 
    
    To validate the reliability of this VLM-based approach, we conducted a rigorous stability analysis, which confirms the low variance of our evaluation scores. The detailed methodology and results are provided in \textbf{Appendix}~\ref{sec:appendix_stability}.

    \paragraph{Onstage Character Count Matching (OCCM).}
    We observe that various models struggle to generate the correct number of onstage characters as specified in the script, often leading to superfluous additions (hallucinations) or omitted figures. To quantify this critical capability, we introduce the Onstage Character Count Matching (OCCM) score. It is important to note that this metric relies on an upstream character detector to obtain the detected character count ($D$). Nevertheless, the OCCM formula itself is designed to provide a fair and robust measure of numerical consistency. The score is calculated as:
    
    \begin{equation}
    \text{OCCM} = \exp{\left(-\frac{|D - E|}{\epsilon + E}\right)} \times 100\%
    \label{eq:occm}
    \end{equation}
    
    where $D$ is the detected number of characters, $E$ is the expected count from the storyboard, and $\epsilon=10^{-6}$ is a small smoothing factor to prevent division by zero. The detailed explanation of performance bounds and the design principle is provided in \textbf{Appendix}~\ref{sec:occm}.

    \paragraph{Copy-Paste Detection.}
    A common shortcut in story visualization is directly reusing the input character reference images, which compromises generation diversity and fidelity to the prompt. To quantify this behavior, we propose a geometrically-normalized \textbf{Copy-Paste Rate} metric. This metric assesses whether a generated character feature $g$ is closer to its specific input reference feature $r$ or to a feature from a different reference image $t$ of the \emph{same character}. Here, the second reference $t$ serves as a proxy for a generalized target, helping to determine if the model is merely copying the input $r$ instead of learning the character's general appearance. The detailed explanation of the computational principle is provided in \textbf{Appendix}~\ref{sec:copy_paste}.

    \paragraph{Image Quality.}
    We evaluate visual quality through two complementary metrics averaged across all  images:
    \begin{itemize}
        \item \textbf{Inception Score (Inc)}~\cite{salimans2016improved} measures diversity and clarity using an Inception V3 backbone.
        \item \textbf{Aesthetic Score}~\cite{aestheticpredictor2024} assesses aesthetic perceptual quality on a 1-10 scale via Aesthetic Predictor V2.5~\cite{aestheticpredictor2024} (a SigLIP-based predictor~\cite{zhai2023sigmoid}), where scores below 5.5 typically indicate blurry, noisy, or unappealing images.
    \end{itemize}

	\section{Experiments and Analysis}
	\subsection{Experimental Setup}

\begin{table*}[tb]
    \caption{
        \small\textbf{Quantitative Results of Various Story Visualization Methods on ViStoryBench and ViStoryBench-Lite}. Results highlighted with a gray background are excluded from ranking. the Copy-Paste Baseline directly pastes the character reference image as the output. For certain methods, we evaluate multiple inference configurations and report all corresponding results. \legendsquare{colorfirst} \legendsquare{colorsecond} \legendsquare{colorthird} \legendsquare{colorfourth} \legendsquare{colorfifth} indicate the first, second, third, fourth, and fifth performance, respectively. We list full results on ViStoryBench-Lite in \textbf{Appendix}~\ref{sec:lite}. \textbf{CSD}: Style Similarity; \textbf{CIDS}: Character Similarity; \textbf{PA}: Prompt Alignment Score (CI: Character Interaction, IA: Individual Action); \textbf{CM}: OCCM; \textbf{Inc}: Inception Score; \textbf{Aes}: Aesthetics Score; \textbf{CP}: Copy-Paste. \imageref: With image reference; \textref: Only text input; \autoregressive: Auto-regressive mode; superscript $^k$ means scale=$k$. \textit{Note: \textbf{PA} scores are based on Gemini-3 Pro.}
    }
    \vspace{1mm}
    \centering
    \resizebox{0.97\textwidth}{!}{ 

    \begin{tabular}{l|c|cccc|ccccc|cccc}
        \toprule		
        \multirow{2}{*}[-0.28em]{\textbf{Method}} 
        &\multirow{2}{*}[-0.28em]{\textbf{Model}} 
        &\multicolumn{2}{c}{\textbf{CSD$\uparrow$}} 
        &\multicolumn{2}{c}{\textbf{CIDS$\uparrow$}} 
        &\multicolumn{5}{c}{\textbf{PA$\uparrow$}} 
        &\multirow{2}{*}[-0.28em]{\textbf{CM}$\uparrow$} 
        &\multirow{2}{*}[-0.28em]{\textbf{Inc}$\uparrow$} 
        &\multirow{2}{*}[-0.28em]{\textbf{Aes}$\uparrow$} 
        &\multirow{2}{*}[-0.28em]{\textbf{CP}}\\
        \cmidrule(l{7pt}r{7pt}){3-4}
        \cmidrule(l{7pt}r{7pt}){5-6}
        \cmidrule(l{7pt}r{7pt}){7-11} 
    
        \vspace{0.2em}
         & & Cross & Self & Cross & Self & Scene & Shot & CI & IA & Avg. \\

        \midrule
        \tablespacing \\
\multicolumn{15}{c}{\textit{\texttt{\textbf{The following results are obtained on ViStoryBench}}}} \\
        \midrule
        Copy-Paste Baseline & - & \cellcolor{mygray}{0.728} & \cellcolor{mygray}{0.712} & \cellcolor{mygray}{0.929} & \cellcolor{mygray}{0.984} & \cellcolor{mygray}{0.34} & \cellcolor{mygray}{1.75} & \cellcolor{mygray}{0.70} & \cellcolor{mygray}{0.81} & \cellcolor{mygray}{0.90} & \cellcolor{mygray}{89.4} & \cellcolor{mygray}{6.71} & \cellcolor{mygray}{4.48} & \cellcolor{mygray}{0.474} \\
        
        \midrule
        \rowcolor{teaserdatasetback}\multicolumn{15}{c}{\myfont \textbf{Story Image Method}} \\ \midrule
        \rowcolor{teaserdatasetback}StoryGen~\cite{Liu_2024_CVPR_Storygen}~\autoregressive & SD1.5 & \rankfifth{0.379} & 0.540 & 0.428 & 0.576 & 0.58 & 1.95 & 0.53 & 0.49 & 0.89 & 51.1 & 8.73 & 4.02 & 0.246 \\ 
        \rowcolor{teaserdatasetback}StoryGen~\cite{Liu_2024_CVPR_Storygen}~\imageref & SD1.5 & 0.371 & 0.531 & 0.417 & 0.568 & 0.63 & 2.05 & 0.53 & 0.40 & 0.90 & 50.8 & 8.89 & 4.02 & 0.228 \\ 
        \rowcolor{teaserdatasetback}StoryGen~\cite{Liu_2024_CVPR_Storygen}~\autoregressive\imageref & SD1.5 & 0.283 & 0.580 & 0.414 & 0.593 & 0.53 & 2.05 & 0.48 & 0.39 & 0.86 & 41.1 & 7.31 & 3.74 & 0.227 \\ 
        \rowcolor{teaserdatasetback}TheaterGen~\cite{cheng2024theatergen} & SD1.5 & 0.184 & 0.392 & 0.348 & 0.578 & 1.99 & 1.86 & 0.46 & 0.33 & 1.16 & 55.4 & \rankfourth{14.89} & 4.90 & 0.189 \\
        \rowcolor{teaserdatasetback}StoryDiffusion~\cite{zhou2024storydiffusion}~\textref & SDXL & 0.269 & \rankfifth{0.628} & 0.397 & \rankfourth{0.622} & 2.62 & 2.30 & 1.33 & 1.15 & 1.85 & 62.9 & \ranksecond{15.72} & \rankthird{5.76} & 0.181\\
        \rowcolor{teaserdatasetback}StoryDiffusion~\cite{zhou2024storydiffusion}~\imageref & SDXL & 0.340 & 0.547 & \rankfifth{0.436} & 0.565 & 1.34 & \rankfifth{2.54} & 1.39 & 1.30 & 1.64 & 57.4 & 10.06 & 5.13 & 0.216 \\
        \rowcolor{teaserdatasetback}SEED-Story~\cite{yang2024seedstory} & SDXL & 0.227 & \textbf{\rankfirst{0.748}} & 0.287 & 0.587 & 1.43 & 1.64 & 0.41 & 0.22 & 0.93 & 44.4 & 6.30 & 3.82 & 0.186 \\
        \rowcolor{teaserdatasetback}Story-Adapter~\cite{mao2024story_adapter}~\imageref$^0$ & SD1.5 & \textbf{\rankfirst{0.456}} & 0.548 & \rankfourth{0.460} & 0.605 & 1.52 & \rankthird{2.60} & 1.81 & 1.66 & 1.90 & \rankfifth{69.0} & 12.98 & 4.99 & 0.198 \\ 
        \rowcolor{teaserdatasetback}Story-Adapter~\cite{mao2024story_adapter}~\imageref$^5$ & SD1.5 & 0.325 & \ranksecond{0.737} & 0.401 & \ranksecond{0.626} & 1.42 & \rankfourth{2.56} & 1.58 & 1.43 & 1.75 & 59.3 & 13.73 & 4.89 & 0.194 \\
        \rowcolor{teaserdatasetback}Story-Adapter~\cite{mao2024story_adapter}~\textref$^0$ & SD1.5 & 0.280 & 0.462 & 0.400 & 0.538 & 1.57 & \textbf{\rankfirst{2.70}} & 1.77 & 1.63 & 1.92 & 63.1 & \textbf{\rankfirst{16.34}} & 5.17 & 0.195 \\ 
        \rowcolor{teaserdatasetback}Story-Adapter~\cite{mao2024story_adapter}~\textref$^5$ & SD1.5 & 0.318 & \rankthird{0.733} & 0.395 & \rankthird{0.624} & 1.41 & 2.54 & 1.53 & 1.36 & 1.71 & 59.6 & 13.13 & 4.90 & 0.193 \\
        \rowcolor{teaserdatasetback}UNO~\cite{wu2025uno} & FLUX1 & \rankthird{0.391} & 0.602 & \ranksecond{0.485} & \rankfifth{0.620} & 3.11 & 2.40 & \rankfifth{1.92} & \rankfifth{1.78} & \rankfifth{2.30} & \rankthird{74.2} & 12.40 & 5.23 & 0.234 \\
        \rowcolor{teaserdatasetback}OmniGen2~\cite{wu2025omnigen2} & DiT & \ranksecond{0.454} & 0.600 & \textbf{\rankfirst{0.548}} & \textbf{\rankfirst{0.647}} & \rankthird{3.16} & \ranksecond{2.68} & \rankfourth{2.14} & \rankfourth{1.98} & \rankfourth{2.49} & \rankfourth{70.2} & 11.05 & 5.25 & 0.275 \\
        \rowcolor{teaserdatasetback}CharaConsist~\cite{wang2025characonsist} & FLUX1 & 0.282 & 0.553 & 0.315 & 0.519 & \rankfourth{3.15} & 2.45 & 1.83 & 1.53 & 2.24 & 57.9 & \rankfifth{13.80} & \ranksecond{5.88} & 0.172 \\
        \rowcolor{teaserdatasetback}QwenImageEdit-2509~\cite{qwen2025qwenimage} & DiT & \rankfourth{0.381} & 0.593 & \rankthird{0.475} & 0.574 & \ranksecond{3.21} & 2.32 & \rankthird{2.39} & \rankfirst{2.13} & \rankthird{2.51} & 59.8 & 13.42 & \rankfifth{5.50} & 0.218 \\

        \midrule
        \rowcolor{blue!3}\multicolumn{15}{c}{\myfont \textbf{Story Video Method}} \\ \midrule
        \rowcolor{blue!3}Vlogger~\cite{zhuang2024vlogger}~\textref  & SD1.4 & 0.201 & 0.407 & 0.346 & 0.548 & 1.15 & 2.32 & 1.56 & 1.34 & 1.59 & \ranksecond{75.4} & 10.29 & 4.28 & 0.194\\
        \rowcolor{blue!3}Vlogger~\cite{zhuang2024vlogger}~\imageref & SD1.4 & 0.259 & 0.453 & 0.362 & 0.554 & 1.18 & 2.28 & 1.55 & 1.39 & 1.60 & \textbf{\rankfirst{76.6}} & 9.77 & 4.28 & 0.200\\ 
        \rowcolor{blue!3}AnimDirector~\cite{li2024anim} & SD3 & 0.288 & 0.510 & 0.401 & 0.578 & \textbf{\rankfirst{3.30}} & 2.34 & \ranksecond{2.48} & \ranksecond{2.09} & \rankfirst{2.55} & 67.4 & 12.02 & \rankfourth{5.59} & 0.212\\ 
        \rowcolor{blue!3}MMStoryAgent~\cite{xu2024mmstoryagent} & SDXL & 0.238 & \rankfourth{0.669} & 0.388 & 0.596 & 2.41 & 2.12 & 1.25 & 1.00 & 1.69 & 61.5 & 9.09 & \textbf{\rankfirst{5.88}} & 0.198\\ 
        \rowcolor{blue!3}MovieAgent~\cite{wu2025movieagent} & SD1.5 & 0.193 & 0.502 & 0.360 & 0.560 & 0.92 & 1.83 & 0.76 & 0.61 & 1.03 & 63.6 & 11.61 & 4.63 & 0.198 \\
        \rowcolor{blue!3}MovieAgent~\cite{wu2025movieagent} & SD3 & 0.299 & 0.479 & 0.400 & 0.544 & \rankfifth{3.14} & 2.50 & \rankfirst{2.53} & \rankthird{2.00} & \ranksecond{2.54} & 64.6 & \rankthird{14.99} & 5.32 & 0.209 \\ 
        
\hline \hline \multicolumn{15}{c}{} \vspace{-0.2cm}\\ 
        \multicolumn{15}{c}{\textit{\texttt{\textbf{The following results are obtained on ViStoryBench-Lite}}}} \\
        \midrule
        \rowcolor{mygray}
        Copy-Paste Baseline & - & 0.911 & 0.994 & 0.550 & 0.735 & 0.42 & 1.60 & 0.76 & 0.85 & 0.91 & 92.76 & 5.46 & 4.39 & 1.000 \\
        
        \midrule
        \rowcolor{SIMback}\multicolumn{15}{c}{\myfont \textbf{Commercial Platform}} \\ \midrule
        \rowcolor{SIMback}MOKI~\cite{moki2024} & - & 0.214 & \rankfourth{0.694} & 0.372 & 0.621 & 2.29 & 1.56 & 0.58 & 0.45 & 1.22 & 45.96 & 10.36 & \textbf{\rankfirst{5.79}} & 0.211 \\
        \rowcolor{SIMback}MorphicStudio~\cite{morphic2024studio} & - & \textbf{\rankfirst{0.577}} & 0.628 & \ranksecond{0.603} & \rankfourth{0.677} & 3.01 & 2.31 & 1.77 & 1.32 & 2.10 & 60.79 & 9.00 & 4.96 & 0.234 \\
        \rowcolor{SIMback}AIbrm~\cite{brmgo2025} & - & \rankfifth{0.412} & \textbf{\rankfirst{0.730}} & \rankfifth{0.557} & \ranksecond{0.740} & 3.06 & 2.18 & 1.67 & 1.55 & 2.12 & \ranksecond{75.53} & 9.53 & \ranksecond{5.72} & 0.223 \\
        \rowcolor{SIMback}ShenBi~\cite{shenbi2025} & - & 0.275 & 0.575 & 0.418 & 0.585 & 3.49 & 2.35 & 2.65 & 2.11 & 2.65 & 61.33 & \rankthird{11.60} & 5.07 & 0.197 \\
        \rowcolor{SIMback}Typemovie~\cite{typemovie2024} & - & 0.325 & 0.646 & 0.464 & 0.621 & 2.34 & 2.17 & 1.62 & 1.40 & 1.88 & \rankfourth{74.14} & \rankfourth{11.15} & 5.32 & 0.168 \\
        \rowcolor{SIMback}Doubao~\cite{doubao2024} & - & 0.367 & \rankthird{0.695} & 0.446 & 0.642 & \rankthird{3.88} & 2.41 & \rankfourth{3.23} & \rankfifth{2.65} & \rankfifth{3.04} & 65.23 & 9.88 & \rankthird{5.61} & 0.255 \\
        
        \midrule
        \rowcolor{red!5}\multicolumn{15}{c}{\myfont \textbf{Multi-modal Large Model}} \\ \midrule
        \rowcolor{red!5}GPT-4o$^*$~\cite{hurst2024gpt4o} & - & \rankthird{0.481} & 0.680 & 0.420 & 0.522 & \rankfifth{3.82} & \ranksecond{2.82} & \ranksecond{3.58} & \ranksecond{3.12} & \ranksecond{3.34} & 69.33 & 9.02 & 5.49 & 0.209 \\
        \rowcolor{red!5}Gemini-2.0$^*$~\cite{gemini2.02025} & - & 0.361 & 0.573 & \rankfourth{0.573} & \rankthird{0.677} & 3.26 & 2.46 & 2.43 & 2.00 & 2.54 & \rankthird{74.82} & 10.12 & 4.91 & 0.266 \\
        \rowcolor{red!5}Gemini-2.5$^*$~\cite{google2025nanobanana} & - & \rankfourth{0.447} & 0.657 & 0.553 & 0.642 & \ranksecond{3.89} & 2.32 & \rankthird{3.26} & \rankthird{2.90} & \rankthird{3.09} & 64.86 & \rankfifth{10.54} & \rankfourth{5.61} & 0.255 \\
        \rowcolor{red!5}Gemini-3.0 Pro$^*$~\cite{google2025nanobananapro} & - & 0.385 & 0.622 & \rankthird{0.581} & \rankfifth{0.653} & \textbf{\rankfirst{3.94}} & \rankfifth{2.58} & \textbf{\rankfirst{3.71}} & \textbf{\rankfirst{3.25}} & \textbf{\rankfirst{3.37}} & 59.97 & \textbf{\rankfirst{12.50}} & \rankfifth{5.54} & 0.244 \\
        \rowcolor{red!5}Seedream-4.0~\cite{gong2025seedream} & - & 0.369 & 0.585 & 0.280 & 0.539 & \rankfourth{3.82} & \rankfourth{2.59} & \rankfifth{3.20} & \rankfourth{2.68} & \rankfourth{3.07} & 49.45 & \ranksecond{12.12} & 5.21 & 0.201 \\
        \rowcolor{red!5}Sora2$^*$~\cite{2025sora2}~\imageref & - & \ranksecond{0.515} & \ranksecond{0.713} & \textbf{\rankfirst{0.766}} & \textbf{\rankfirst{0.839}} & 3.08 & \rankthird{2.76} & 2.91 & 2.50 & 2.81 & \textbf{\rankfirst{81.42}} & 6.53 & 4.72 & 0.158 \\
        \rowcolor{red!5}Sora2$^*$~\cite{2025sora2}~\textref & - & 0.365 & 0.685 & 0.364 & 0.561 & 3.01 & \rankfirst{2.83} & 2.96 & \rankfifth{2.33} & 2.78 & \rankfifth{72.67} & 9.68 & 4.52 & 0.158 \\
        \bottomrule
    \end{tabular}
    }
    \label{table:main}
    \vspace{-1em}
\end{table*}

    To address the high costs associated with user studies and commercial platform evaluations, we introduce ViStoryBench-Lite, a strategically curated subset for efficient yet representative assessment. This one-quarter subset of the full ViStoryBench was meticulously constructed to preserve statistical alignment with the full dataset in text styles and character reference distributions, as illustrated by the theme distribution alignment in Figure ~\ref{fig:dataset_stats_main}. It includes 20 carefully selected stories with 36 animated, 41 realistic, and 43 non-human characters, closely mirroring the full dataset's composition. As detailed in \textbf{Appendix}~\ref{sec:lite}, performance correlation between the Lite and full versions is consistently high, validating its reliability as a cost-effective alternative for large-scale evaluation.

    In our main experiments (Table~\ref{table:main}), we evaluate a diverse range of image and video generation methods. For image generation, we assess StoryGen~\cite{Liu_2024_CVPR_Storygen}, TheaterGen~\cite{cheng2024theatergen}, StoryDiffusion~\cite{zhou2024storydiffusion}, SEED-Story~\cite{yang2024seedstory}, Story-Adapter~\cite{mao2024story_adapter}, UNO~\cite{wu2025less}, OmniGen2~\cite{wu2025omnigen2}, CharaConsist~\cite{wang2025characonsist}, and QwenImageEdit-2509~\cite{qwen2025qwenimage}. We also evaluate multi-modal large models, including GPT-4o~\cite{hurst2024gpt4o}, Gemini-2.0~\cite{gemini2.02025}, Gemini-2.5~\cite{comanici2025gemini2.5} (Nano Banana), and Seedream-4.0~\cite{gong2025seedream}. For video generation, we test Vlogger~\cite{zhuang2024vlogger}, AnimDirector~\cite{li2024anim}, MMStoryAgent~\cite{xu2024mmstoryagent}, MovieAgent~\cite{wu2025automated}, and the native multi-shot model Sora2~\cite{2025sora2}. Additionally, we include a simple Copy-Paste Baseline that programmatically places reference images on a 1080p canvas. Due to resource constraints, commercial software results, including MOKI~\cite{moki2024}, MorphicStudio~\cite{morphic2024studio}, AIbrm~\cite{brmgo2025}, ShenBi~\cite{shenbi2025}, Typemovie~\cite{typemovie2024}, and Doubao~\cite{doubao2024}, are reported only on the Lite version of the dataset using their May 2025 releases.

    Detailed adaptations for the varying method (Image and video) requirements are discussed in \textbf{Appendix}~\ref{sec:implementation_detail}. For video methods without intermediate images, we use the keyframe of each shot. Most methods produce 1080p outputs, except for some like Gemini-2.0 with limited resolution control.

    A public leaderboard will be maintained to foster community competition, ranking models by averaging metric ranks for balanced evaluation.

	\paragraph{User Study.}
	
	To evaluate the consistency and aesthetic quality of the generated images, we conduct a user study involving participants who assessed the results across three dimensions: Environment Consistency, Character Identification Consistency, and Subjective Aesthetics Score. Environment Consistency focused on whether scenes corresponding to the same environment description appeared visually cohesive. For Character Identification Consistency, participants rated how consistently the main characters were identifiable and coherent throughout the story. Subjective Aesthetics Score assessed the overall artistic appeal, detail richness, and storytelling effectiveness of the visualizations. Detailed scoring criteria for each dimension are provided in \textbf{Appendix}~\ref{sec:user_study}. The detailed scoring results will also be open-sourced. We list the top four methods in each category based on user ratings (0-4 scores, scale to 100\%):

    \begin{enumerate}[label=\textcircled{\scriptsize\arabic*}, leftmargin=1.4em, parsep=-0.05em]
    
    \item \textbf{Environment Consistency}: UNO~(3.28 82.0\%), GPT-4o~(3.25 81.2\%), Doubao~(3.22 80.4\%), Story-Adapter~\raisebox{-0.2em}{\includegraphics[width=0.042\linewidth]{image/image-ref.png}} (3.18 79.6\%) 


    \item \textbf{Character Identification Consistency}: Doubao~(3.70 92.6\%), AIbrm~(3.54 88.4\%), UNO~(3.36 84.0\%), GPT-4o~(3.26 81.6\%)
    
    \item \textbf{Subjective Aesthetics Score}: GPT-4o~(3.42 85.6\%), Doubao~(3.4 85.0\%), AIbrm~(3.32 83.0\%), UNO~(3.21 80.4\%)
    \end{enumerate}
	
	We establish one-to-one correspondences between three human evaluation metrics and automated metrics, similar to Theatergen~\cite{cheng2024theatergen}: Character Identification Consistency corresponds to CIDS, Environment Consistency corresponds to Style Similarity, and Subjective Aesthetics Score to Aesthetic Quality. Subsequently, we evaluate the correlation coefficients between automated metrics and human evaluation results, including Kendall's $\tau$, Spearman's $\rho$, and Pearson's $\sigma$, as shown in Table~\ref{table:correlation_analysis}. These demonstrate that automated metrics can effectively reflect human preferences.

    Due to time constraints, the user study was limited to commercial method versions available until mid-May 2025, and we excluded obvious outliers like StoryGen~\cite{Liu_2024_CVPR_Storygen} from correlation analysis to ensure valid metric validation. As detailed in \textbf{Appendix}~\ref{sec:user_study}.

\begin{table}[tb]
\caption{
		\small \textbf{Correlation Analysis}. Kendall's $\tau$, Spearman's $\rho$, and Pearson's $\sigma$ coefficients between human evaluations and automated metrics are reported.
	}
	\centering
	\resizebox{0.4\textwidth}{!}{
    \begin{tabular}{l|ccc}
		\toprule		
		\textbf{Metrics} 
&\textbf{Self CSD} &\textbf{Self CIDS} & \textbf{Aesthetics} \\
\addlinespace[-1pt]
		\midrule
		$\tau$ & 0.4165 & 0.4978 & 0.2565 \\
        $\rho$ & 0.5648 & 0.6759 & 0.3966 \\
		$\sigma$ & 0.6042 & 0.7956 & 0.5411 \\
        \bottomrule
         \end{tabular}
	 \label{table:correlation_analysis}
    }
\end{table}

	\subsection{Discussion: Insights and Limitations}

    \paragraph{Insights.} Based on the comprehensive evaluation results across ViStoryBench and ViStoryBench-Lite, we derive and briefly outline the following key findings: 
    \begin{enumerate}[label=\textcircled{\scriptsize\arabic*}, leftmargin=1.4em, parsep=-0.05em]
        \item Multi-modal large models (e.g., GPT-4o~\cite{hurst2024gpt4o}) excel in narrative alignment but lag in visual quality; 
        \item Commercial tools exhibit strengths in aesthetics and style sim, yet often lack fine-grained narrative control; 
        \item Story image methods achieve high character consistency but show limited generalization across scenarios; 
        \item Story video methods face challenges in per-frame quality when modeling temporal dynamics; 
        \item Multi-shot video methods exhibit high self-consistency due to training on movie data, while showing weaker customization capability; 
        \item A clear trade-off exists between consistency and diversity, underscoring the need for balanced evaluation; 
        \item This Real-world oriented evaluation reveals that current methods still struggle to jointly optimize semantic and visual qualities. 
    \end{enumerate}
    These insights emphasize the value of task-specific model selection and highlight future directions such as multi-shot data, unified multi-modal architectures, and more holistic evaluation metrics. Details are provided in \textbf{Appendix}~\ref{sec:insights}

    \paragraph{Limitations.} 

    The design and scope of ViStoryBench are subject to several limitations:
    \begin{enumerate}[label=\textcircled{\scriptsize\arabic*}, leftmargin=1.4em, parsep=-0.05em]
    \item Focus on multi-image consistency: Although synchronized audio-visual storytelling is the long-term goal, current evaluation is limited to multi-image generation with emphasis on inter-frame consistency.  
    \item Lack of background-aware evaluation: Due to the absence of background-reference support in existing open-source methods, the benchmark does not include background reference images or scene-level image-and-image similarity evaluation.  
    \item Trade-offs in evaluation metrics: Our hybrid evaluation strategy employs expert models for stability and VLM-based scoring for semantic richness. However, expert models may underperform in complex contexts, while VLMs remain prone to hallucinations.  
    \end{enumerate}

    Despite our extensive efforts to select the most reliable evaluation method for each metric, these limitations remain.

    \section{Conclusion}
    
    We present ViStoryBench, a high-fidelity benchmark for evaluating story visualization under diverse scenarios. It contains 80 multi-shot stories across 10 visual styles, each with character references and script annotations. The benchmark introduces 12 human-validated automated metrics assessing character similarity, style similarity, prompt alignment, aesthetics, artifacts, and copy-paste behavior. Extensive experiments on open-source and commercial systems offer the first large-scale, multi-dimensional comparison of SOTA methods, revealing key strengths, limitations, and future directions.

    Due to space limitations, we have placed a large number of technical details in the appendix and supplementary materials, such as the video display on our homepage, the evaluation of multi-shot video generation models, and the prompt alignment evaluation based on Qwen.

\section*{Acknowledgement}
This work was supported by the National Natural Science Foundation of China (No. 6250070674) and the Zhejiang Leading Innovative and Entrepreneur Team Introduction Program (2024R01007). 
\definecolor{mygray}{gray}{.9}
\definecolor{best}{rgb}{1, 0.7, 0.8}
\definecolor{best2}{rgb}{1, 0.8, 0.9}


\definecolor{colorfirst}{RGB}{252,151,100}
\definecolor{colorsecond}{RGB}{253,187,132}
\definecolor{colorthird}{RGB}{253,212,158}
\definecolor{colorfourth}{RGB}{254,232,200}
\definecolor{colorfifth}{RGB}{255,247,236}

\DeclareRobustCommand{\legendsquare}[1]{%
  \textcolor{#1}{\rule{2ex}{2ex}}%
}
\DeclareRobustCommand{\legendsquarebox}[1]{%
  \tikz[] \draw[black, fill=#1, line width=0.4pt] (0,0) rectangle (1.5ex,1.5ex);%
}
\setlength{\fboxsep}{0pt} 
\definecolor{datasetback}{RGB}{254,232,207}
\definecolor{datasettext}{RGB}{191,144,1}
\definecolor{castback}{RGB}{255,255,229}
\definecolor{casttext}{RGB}{65,171,93}
\definecolor{statsback}{RGB}{255,245,240}
\definecolor{statstext}{RGB}{239,101,78}

\definecolor{PAback}{RGB}{255,241,217}
\definecolor{PAtext}{RGB}{191,144,1}
\definecolor{CPback}{RGB}{255,255,217}
\definecolor{CPtext}{RGB}{192,79,21}
\definecolor{OSback}{RGB}{255,240,217}
\definecolor{OStext}{RGB}{255,153,0}
\definecolor{SIMback}{RGB}{253,221,215}
\definecolor{SIMtext}{RGB}{250,81,50}

\newpage
\appendix
\setcounter{equation}{0}
\setcounter{figure}{0}
\setcounter{table}{0}
\setcounter{section}{0}
\makeatletter
\renewcommand{\theequation}{S\arabic{equation}}
\renewcommand{\thefigure}{S\arabic{figure}}
\renewcommand{\thetable}{S\arabic{table}}


\noindent
\textbf{\LARGE Appendix}
\thispagestyle{empty}

\begin{figure*}[htbp]
    \centering
    \includegraphics[width=\linewidth]{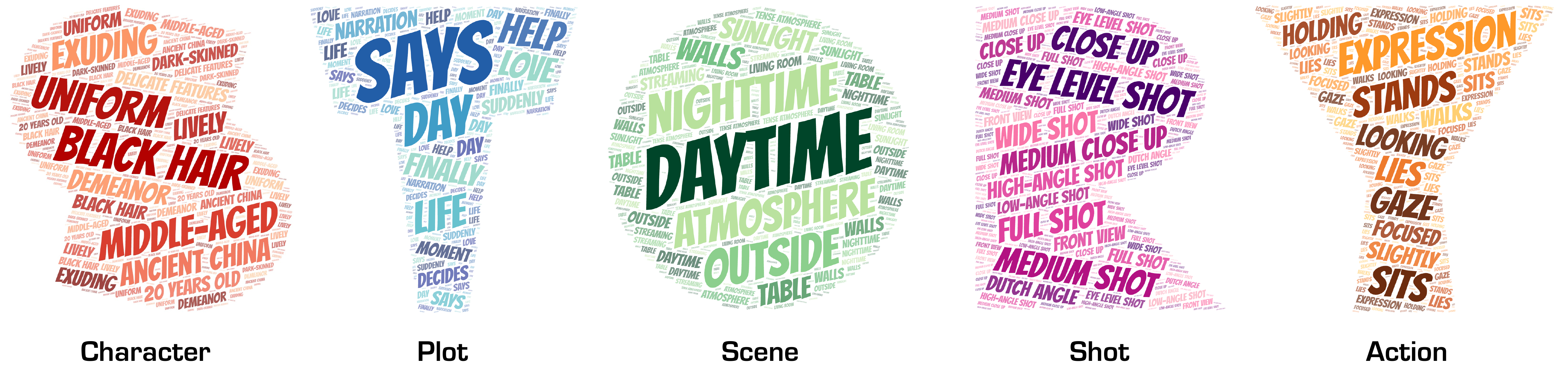}
    \caption{\small \textbf{Words Cloud}. Visualization of narrative elements from the stories in our ViStoryBench dataset: spanning character traits (e.g., black hair, middle-aged, uniform), plot points (e.g., says, finally, love), scene settings (e.g., daytime, atmosphere, outside, sunlight, living room), shot types (e.g., eye-level, close-up, high-angle), and character actions (e.g., expression, stands, gaze).}

    \label{fig:words_cloud}
\end{figure*}

\section{Broader Limitations and Societal Impact}

The design and scope of ViStoryBench are subject to several limitations, which reflect current technological constraints and evaluation challenges in the field of story-oriented generation:

\paragraph{Focus on multi-image consistency:} While the long-term objective of this research community is synchronized audio-visual storytelling with full scene dynamics, the current benchmark is deliberately scoped to multi-image generation with an emphasis on inter-shot consistency. This focus allows us to address the most immediate challenges in visual narrative coherence without introducing additional complexity from temporal modeling or audio alignment, which are active areas of research in their own right. Since established benchmarks like VBench~\cite{zheng2025vbench} already provide comprehensive metrics for single-shot video temporal modeling, we avoid redundant efforts in this area.

\paragraph{Lack of background-reference evaluation:} 
Due to limited support for background-conditioned generation in current open-source story visualization models, ViStoryBench does not incorporate background reference images or include scene-level image-to-image similarity evaluation, with only a minimal number of single-scene stories included. Instead, scene descriptions are provided via text prompts, and scene consistency is assessed through prompt alignment. Future work will expand the dataset to include more single-scene multi-shot stories along with corresponding scene reference images to enhance the scope and functionality of the benchmark.

\paragraph{Inherent trade-offs in evaluation metrics:} The benchmark employs a hybrid evaluation strategy that combines expert models (for stability and continuity) and VLM-based scorers (for semantic richness and narrative alignment). However, this approach involves fundamental compromises: expert models may lack adaptability in visually or narratively complex scenarios, while VLMs remain susceptible to hallucination and may not always align with human perceptual judgments. These trade-offs underline the lack of a universally optimal automated metric for all aspects of story visualization.

\paragraph{Dataset Limitations and Copyright Concerns.}
Some images in our dataset are derived from well-known movies, TV shows, and animations. While these samples are used solely for academic research and benchmarking purposes, they may raise copyright concerns. We do not claim ownership of any copyrighted material, and all third-party content is included under fair use principles for non-commercial research and analysis. Nevertheless, users of our dataset should be aware of potential legal constraints when repurposing or redistributing the data.
Additionally, the inclusion of well-known visual content may cause certain metrics to become overfitted to these familiar styles or characters, potentially leading to metric manipulation or over-optimization.

\paragraph{Language Sensitivity.}
Our benchmark supports both Chinese and English story prompts. While we select the appropriate language for each model based on its design and documentation, we do not control for potential discrepancies in generation quality caused by language differences. This may introduce variation that is not attributable to model capability alone.

\paragraph{Scope of Evaluation.}
Our benchmark currently does not support accurate evaluation of comic-style or manga generation tasks that involve multi-panel layouts within a single image, due to the lack of a robust panel segmentation method. Similarly, we do not assess inference efficiency or runtime performance across models.
For video-based story generation methods, our benchmark does not yet include a comprehensive evaluation of temporal coherence, frame-level consistency, or other video-specific quality metrics. These remain important future directions.

\paragraph{Societal Impact.}
We envision story visualization models as promising tools for education, creativity, and cultural preservation. In curating the dataset, we made conscious efforts to include diverse narratives from multiple cultures and regions. However, generative models are still susceptible to reproducing stereotypes and amplifying data-driven biases. It is vital that these tools are developed and deployed responsibly.

\paragraph{Conclusion.}
Despite extensive efforts to align metric selection with the requirements of each evaluation dimension, these limitations reflect persistent challenges in the current evaluation paradigm. They also indicate meaningful directions for future work in developing more comprehensive and reliable story visualization benchmarks.

Finally, we emphasize that generative models should not be used to create or disseminate false or misleading content. Addressing such risks requires active collaboration between researchers, platform providers, and policymakers to ensure safe and ethical applications.

\section{Overall Insights}\label{sec:insights}

Based on the table of automated test results on ViStoryBench and ViStoryBench-Lite, we have analyzed the performance of various story visualization methods across multiple metrics. Here are the key insights and patterns observed from a research perspective:

\paragraph{Performance of Multi-modal Large Models is Dominant, Especially GPT-4o.}
\begin{enumerate}[label=\textcircled{\scriptsize\arabic*}, leftmargin=1.4em, parsep=-0.05em]
    \item GPT-4o~\cite{hurst2024gpt4o} achieves the best or second-best performance in critical metrics like Alignment Score (3.673), OCCM Score (93.5), CIDS (Cross: 0.571 and Self: 0.679), and CSD (Cross: 0.481 and Self: 0.680), indicating strong narrative understanding, character consistency and style similarity. This suggests that large-scale multi-modal models excel at high-level semantic alignment and coherence, likely due to their extensive pre-training on diverse data.
    \item However, GPT-4o~\cite{hurst2024gpt4o} lags in Inception Score (9.02) and Aesthetics Score (5.49), which measure image diversity and visual quality. This implies a trade-off: while LLMs handle narrative complexity well, they may struggle with low-level visual fidelity compared to specialized methods.
\end{enumerate}

\paragraph{Commercial Software Shows Strengths in Visual Quality but Inconsistencies in Narrative Tasks.}
\begin{enumerate}[label=\textcircled{\scriptsize\arabic*}, leftmargin=1.4em, parsep=-0.05em]
   \item Commercial tools like MorphicStudio~\cite{morphic2024studio} excel in style consistency (CSD\_Cross: 0.653), while Doubao~\cite{doubao2024} perform well in alignment (3.494). This highlights their optimization for production-ready output.

   \item However, they exhibit variability: for example, MOKI~\cite{moki2024} has high aesthetics (5.79) but poor character consistency (CIDS\_Cross: 0.214). This indicates that commercial tools may prioritize aesthetic appeal over fine-grained narrative controls, leading to imbalances in evaluation dimensions.
\end{enumerate}

\paragraph{Story Image Methods Excel in Specific Niches, but Lack Uniformity.}
\begin{enumerate}[label=\textcircled{\scriptsize\arabic*}, leftmargin=1.4em, parsep=-0.05em]
   \item Methods like OmniGen2~\cite{wu2025omnigen2} lead in character consistency (CIDS\_Self: 0.537) and OCCM (90.8), demonstrating strengths in maintaining identity across frames. Story-Adapter~\cite{mao2024story_adapter} variants achieve high style consistency (CSD\_Cross: 0.456), showing progress in specialized tasks.

   \item However, performance varies widely: SEED-Story~\cite{yang2024seedstory} and TheaterGen~\cite{cheng2024theatergen} score low on multiple metrics (e.g., Alignment $\textless2.00$ and CIDS\_Cross $\textless0.35$), indicating that some methods overfit to specific scenarios or lack generalization. The reliance on reference images (e.g., Story-Adapter~\cite{mao2024story_adapter} with image-ref) often boosts consistency but may limit creativity.
\end{enumerate}

\paragraph{Story Video Methods Underperform in Key Metrics, Revealing Challenges in Temporal Modeling.}
\begin{enumerate}[label=\textcircled{\scriptsize\arabic*}, leftmargin=1.4em, parsep=-0.05em]
   \item Video-based Methods like Vlogger~\cite{zhuang2024vlogger} and MovieAgent~\cite{wu2025movieagent} generally score lower in style and character consistency (e.g., CSD\_Cross $\textless0.3$) compared to image-based methods. This suggests that temporal modeling introduces additional complexity, hindering per-frame quality.

   \item An exception is MovieAgent~\cite{wu2025movieagent} (SD3), which achieves strong alignment (3.16), implying that leveraging advanced image-based diffusion models (e.g., SD3~\cite{esser2024sd3}) can mitigate some issues. Yet, overall, video methods lag in metrics like Inception Score, indicating limited diversity in generated sequences.
\end{enumerate}

\paragraph{Excellent Performance in Multi-shot Video Models, Especially Sora2.}
\begin{enumerate}[label=\textcircled{\scriptsize\arabic*}, leftmargin=1.4em, parsep=-0.05em]
    \item Multi-shot video generation models excel in cross-shot character and scene consistency, as evidenced by high self-similarity score (e.g., CIDS\_Self 0.813 and CSD\_Self 0.713 for Sora2~\cite{2025sora2}), outperforming many image-based methos. This likely results from training on large-scale character-consistent multi-shot movie data. In contrast, most image-based methods lack specialized training on multi-shot storyboard data.

    \item However, current multi-shot video models struggle with visual reference adherence, indicated by relatively lower cross-similarity scores (e.g., CIDS\_Cross 0.738 and CSD\_Cross 0.515 for Sora2~\cite{2025sora2}), revealing a gap in style and scene constraint preservation compared to reference-based image methods. 
\end{enumerate}

\paragraph{Trade-offs Between Consistency and Creativity Are Evident.}
Methods with high character consistency and high copy-paste score (lower is better) often have lower diversity (e.g., GPT-4o). This underscores a fundamental tension in story visualization: optimizing for one dimension may compromise another.

\paragraph{In story visualization tasks, comprehensive evaluation metrics are extremely important.} For instance, the simple Copy-Paste Baseline achieves optimal results across numerous metrics. However, its alignment score is notably low. Although IS can generally measure the quality and diversity of image generation, it is quite challenging to compare different models by examining the IS metric alone.  When using only text as input, StoryDiffusion~\cite{zhou2024storydiffusion} and Story-Adapter~\cite{mao2024story_adapter} achieve excellent IS and aesthetic quality. However, relying solely on text input clearly cannot produce results that resemble the features and styles of the character reference images. 

\paragraph{Our quantitative metrics demonstrate alignment with qualitative observations.} For Story-Adapter~\cite{mao2024story_adapter}, the scoring consistency between automated metrics and human evaluation is particularly evident: (1) In text-only mode (its native setting), the overall quality score (scale=5) systematically surpasses the baseline (scale=0), as theoretically expected; (2) When using image references, scale=0 achieves higher cross-similarity but lower self-similarity compared to scale=5 in both CIDS and CSD.

\paragraph{ViStoryBench-Lite Reveals Real-World Gaps}
Results on ViStoryBench-Lite (focused on practical scenarios) show that commercial and LLM methods perform well in alignment but struggle with low-level metrics (e.g., Gemini-2.0~\cite{gemini2.02025} has Align: 3.150 but CSD\_Cross: 0.361). This indicates that real-world applications require balancing semantic and visual qualities, and current methods may not fully address this.

\paragraph{Conclusion and Future Directions:}
\begin{enumerate}[label=\textcircled{\scriptsize\arabic*}, leftmargin=1.4em, parsep=-0.05em]
    \item No single method dominates all metrics, emphasizing the need for task-specific model selection. For narrative-heavy tasks, LLMs like GPT-4o~\cite{hurst2024gpt4o} are preferable; for visual quality, commercial tools or specialized image methods may suffice.
    
    \item Future work should focus on hybrid approaches: integrating LLMs for planning with diffusion models for visual quality, and improving evaluation metrics to better capture storytelling aspects like pacing and emotion.

    \item A promising direction for story visualization is to combine the temporal coherence of video models with the precise reference alignment of image-based generators, enabling customized multi-shot output with both consistency and control.
    
    \item The variability in results underscores the value of benchmarks like ViStoryBench for guiding progress. Researchers should prioritize methods that balance consistency, diversity, and alignment.
\end{enumerate}

\begin{figure*}
    \centering
    \includegraphics[width=\linewidth]{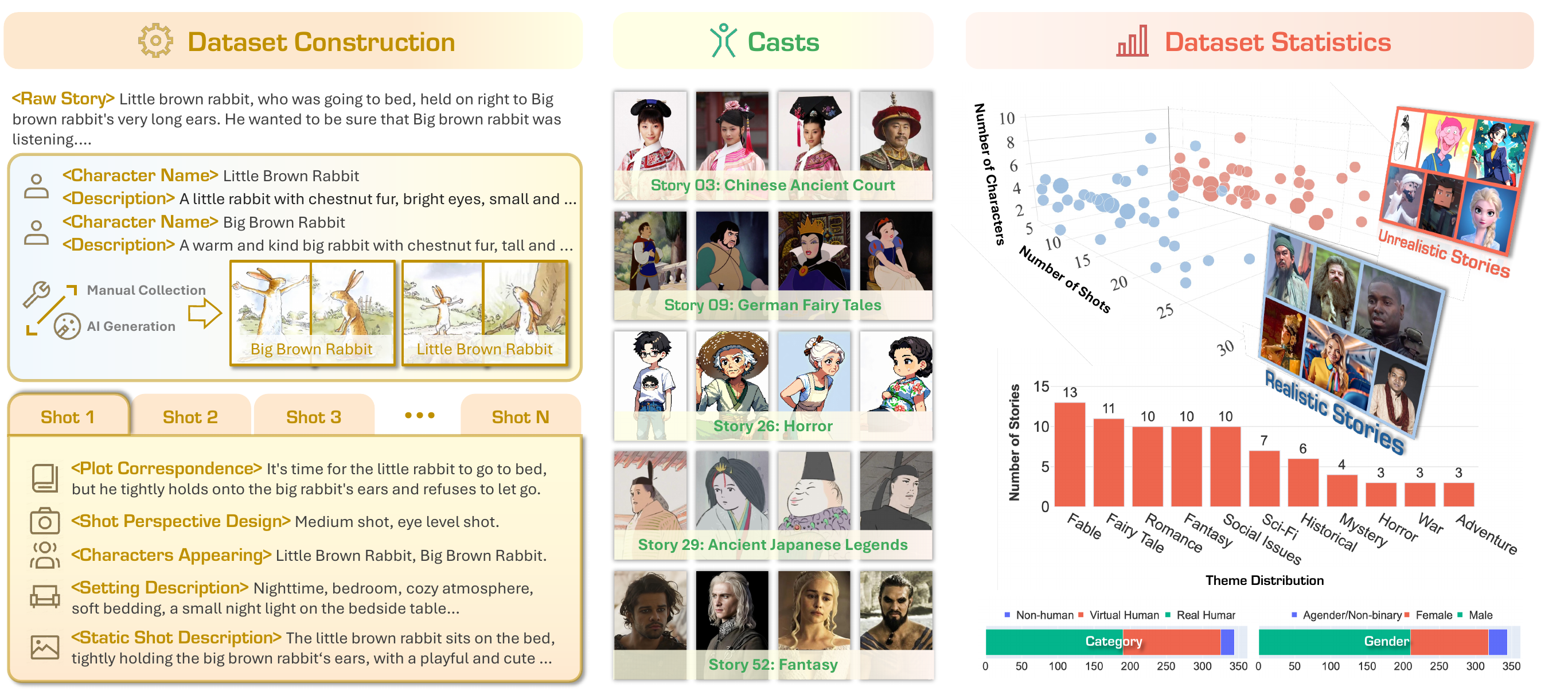}
    \caption{
    \small \textbf{Overview of ViStoryBench Dataset.} 
\highlighttext{datasettext}{datasetback}{\small \textbf{Dataset Construction}}: We build a story generation pipeline powered by large language models (LLMs), followed by human verification to ensure quality and consistency. 
\highlighttext{casttext}{castback}{\small \textbf{Casts}}: Reference images for characters are manually curated to maintain a consistent visual style. 
\highlighttext{statstext}{statsback}{\small \textbf{Dataset Statistics}}: 
ViStoryBench dataset exhibits a broad distribution across story categories, stylistic variations, and character diversity, enabling comprehensive evaluation of storytelling generation models.
}
    \label{fig:main}
    \vspace{-1em}
\end{figure*}

\section{Details of Dataset Collection and Statistics}\label{sec:dataset}

Story visualization datasets have shown notable growth and evolution in terms of scale, image resolution, automated generation pipelines, and stylistic diversity—reflecting both technological advancements and the expanding scope of research interests. We summarize story visualization datasets~\cite{huang2016visual,gupta2018imagine,li2019storygan,yang2024seedstory,ye2024openstory,Liu_2024_CVPR_Storygen} in Table~\ref{tab:visual_story_datasets}. A key distinction of ViStoryBench compared to existing datasets lies in our construction methodology: rather than extracting captions from visual keyframes to construct a narrative, we take a top-down approach by generating structured shot descriptions directly from full textual stories. 
Additionally, when curating benchmarks, we place particular emphasis on ensuring a broad range of styles and thematic diversity. Table~\ref{tab:style_distribution} presents a detailed breakdown of the 10 visual styles curated in our dataset, which are classified based on their character reference images.

\begin{table}[htbp]
\centering
\caption{Distribution of the 10 Visual Styles in ViStoryBench. The classification is based on the visual style of the character reference images for each story.}
\label{tab:style_distribution}
\resizebox{\columnwidth}{!}{%
\begin{tabular}{@{}lc@{}}
\toprule
\textbf{Style Category} & \textbf{Number of Stories} \\
\midrule
Photorealistic / Live-action photo & 39 \\
\rowcolor{teaserdatasetback}Anime / Cel-shading style & 14 \\
Children's book / Cartoon & 7 \\
\rowcolor{teaserdatasetback}Classic fairy tale illustration / Vintage & 5 \\
Classical oil painting / Religious art & 4 \\
\rowcolor{teaserdatasetback}Flat vector illustration & 4 \\
3D / Voxel / Claymation style & 3 \\
\rowcolor{teaserdatasetback}Chinese ink painting / Line art / Silhouette & 2 \\
Action manga / American comics style & 1 \\
\rowcolor{teaserdatasetback}CG realistic fantasy illustration & 1 \\
\midrule
\textbf{Total} & \textbf{80} \\
\bottomrule
\end{tabular}%
}
\end{table}

\begin{figure*}[ht!]
    \centering
    \includegraphics[width=0.92\linewidth]{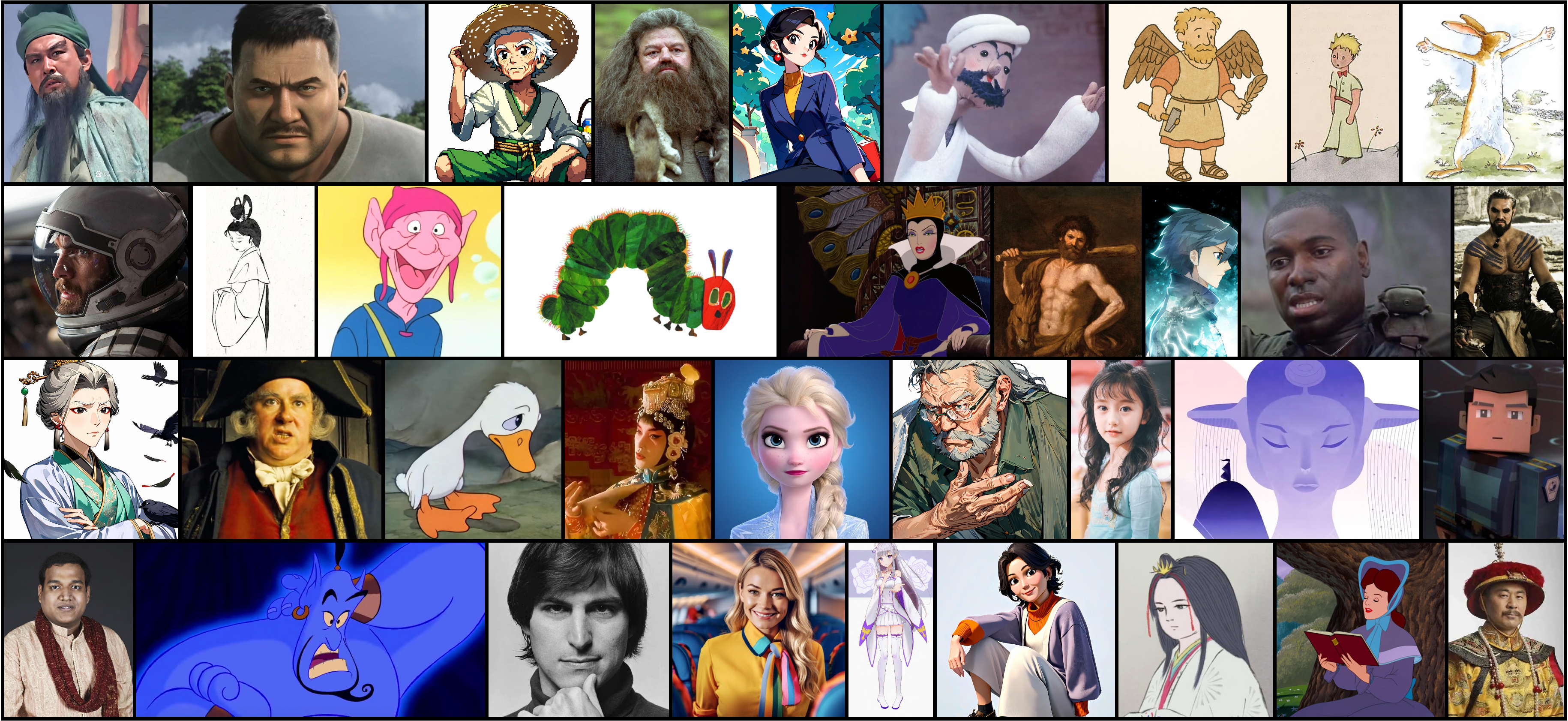}
\caption{\small \textbf{Random Character Reference Samples from the Dataset}. The reference images include full-body shots, half-body shots, or portraits, spanning diverse visual styles from photorealistic to a variety of animation-inspired designs.}
    \label{fig:characterstyle}
\end{figure*}

The collected dataset spans a wide array of genres, including 13 folktales, 10 romance stories, 4 suspense/crime stories, 3 horror narratives, 6 historical tales, 10 fantasy stories, 7 science fiction stories, 3 war stories, 10 stories about social life, 3 survival/adventure stories, and 11 fairy tales. These stories are segmented and adapted into detailed shot scripts using the in-house LLM model.
The full dataset contains 1,317 shots in total, with each story comprising between 4 and 30 shots (averaging 16.5 shots per story). Basic statistics shown in Figure~\ref{fig:main}

To support a wide range of methods, all test-related textual prompts are provided in both English and Chinese. For methods that only support Chinese, or perform significantly better with Chinese input, we use the Chinese version; otherwise, English inputs are used. Each individual shot is annotated with five structured fields: \textit{Setting Description}, \textit{Plot Correspondence}, \textit{Onstage Characters}, \textit{Static Shot Description}, and \textit{Shot Perspective Design}.

In our curation process, we made a conscious effort to incorporate narratives from diverse cultural backgrounds. The cultural origins of the 80 stories are distributed across several major regions, including Chinese (39 stories), Euro-American (27), Japanese (8), African (3), Islamic/Middle Eastern (2), and Indian (1). This diversity is complemented by a wide array of thematic genres: the dataset includes 13 folktales, 10 romance stories, 4 suspense/crime stories, 3 horror narratives, 6 historical tales, 10 fantasy stories, 7 science fiction stories, 3 war stories, 10 stories about social life, 3 survival/adventure stories, and 11 fairy tales.

\subsection{Character Reference Image}

For most well-known stories in our dataset, character reference images are directly sourced from existing visual works such as movies, animated films, or television series. For lesser-known or original stories, we adopt one of two strategies to obtain reference images for the main characters: (1) retrieving representative screenshots from films or TV shows with similar settings and styles (covering 16 stories), or (2) using the SDXL model~\cite{podell2023sdxl} to generate high-quality stylized animation portraits (covering 7 stories).

In total, our dataset includes \textbf{344 unique characters}, which can be categorized into \textbf{190 real humans}, \textbf{135 virtual humans} (e.g., animated or game characters), and \textbf{19 non-human entities} (e.g., animals or creatures). Regarding gender annotation, there are \textbf{210 male}, \textbf{108 female}, and \textbf{26 characters} who are either genderless or non-binary. Each character is associated with between 1 and 10 reference images, resulting in a total of \textbf{509 reference images}, with 89 characters having more than one image. A selection of these reference images is visualized in Figure~\ref{fig:characterstyle}.

Furthermore, we categorize all 80 stories into two distinct types based on the visual style of the main character references: \textbf{realistic stories} and \textbf{unrealistic stories}. The realistic category includes 39 stories whose characters are portrayed using photographic or cinematic images. 
These characters are additionally labeled with ethnicity information following prior works~\cite{cheng2023dna,cheng2022generalizable,pan2023renderme}. 
The remaining 41 stories are labeled as unrealistic, typically involving animation, stylized art, or fantasy characters.

This classification allows us to conduct stratified evaluations and analyze how different generation methods perform across story types with distinct visual and semantic characteristics.

\begin{table*}[ht]
\centering
\caption{\small \textbf{Comparison between Lite and Full Settings across Multiple Evaluation Dimensions}. Quantitative comparison between the full dataset and the lite subset under a single method. The close alignment across all metrics, including style and character consistency, generative quality, diversity, and prompt alignment, which demonstrates that the lite subset serves as a representative proxy for the full dataset, enabling efficient yet reliable evaluation.}
\label{tab:lite_full}
\resizebox{\linewidth}{!}{
\begin{tabular}{l|cc|cc|c|cc|cccc}
\toprule
\textbf{} & \multicolumn{2}{c|}{\textbf{Style Consistency}} & \multicolumn{2}{c|}{\textbf{Character Consistency}} & \textbf{Character Matching} & \textbf{Generative Quality} & \textbf{Diversity} & \multicolumn{4}{c}{\textbf{Prompt Alignment}} \\
& Cross & Self & Cross & Self & OCCM & Aesthetic & Inception & Scene & Shot & CI & IA \\
\midrule
\textbf{Full} & 0.325 & 0.569 & 0.427 & 0.600 & 62.487 & 4.894 & 11.510 & 2.331 & 2.891 & 2.176 & 2.101 \\
\textbf{Lite} & 0.373 & 0.626 & 0.469 & 0.620 & 65.351 & 5.027 & 9.787 & 2.779 & 2.965 & 2.564 & 2.290 \\
\midrule
\rowcolor{teaserdatasetback}\textbf{Diff} & 13.78\% & 9.55\% & 9.34\% & 3.21\% & 4.48\% & 2.68\% & 16.17\% & 17.57\% & 2.52\% & 16.34\% & 8.62\% \\
\bottomrule
\end{tabular}
}
\end{table*}

\begin{table*}[ht!]
    \centering
\caption{\small \textbf{Comparison of Multi-modal Storytelling Datasets}. Representative story datasets are summarized in terms of annotation method, image scale, resolution, number of shots per story, and visual style.}
    \resizebox{0.65\textwidth}{!}{\begin{tabular}{l|cccccc}
            \toprule
            \textbf{Datasets}&\textbf{Caption} & \textbf{\# Images} & \textbf{Resolution} & \textbf{\# Shots} & \textbf{Style}\\ \midrule
            VIST~\cite{huang2016visual} & Manual & $146$k & - & 5 & Realistic \\
            \rowcolor{teaserdatasetback}Flintstones~\cite{gupta2018imagine} & Manual & $123$k & $128\times128$ & 5 &Anime  \\
            Pororo~\cite{li2019storygan} & Manual & $74$k & $128\times128$ & 5  &Anime \\ 
            \rowcolor{teaserdatasetback}StorySalon~\cite{Liu_2024_CVPR_Storygen} &ASR & $160$k & $432\times803$ & 14 &Anime \\ 
            StoryStream~\cite{yang2024seedstory} &Generated & $258$k & $480\times854$ & 30 &Anime \\
            \rowcolor{teaserdatasetback}OpenStory~\cite{ye2024openstory} & Generated & $107$M & $720$p \& $1080$p & 28 & Realistic \\
            \bottomrule
        \end{tabular}
    }
    \label{tab:visual_story_datasets}
\end{table*}

\begin{table*}[htpb]
\centering
\begin{tcolorbox}[title=Prompt for Dataset Construction (Part 1),
enhanced, 
skin first=enhanced,
skin middle=enhanced,
skin last=enhanced,
colback=yellow!10!white,
colframe=orange!80!black,
fonttitle=\bfseries]{}
You are a seasoned film script artist skilled at transforming descriptive text from novel scripts into visual content descriptions. You also adapt scripts into static shot scripts. Your designs must incorporate a wide variety of compositions to perfectly capture the script's content through imagery, ensuring the storyline is effectively conveyed in the visual descriptions. In addition, you also need to provide a comprehensive introduction to the characters that appear in the shot scripts, mainly describing their appearance and clothing.\\

\textbf{Task Description}

You are required to write shot scripts based on the user's input story, totaling \textcolor{blue}{\textless num\_of\_shots\textgreater} shots. Each shot can feature 0 to 3 characters, meaning it can be a scene shot, single-character shot, two-character shot, or three-character shot. For each shot in the shot script, you need to output \textless Plot Correspondence\textgreater, \textless Setting Description\textgreater, \textless Shot Perspective Design\textgreater, \textless Onstage characters\textgreater, \textless Static Shot Description\textgreater. 

The \textless Plot Correspondence\textgreater section requires dividing the original plot into \textcolor{blue}{\textless num\_of\_shots\textgreater} scenes, presenting the plot of each scene in the form of narration and dialogue. Note that when dividing the input plot into different scenes, the rationality of the plot needs to be considered.\\

\textbf{Concept Explanation of Various Fields in Shot Scripts}
\begin{itemize}[itemsep=-1.4pt,leftmargin=2em]
\item Setting Description: The story will be divided into \textcolor{blue}{\textless num\_of\_shots\textgreater} scenes. The setting refers to the environmental setup of each scene. It should not include any characters. You need to describe all elements in the environment in detail in this field, so that the scene where the story takes place can be vividly recreated. Standard writing format: time, location, atmosphere description, other elements in the environment, lighting effects.
\item Shot Perspective Design: Shot Perspective Design refers to information from several dimensions: shot distance, camera angle, and camera type.
\item Onstage characters: Please select the characters appearing in this scene from the character list. The number of characters should be controlled between 0-3. If no characters appear, leave it blank.
\item Static Shot Description: This part describes the static actions or positions of characters and items in the scene, ensuring that it describes a fixed state. Writing format: \textless character position\textgreater, \textless character expression\textgreater, \textless character action\textgreater, \textless position of elements in the scene\textgreater.
\end{itemize}

\textbf{Requirements for Creating Setting Description in Shot Scripts}
\begin{itemize}[itemsep=-1.4pt,leftmargin=2em]
\item You need to design a shot script for the user's input story. Break the story into \textcolor{blue}{\textless num\_of\_shots\textgreater} main scenes and write scene descriptions for each of these \textcolor{blue}{\textless num\_of\_shots\textgreater} scenes.
\item Note that character descriptions should not appear in Setting Descriptions. Only describe the scene itself, ensuring that the scene is consistent with the original story. Do not include backgrounds, items, or other elements that are not present in the story.
\item Think from the perspective of the visuals, using the visuals to drive the content of the shots. Ensure that all plot elements can be directly depicted through the visuals. Avoid thinking from the perspective of a screenwriter's script and refrain from using abstract or metaphorical expressions.
\item Pay attention to the consistency between the characters' locations and the Setting Descriptions.
\item When writing the background content, do not directly use the expressions from the origin story. Use clear and concise sentences to describe in detail all the elements included in the background visuals and their relationships.
\item If there are characters in the visuals, clearly express their facial expressions, demeanor, and actions.
\item Pay attention to the visual narrative continuity between adjacent shot panels.
\end{itemize}

\textbf{Requirements for Creating Shot Perspective Design in Shot Scripts}
\begin{itemize}[itemsep=-1.4pt,leftmargin=2em]
    \item Continuity in Shot Composition: Adjacent shots should maintain coherence in shot composition and camera angles. By employing a diverse yet consistent combination of shot compositions and camera angles, create a viewing experience that is both spatially immersive and visually engaging.
    \item Shot Distance Selection: Choose the most appropriate shot distance from "wide shot, full shot, medium long shot, medium shot, medium close up, close up" based on the emotional atmosphere of the current scene. A smaller subject size results in a more relaxed emotional tone, while a larger subject size creates a tenser atmosphere. Consider the shot distance design of preceding and following shots as well.

\end{itemize}

\end{tcolorbox}
\end{table*}

\begin{table*}[htpb]
\centering
\begin{tcolorbox}[title=Prompt for Dataset Construction (Part 2),
enhanced, 
skin first=enhanced,
skin middle=enhanced,
skin last=enhanced,
colback=yellow!10!white,
colframe=orange!80!black,
fonttitle=\bfseries]{}

\begin{itemize}[itemsep=-1.4pt,leftmargin=2em]
    \item Guidelines for Shot Composition Combinations:
    \begin{itemize}[itemsep=-1.4pt,leftmargin=2em]
        \item Smooth Narrative Transitions: The way shot compositions are connected impacts narrative fluidity. Effective transitions involve gradual tightening or loosening. Moving from wider to tighter shots is called "tightening," 
        while moving from tighter to wider shots is called "loosening." Avoid abrupt shifts from Wide Shot to extreme close-ups or vice versa.
        \item Avoid Repetitive Compositions: Ensure that adjacent shots have different compositions to prevent visual monotony.
        \item Emotional Resonance: Since shot composition affects emotional tone, match shot compositions with the emotional intensity of the plot. For instance, intense emotional scenes are better suited for tighter shots.
        \item Rhythmic Pacing: The combination of shot compositions influences the visual rhythm of the scene. Use an appropriate mix of shot compositions within each episode to convey the narrative's pace effectively.
    \end{itemize}
\item Camera Angle Selection: Opt for the camera angle—"front view, side view, back view"—that best suits the current scene's visual requirements.

\item Camera Type Selection: Choose the camera type—"Eye level shot, low-angle shot, high-angle shot, bird's eye view shot, dutch angle shot, foreshortening, inverted shot"—that aligns with the scene's content and emotional tone. Consider the camera types used in preceding and following shots for continuity.

\item Camera Type Combinations: Select camera types that complement the scene's content and emotional context. Pay attention to how different camera types interact to enhance the visual storytelling.

\item Please refer to the provided materials to choose the most fitting camera type for the current scene's content and emotional tone, ensuring that the combination of camera types supports the overall narrative effectively.
\end{itemize}

\textbf{Reference Materials for Camera Design}

For Shot Distance selection:
\begin{itemize}[itemsep=-1.4pt,leftmargin=2em]
\item Wide Shot: Displays the relationship between characters and their environment, commonly used to showcase scenes and background settings.
\item Full shot: Shows the entire body of a character, often used to present full actions or the overall view of a scene.
\item Medium long shot: Captures from above the character's knees.
\item Medium shot: Captures from above the character's waist.
\item Medium close up: Captures from above the character's chest.
\item Close up: focus on a close-up of the character's head or face, with the background and environment typically blurred or entirely out of view.
\end{itemize}
For determining the relationship between Camera Type and content:
\begin{itemize}[itemsep=-1.4pt,leftmargin=2em]
\item Eye level shot: The camera is positioned at the same level as the eyeline.
\item Low-angle shot: The camera shoots upward from below the eyeline, enhancing the subject's authority or size, often conveying power or intimidation. Suitable for emphasizing an individual's dominance or creating visual pressure, such as highlighting a hero or villain.
\item High-angle shot: The camera looks down from above the eyeline, showing the breadth of a scene or diminishing the visual importance of the subject. Effectively reduces the visual scale of characters or objects, used to depict a sense of isolation or helplessness in characters, or to present vast landscapes.
\item Bird's eye view shot: The camera shoots downward from a high altitude, providing a top-down perspective, usually covering extensive geographical areas. Highly effective when a global view of events or environments is needed.
\item Dutch angle shot: The camera is deliberately tilted during filming, often used to create a sense of imbalance or tension. Particularly effective in portraying scenes of chaos, tension, or psychological instability.
\item Foreshortening: Emphasizes depth of field through perspective techniques, making the relationship between foreground and background more prominent. Suitable for highlighting spatial relationships and depth, commonly used to enhance visual guidance and a sense of depth.
\item Inverted shot: The image is filmed upside down, challenging the audience's visual habits, often used to represent confusion or an unstable mental state.
\end{itemize}
\textbf{Notes}
\begin{itemize}[itemsep=-1.4pt,leftmargin=2em]
\item Each Static Shot Description in the shot scripts should correspond sequentially to each segment of the story, providing detailed descriptions of characters' expressions, actions, and states.

\end{itemize}
\end{tcolorbox}
\end{table*}

\begin{table*}[htpb]
\centering
\begin{tcolorbox}[title=Prompt for Dataset Construction (Part 3),
enhanced, 
skin first=enhanced,
skin middle=enhanced,
skin last=enhanced,
colback=yellow!10!white,
colframe=orange!80!black,
fonttitle=\bfseries]{}
\textbf{Notes}
\begin{itemize}[itemsep=-1.4pt,leftmargin=2em]
\item Do not include any characters in the Setting Description. Ignore the characters in the story and only describe the environmental setting.
\item Do not introduce items or backgrounds in the Static Shot Description that are not mentioned in the story. Ensure that the location of actions aligns with the story.
\item Each sentence in the plot has context. Use this context to determine the best shot design.
\item Ensure that the transitions between different  shots in the shot scripts follow creative requirements.
\item Approach the creation of shot content from a visual perspective. Consider the best way to present the script visually.
\item Only provide content that meets the output format requirements; do not include any explanations.
\item When dividing the original plot content, rewrite it into a format suitable for presentation.
\item Strictly follow the format and requirements of the output example when writing.
\end{itemize}

\textbf{A Sample output}

{\small
\begin{lstlisting}[breaklines=true,language=json]
{
  "Shot 1": {
    "Plot Correspondence": "Fern is performing a magical ritual at the bell tower to awaken an ancient power, which is crucial for her team's quest to protect the town from impending danger.",
    "Setting Description": "Twilight, stone street leading to the bell tower, bell tower with arches, town streets, distant bell tower with arches, romantic atmosphere, mysterious atmosphere, glowing arrow on the ground, fallen leaves, pigeons perched on rooftops, soft golden light",
    "Shot Perspective Design": "Medium shot, eye level shot",
    "Onstage characters": ["Fern"],
    "Static Shot Description": "Fern is pressed against the bell tower pillar, with an expectant and serious expression. Her right index and middle fingers are together, suspended in mid-air, and fine golden particles are pouring out from her fingertips."
  },
  "Characters": {
    "Fern": "A beautiful girl with long purple hair and purple eyes, wearing a silver butterfly hair accessory, a black coat and black boots, and a white dress under the coat. She holds a wooden staff in her hand.",
    "Frieren": "A young white-haired female elf with twin ponytails, blue-green eyes, pointed elf ears, a pair of red earrings, a white wizard robe with gold trim and brown boots, holding a staff with a round ruby on the top.",
    "Himmel": "A young boy with short blue hair and blue eyes, wearing a blue knight uniform and a white cape, holding a long sword in his hand, with a gentle yet firm expression."
  }
}
\end{lstlisting}

}
\end{tcolorbox}
\end{table*}

\subsection{Prompts for Dataset Construction}
To automate the transformation of story narratives into detailed multi-shot descriptions, we leverage a large language model (LLM) as a shot planner. This LLM is tasked with segmenting each narrative into a coherent sequence of visual shots, ensuring consistency in character presence, camera composition, environment description, and story progression. 

We propose a structured prompt engineering approach for data generation, which systematically decomposes the complex story visualization task into well-defined, verifiable subtasks to achieve precise control over LLM outputs. This methodology aligns with rigorous benchmark construction standards in academic research and offers a valuable technical framework for building complex, structured multimodal datasets. Our approach converts an MLLM into a controllable visual narrative script generator and employs five core strategies:
\begin{enumerate}[label=\textcircled{\scriptsize\arabic*}, leftmargin=1.4em, parsep=-0.05em]
    \item\textbf{Multi-grained task decomposition}, breaking the task into five structured modules—Plot Correspondence, Setting Description, Shot Perspective Design, Onstage Characters, and Static Shot Description—enabling the LLM to focus on simpler subproblems for improved accuracy and stability.
    \item\textbf{Professional knowledge infusion}, incorporating cinematic expertise such as standardized shot terminology (e.g., Wide Shot, Low-angle) and narrative principles (e.g., Smooth Narrative Transitions, Emotional Resonance).
    \item\textbf{Multi-dimensional information isolation}, enforcing modality separation (e.g., excluding characters from setting descriptions) to prevent spurious correlations and support combinatory generalization.
    \item\textbf{Visual-friendly description}, ensuring all textual content is concrete and directly depictable (e.g., avoiding abstract expressions in favor of visually grounded descriptions).
    \item\textbf{Contextual narrative modeling}, maintaining coherence across shot sequences by considering preceding and subsequent shots.
    Thus, our prompt set defines an LLM-powered automated data generation framework, enabling efficient construction of a high-quality, consistent, and domain-informed benchmark for visual storytelling.
\end{enumerate}
Below, we present the specific system prompt utilized for this purpose.

\section{Qualitative Results}\label{sec:qualitative}
\begin{figure*}[htpb]
    \centering
    \includegraphics[width=0.9\linewidth]{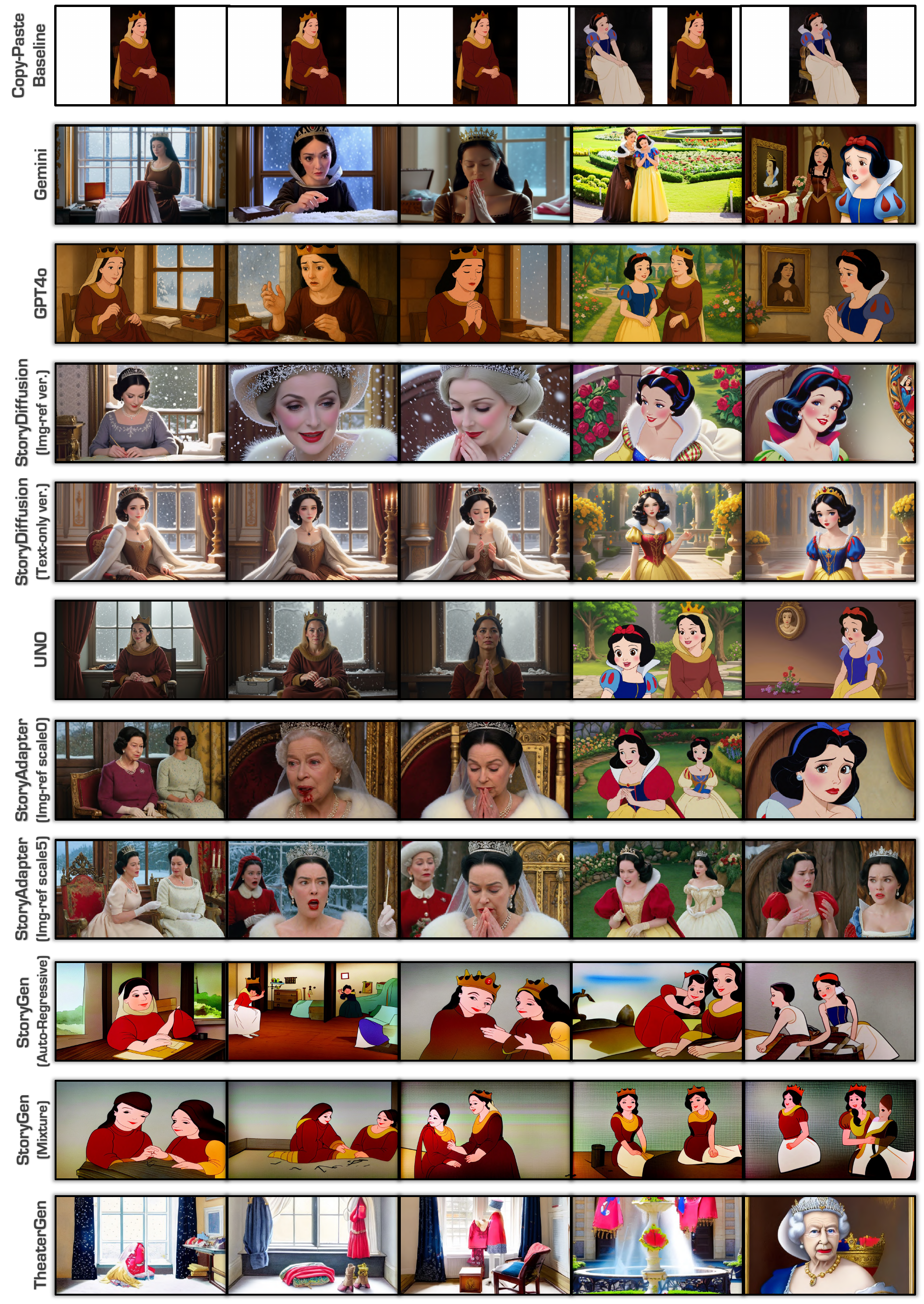}
\end{figure*}
\begin{figure*}[htpb]
    \centering
    \includegraphics[width=0.95\linewidth]{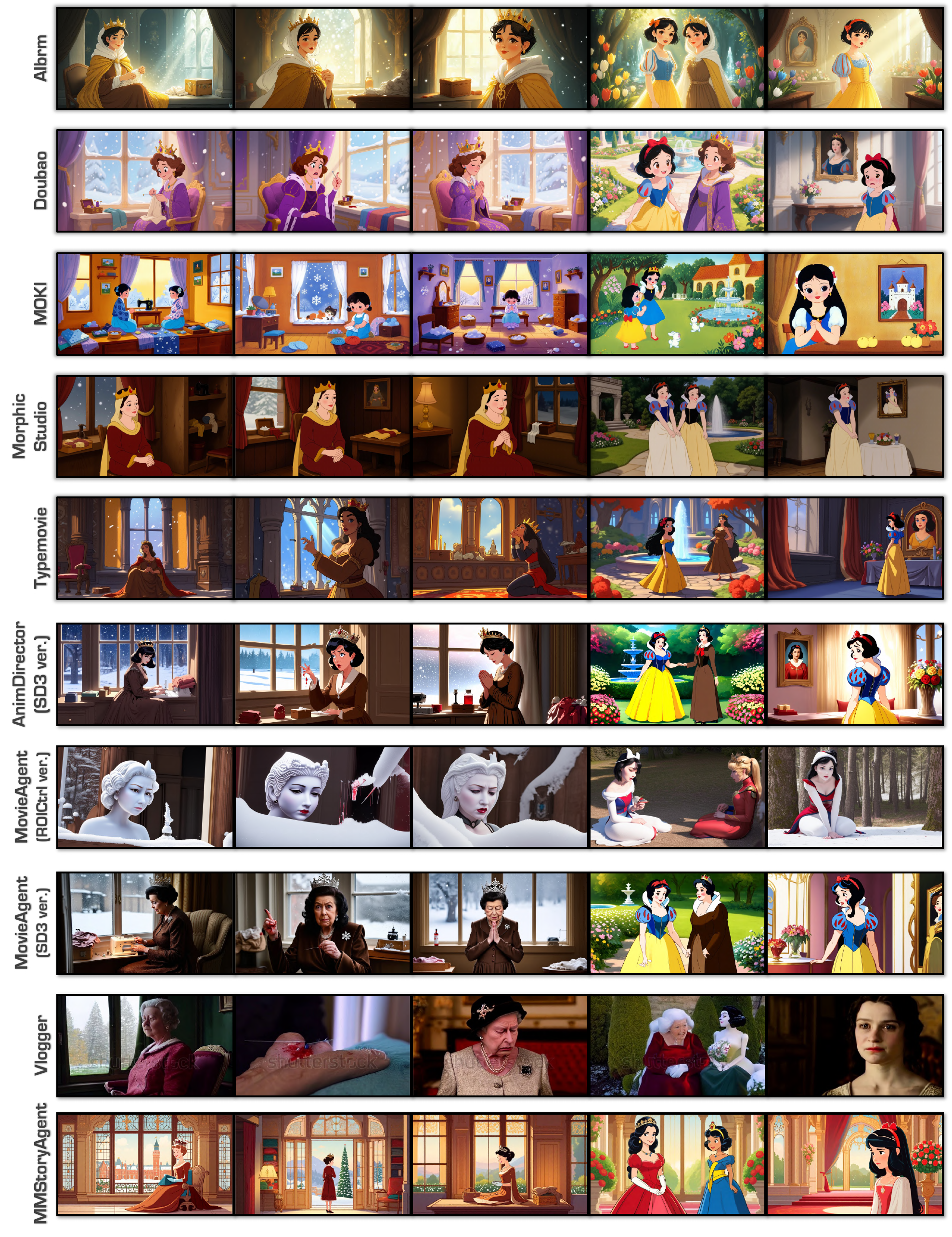}
\end{figure*}
\begin{figure*}[htpb]
    \centering
    \includegraphics[width=0.95\linewidth]{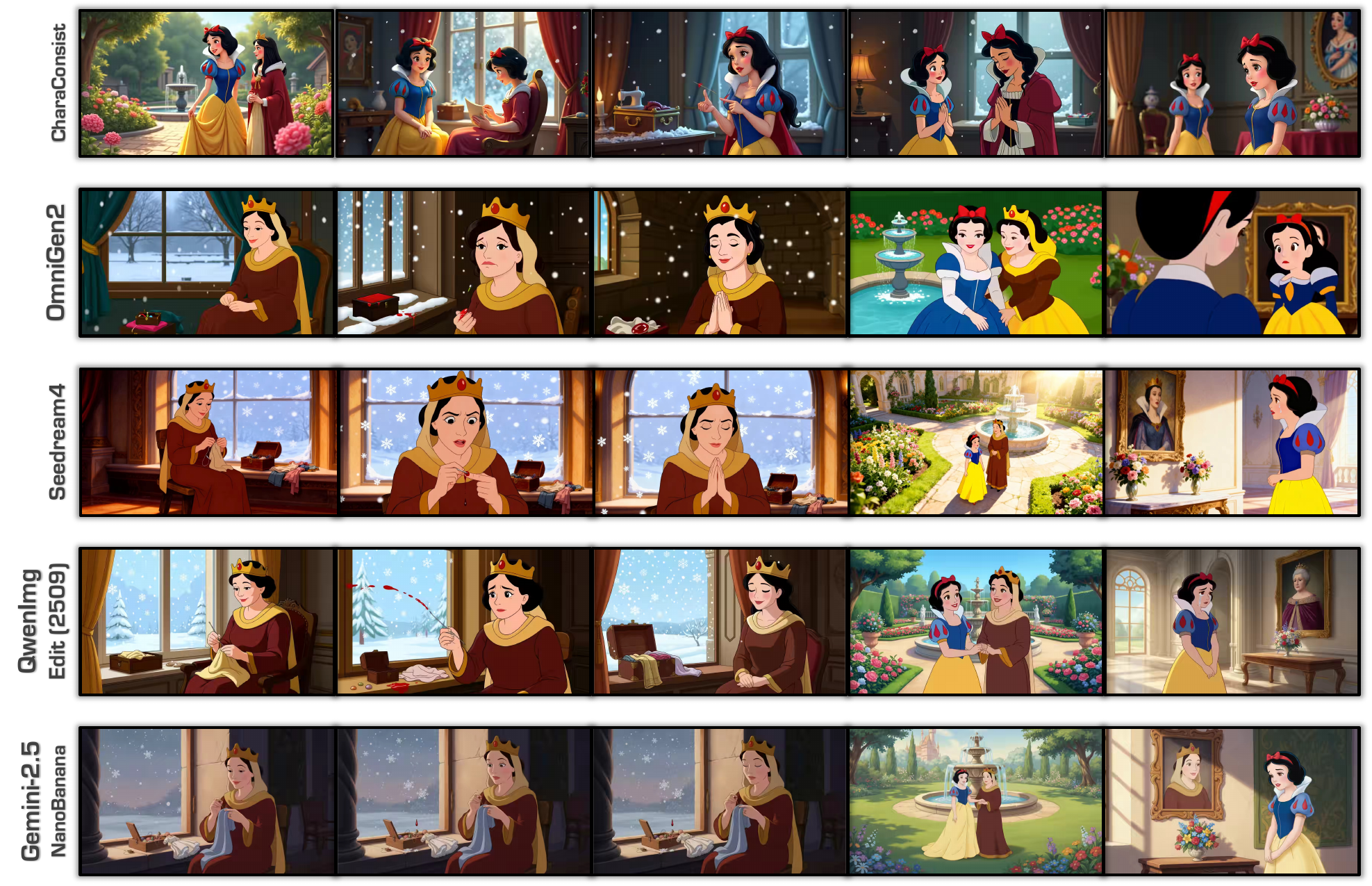}
    \caption{\small \textbf{Qualitative Result on Story 09}. From left to right are shot1 to shot5. Reference images of each shot's onstage characters is shown in Copy-Paste baseline results. 
    }
    \label{fig:qualitative}
\end{figure*}

\begin{figure*}[htpb]
    \centering
    \includegraphics[width=0.95\linewidth]{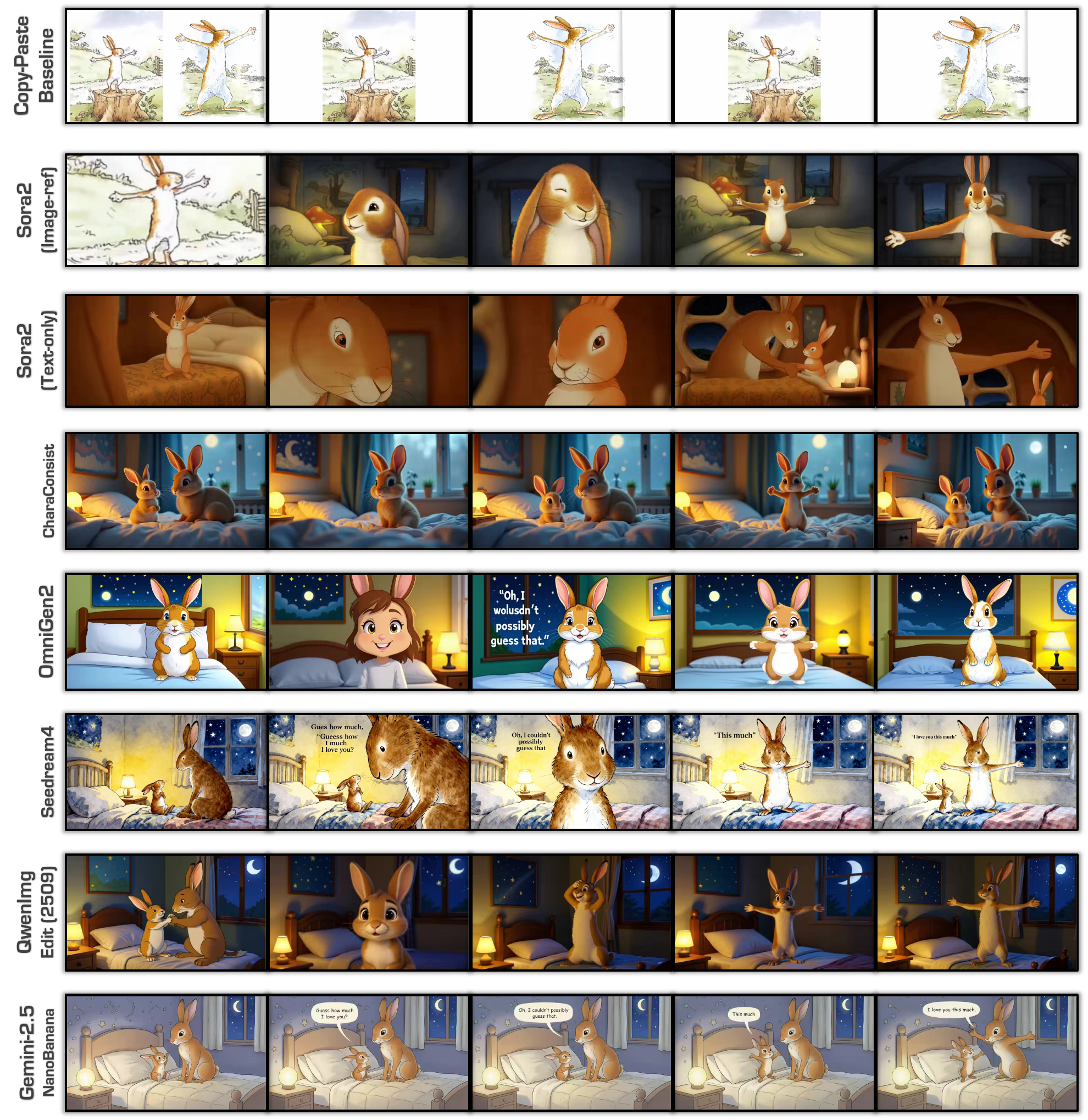}
    \caption{\small \textbf{Qualitative Result on Story 01}. From left to right are shot1 to shot5. Reference images of each shot's onstage characters is shown in Copy-Paste baseline results. 
    }
    \label{fig:qualitative_story01}
\end{figure*}
We provide the visualization generation results of the methods tested on \textbf{Story 09} and \textbf{Story 01}, as shown in Figure~\ref{fig:qualitative} and Figure~\ref{fig:qualitative_story01}. We sample the first five shots of the story for display to offer a concise yet representative comparison. At the top of the figure, we include a ``Copy-Paste Baseline'' that visually presents the ground-truth character presence in each frame using manually cropped reference images. This serves as a reference for evaluating the accuracy of character depiction across methods.
Below the baseline, we showcase the generation results from all 18 evaluated open-source methods (including their key variants), as well as several leading commercial tools. These comparisons highlight differences in prompt alignment, character consistency, and visual quality across models.

To facilitate better examination of the visual outputs, we also provide a full frame-by-frame visualization of the entire story. (Due to policy restrictions, the demo video was recorded anonymously and submitted as supplementary material.)

This detailed visualization enables a more comprehensive evaluation of each method’s performance over the complete narrative sequence, especially in terms of temporal coherence, character persistence, and scene transitions. We encourage readers to explore this link for in-depth visual analysis beyond the summarized frames shown in the main figure.

More results can be found on the website: \href{https://huggingface.co/datasets/ViStoryBench/ViStoryBenchResult}{\url{https://huggingface.co/datasets/ViStoryBench/ViStoryBenchResult}}

\section{ViStoryBench-Lite Benchmark}\label{sec:lite}

\subsection{Effectiveness of ViStoryBench-Lite}

To assess the representativeness and reliability of ViStoryBench-Lite, we conduct a comprehensive comparison between the full dataset and the \textbf{Lite} subset. We first perform a distributional analysis over story categories to examine the content diversity across both subsets. As shown in Figure~\ref{fig:dataset_stats_main}, the \textbf{Lite} subset exhibits a highly similar category distribution to the full dataset, suggesting a well-preserved narrative and visual diversity.

Further, we apply a unified generation method across both subsets and report quantitative results over all major evaluation dimensions. The detailed comparison is presented in Table~\ref{tab:lite_full}, which includes metrics on style consistency, character consistency, generative quality, diversity, and prompt alignment. The results demonstrate that the performance difference between the \textbf{Lite} and full datasets is minimal across most dimensions. Notably, only marginal discrepancies are observed in certain VLM-based prompt alignment metrics, such as scene-level consistency and Character Interaction, while the overall alignment score remains within a narrow error margin.

These findings validate the effectiveness of ViStoryBench-Lite as a representative subset of the full benchmark. This is particularly important for settings where large-scale human involvement (e.g. user study) or commercial API evaluation is required, such as user studies or commercial platform assessments. By using ViStoryBench-Lite, researchers and practitioners can achieve efficient and cost-effective evaluation without compromising result reliability.

\subsection{Full Evaluation on ViStoryBench-Lite}

\begin{table*}[h]
    \caption{
        \small\textbf{Quantitative Results of Various Story Visualization Methods on ViStoryBench-Lite}. Results highlighted with a gray background are excluded from ranking, for example, SEED-Story is trained on only three animations and does not aim for generalization, while the Copy-Paste Baseline directly pastes the character reference image as the output. For certain methods, we evaluate multiple inference configurations and report all corresponding results. \legendsquare{colorfirst} \legendsquare{colorsecond} \legendsquare{colorthird} \legendsquare{colorfourth} \legendsquare{colorfifth} indicate the first, second, third, fourth, and fifth performance, respectively. \textbf{CSD}: Style Similarity; \textbf{CIDS}: Character Similarity; \textbf{PA}: Prompt Alignment Score (CI: Character Interaction, IA: Individual Action); \textbf{CM}: OCCM; \textbf{Inc}: Inception Score; \textbf{Aes}: Aesthetics Score; \textbf{CP}: Copy-Paste. \imageref: With image reference; \textref: Only text input; \autoregressive: Auto-regressive mode; superscript $^k$ means scale=$k$.
        \textit{Note: \textbf{PA} scores are based on Gemini-3 Pro.}
    }
    \vspace{1mm}
    \centering
    \resizebox{1\textwidth}{!}{ 

    \begin{tabular}{l|c|cccc|ccccc|cccc}
        \toprule		
        \multirow{2}{*}[-0.28em]{\textbf{Method}} 
        &\multirow{2}{*}[-0.28em]{\textbf{Model}} 
        &\multicolumn{2}{c}{\textbf{CSD$\uparrow$}} 
        &\multicolumn{2}{c}{\textbf{CIDS$\uparrow$}} 
        &\multicolumn{5}{c}{\textbf{PA$\uparrow$}} 
        &\multirow{2}{*}[-0.28em]{\textbf{CM}$\uparrow$} 
        &\multirow{2}{*}[-0.28em]{\textbf{Inc}$\uparrow$} 
        &\multirow{2}{*}[-0.28em]{\textbf{Aes}$\uparrow$} 
        &\multirow{2}{*}[-0.28em]{\textbf{CP}}\\
        \cmidrule(l{7pt}r{7pt}){3-4}
        \cmidrule(l{7pt}r{7pt}){5-6}
        \cmidrule(l{7pt}r{7pt}){7-11} 
    
        \vspace{0.2em}
         & & Cross & Self & Cross & Self & Scene & Shot & CI & IA & Avg. \\

        \midrule
       \rowcolor{mygray}
        Copy-Paste Baseline & - & \cellcolor{mygray}{0.735} & \cellcolor{mygray}{0.770} & \cellcolor{mygray}{0.911} & \cellcolor{mygray}{0.993} & \cellcolor{mygray}{0.42} & \cellcolor{mygray}{1.60} & \cellcolor{mygray}{0.76} & \cellcolor{mygray}{0.85} & \cellcolor{mygray}{0.91} & \cellcolor{mygray}{92.76} & \cellcolor{mygray}{5.46} & \cellcolor{mygray}{4.39} & \cellcolor{mygray}{0.550} \\
        
        \midrule
        \rowcolor{teaserdatasetback}\multicolumn{15}{c}{\myfont \textbf{Story Image Method}} \\ \midrule
        
        \rowcolor{teaserdatasetback}StoryGen~\cite{Liu_2024_CVPR_Storygen}~\autoregressive & SD1.5 & \rankfifth{0.405} & 0.562 & \rankfifth{0.405} & 0.591 & 0.59 & 1.78 & 0.58 & 0.55 & 0.88 & 52.98 & 7.15 & 4.09 & 0.277 \\ 
        \rowcolor{teaserdatasetback}StoryGen~\cite{Liu_2024_CVPR_Storygen}~\imageref & SD1.5 & 0.396 & 0.551 & 0.396 & 0.602 & 0.59 & 1.94 & 0.55 & 0.25 & 0.83 & 52.53 & 7.67 & 4.09 & 0.224 \\ 
        \rowcolor{teaserdatasetback}StoryGen~\cite{Liu_2024_CVPR_Storygen}~\autoregressive\imageref & SD1.5 & 0.316 & 0.617 & 0.316 & \rankfourth{0.610} & 0.46 & 2.02 & 0.52 & 0.30 & 0.82 & 40.13 & 6.25 & 3.86 & 0.240 \\

        \rowcolor{teaserdatasetback}TheaterGen~\cite{cheng2024theatergen} & SD1.5 & 0.221 & 0.411 & 0.354 & 0.537 & 1.99 & 1.92 & 0.49 & 0.40 & 1.20 & 54.93 & \rankfirst{13.60} & 4.94 & 0.204 \\
        \rowcolor{teaserdatasetback}StoryDiffusion~\cite{zhou2024storydiffusion}~\textref & SDXL & 0.293 & \rankfourth{0.680} & 0.409 & 0.641 & 2.61 & 2.02 & 1.25 & 1.05 & 1.73 & 67.07 & \ranksecond{12.99} & \ranksecond{5.83} & 0.186 \\
        \rowcolor{teaserdatasetback}StoryDiffusion~\cite{zhou2024storydiffusion}~\imageref & SDXL & \rankfourth{0.409} & 0.611 & 0.460 & 0.575 & 1.35 & \rankfirst{2.68} & 1.34 & 1.25 & 1.66 & 62.48 & 8.18 & 5.21 & 0.251 \\
        \rowcolor{teaserdatasetback}SEED-Story~\cite{yang2024seedstory} & SDXL & 0.258 & \rankfirst{0.763} & 0.559 & 0.656 & 1.44 & 1.67 & 0.32 & 0.16 & 0.90 & 64.33 & 4.90 & 3.81 & 0.306 \\
        \rowcolor{teaserdatasetback}Story-Adapter~\cite{mao2024story_adapter}~\imageref$^0$ & SD1.5 & \rankfirst{0.518} & 0.609 & \rankfourth{0.490} & \rankfifth{0.605} & 1.42 & \rankthird{2.53} & 1.75 & 1.45 & 1.79 & 70.34 & 11.49 & 4.89 & 0.250 \\ 
        \rowcolor{teaserdatasetback}Story-Adapter~\cite{mao2024story_adapter}~\imageref$^5$ & SD1.5 & 0.371 & \ranksecond{0.758} & 0.425 & \rankthird{0.619} & 1.35 & \rankfifth{2.40} & 1.46 & 1.40 & 1.65 & 61.39 & \rankfifth{12.03} & 4.80 & 0.217 \\
        \rowcolor{teaserdatasetback}Story-Adapter~\cite{mao2024story_adapter}~\textref$^0$ & SD1.5 & 0.343 & 0.515 & 0.430 & 0.547 & 1.48 & \ranksecond{2.67} & 1.76 & 1.60 & 1.88 & 65.32 & \rankthird{12.72} & 5.12 & 0.203 \\ 
        \rowcolor{teaserdatasetback}Story-Adapter~\cite{mao2024story_adapter}~\textref$^5$ & SD1.5 & 0.353 & \rankthird{0.752} & 0.416 & 0.634 & 1.31 & \rankfourth{2.44} & 1.40 & 1.30 & 1.61 & 61.57 & 10.59 & 4.85 & 0.220 \\
        \rowcolor{teaserdatasetback}UNO~\cite{wu2025uno} & FLUX1 & \rankthird{0.425} & 0.648 & \ranksecond{0.512} & \ranksecond{0.630} & \rankfifth{3.12} & 2.25 & \rankfifth{1.98} & \rankfifth{1.75} & \rankfifth{2.28} & \rankfifth{70.88} & 10.50 & 5.13 & 0.287 \\
        \rowcolor{teaserdatasetback}OmniGen2~\cite{wu2025omnigen2} & DiT & \ranksecond{0.491} & \rankfifth{0.648} & \rankfirst{0.576} & \rankfirst{0.668} & \rankfourth{3.18} & 2.38 & \rankfourth{2.15} & \rankthird{1.90} & \rankfourth{2.40} & \rankthird{73.44} & 8.21 & 5.21 & 0.298 \\
        \rowcolor{teaserdatasetback}CharaConsist~\cite{wang2025characonsist} & FLUX1 & 0.333 & 0.646 & 0.347 & 0.539 & \rankthird{3.24} & 2.37 & 1.82 & 1.63 & 2.27 & 62.10 & 10.84 & \rankthird{5.78} & 0.216 \\
        \rowcolor{teaserdatasetback}QwenImageEdit-2509~\cite{qwen2025qwenimage} & DiT & 0.404 & 0.614 & \rankthird{0.482} & 0.541 & \rankfirst{3.37} & 2.10 & \ranksecond{2.45} & \rankfirst{2.20} & \rankfirst{2.53} & 61.27 & 10.56 & \rankfifth{5.46} & 0.249 \\

        \midrule
        \rowcolor{blue!3}\multicolumn{15}{c}{\myfont \textbf{Story Video Method}} \\ \midrule
        \rowcolor{blue!3}Vlogger~\cite{zhuang2024vlogger}~\textref  & SD1.4 & 0.240 & 0.462 & 0.369 & 0.524 & 1.12 & 2.25 & 1.60 & 1.35 & 1.58 & \ranksecond{77.13} & 8.41 & 4.24 & 0.209 \\
        \rowcolor{blue!3}Vlogger~\cite{zhuang2024vlogger}~\imageref & SD1.4 & 0.299 & 0.497 & 0.373 & 0.519 & 1.14 & 2.24 & 1.53 & 1.35 & 1.57 & \rankfirst{79.14} & 8.83 & 4.24 & 0.211 \\ 
        \rowcolor{blue!3}AnimDirector~\cite{li2024anim} & SD3 & 0.305 & 0.558 & 0.423 & 0.593 & \ranksecond{3.27} & 2.11 & \rankthird{2.44} & \ranksecond{2.05} & \ranksecond{2.47} & \rankfourth{72.03} & 9.94 & \rankfourth{5.60} & 0.206 \\ 
        \rowcolor{blue!3}MMStoryAgent~\cite{xu2024mmstoryagent} & SDXL & 0.261 & 0.661 & 0.385 & 0.598 & 2.44 & 1.94 & 1.14 & 0.95 & 1.62 & 58.24 & 8.09 & \rankfirst{5.91} & 0.189 \\ 
        \rowcolor{blue!3}MovieAgent~\cite{wu2025movieagent} & SD1.5 & 0.236 & 0.564 & 0.372 & 0.568 & 0.93 & 1.81 & 0.88 & 0.85 & 1.12 & 64.41 & 10.06 & 4.69 & 0.261 \\
        \rowcolor{blue!3}MovieAgent~\cite{wu2025movieagent} & SD3 & 0.346 & 0.539 & 0.433 & 0.582 & 3.09 & 2.28 & \rankfirst{2.45} & \rankfourth{1.90} & \rankthird{2.43} & 67.29 & \rankfourth{12.04} & 5.32 & 0.199 \\

        \midrule
        \rowcolor{SIMback}\multicolumn{15}{c}{\myfont \textbf{Commercial Platform}} \\ \midrule
        \rowcolor{SIMback}MOKI~\cite{moki2024} & - & 0.214 & \rankfourth{0.694} & 0.372 & \rankfourth{0.621} & 2.29 & 1.56 & 0.58 & 0.45 & 1.22 & 45.96 & 10.36 & \textbf{\rankfirst{5.79}} & 0.211 \\
        \rowcolor{SIMback}MorphicStudio~\cite{morphic2024studio} & - & \textbf{\rankfirst{0.577}} & 0.628 & \ranksecond{0.603} & 0.677 & 3.01 & 2.31 & 1.77 & 1.32 & 2.10 & 60.79 & 9.00 & 4.96 & 0.234 \\
        \rowcolor{SIMback}AIbrm~\cite{brmgo2025} & - & \rankfifth{0.412} & \textbf{\rankfirst{0.730}} & \rankfifth{0.557} & \ranksecond{0.740} & 3.06 & 2.18 & 1.67 & 1.55 & 2.12 & \ranksecond{75.53} & 9.53 & \ranksecond{5.72} & 0.223 \\
        \rowcolor{SIMback}ShenBi~\cite{shenbi2025} & - & 0.275 & 0.575 & 0.418 & 0.585 & 3.49 & 2.35 & 2.65 & 2.11 & 2.65 & 61.33 & \rankthird{11.60} & 5.07 & 0.197 \\
        \rowcolor{SIMback}Typemovie~\cite{typemovie2024} & - & 0.325 & 0.646 & 0.464 & 0.621 & 2.34 & 2.17 & 1.62 & 1.40 & 1.88 & \rankfourth{74.14} & \rankfourth{11.15} & 5.32 & 0.168 \\
        \rowcolor{SIMback}Doubao~\cite{doubao2024} & - & 0.367 & \rankthird{0.695} & 0.446 & 0.642 & \rankthird{3.88} & 2.41 & \rankfourth{3.23} & \rankfifth{2.65} & \rankfifth{3.04} & 65.23 & 9.88 & \rankthird{5.61} & 0.255 \\

        \midrule
        \rowcolor{red!5}\multicolumn{15}{c}{\myfont \textbf{Multi-modal Large Model (Language, Image and Video)}} \\ \midrule
        \rowcolor{red!5}GPT-4o$^*$~\cite{hurst2024gpt4o} & - & \rankthird{0.481} & 0.680 & 0.420 & 0.522 & \rankfifth{3.82} & \ranksecond{2.82} & \ranksecond{3.58} & \ranksecond{3.12} & \ranksecond{3.34} & 69.33 & 9.02 & 5.49 & 0.209 \\
        \rowcolor{red!5}Gemini-2.0$^*$~\cite{gemini2.02025} & - & 0.361 & 0.573 & \rankfourth{0.573} & 0.677 & 3.26 & 2.46 & 2.43 & 2.00 & 2.54 & \rankthird{74.82} & 10.12 & 4.91 & 0.266 \\
        \rowcolor{red!5}Gemini-2.5$^*$~\cite{google2025nanobanana} & - & \rankfourth{0.447} & 0.657 & 0.553 & 0.642 & \ranksecond{3.89} & 2.32 & \rankthird{3.26} & \rankthird{2.90} & \rankthird{3.09} & 64.86 & \rankfifth{10.54} & \rankfourth{5.61} & 0.255 \\
        \rowcolor{red!5}Gemini-3.0 Pro$^*$~\cite{google2025nanobananapro} & - & 0.385 & 0.622 & \rankthird{0.581} & 0.653 & \rankfirst{3.94} & \rankfifth{2.58} & \rankfirst{3.71} & \rankfirst{3.25} & \rankfirst{3.37} & 59.97 & \textbf{\rankfirst{12.50}} & \rankfifth{5.54} & 0.244 \\
        \rowcolor{red!5}Seedream-4.0~\cite{gong2025seedream} & - & 0.369 & 0.585 & 0.280 & 0.539 & \rankfourth{3.82} & \rankfourth{2.59} & \rankfifth{3.20} & \rankfourth{2.68} & \rankfourth{3.07} & 49.45 & \ranksecond{12.12} & 5.21 & 0.201 \\
        \rowcolor{red!5}Sora2$^*$~\cite{2025sora2}~\imageref & - & \ranksecond{0.515} & \ranksecond{0.713} & \textbf{\rankfirst{0.766}} & \textbf{\rankfirst{0.839}} & 3.08 & \rankthird{2.76} & 2.91 & 2.50 & 2.81 & \textbf{\rankfirst{81.42}} & 6.53 & 4.72 & 0.158 \\
        \rowcolor{red!5}Sora2$^*$~\cite{2025sora2}~\textref & - & 0.365 & \rankfifth{0.685} & 0.364 & 0.561 & 3.01 & \rankfirst{2.83} & 2.96 & 2.33 & 2.78 & \rankfifth{72.67} & 9.68 & 4.52 & 0.158 \\
        \bottomrule
    \end{tabular}
    }
    \label{table:lite}
    \label{tab:comparison_lite}
    \vspace{-1em}
\end{table*}

\begin{table*}[h]
    \caption{
        \small\textbf{Qwen-Based Prompt Alignment Evaluation of Story Visualization Methods on ViStoryBench and ViStoryBench-Lite}. While the Copy-Paste Baseline directly pastes the character reference image as the output. For certain methods, we evaluate multiple inference configurations and report all corresponding results. \legendsquare{colorfirst} \legendsquare{colorsecond} \legendsquare{colorthird} \legendsquare{colorfourth} \legendsquare{colorfifth} indicate the first, second, third, fourth, and fifth performance, respectively. \textbf{PA}: Prompt Alignment Score (CI: Character Interaction, IA: Individual Action); \imageref: With image reference; \textref: Only text input; \autoregressive: Auto-regressive mode; superscript $^k$ means scale=$k$.
    }
    \label{tab:comparison_full_lite_pa_qwen}
    \vspace{1mm}
    \centering
    \resizebox{0.8\textwidth}{!}{ 

    \begin{tabular}{l|c|ccccc|ccccc}
        \toprule		
        \multirow{2}{*}[-0.28em]{\textbf{Method}} 
        &\multirow{2}{*}[-0.28em]{\textbf{Model}} 
        &\multicolumn{5}{c}{\textbf{PA (lite)$\uparrow$}} 
        &\multicolumn{5}{c}{\textbf{PA (full)$\uparrow$}}
    
        \\
         & &Scene & Shot & CI & IA & Avg. & Scene & Shot & CI & IA & Avg. \\

        \midrule
        \rowcolor{mygray}Copy-Paste Baseline & - & 
        0.30 & 2.29 & 0.80 & 1.11 & 1.12 & 
        0.30 & 2.29 & 0.80 & 1.11 & 1.12 \\
         
        \midrule
        \rowcolor{teaserdatasetback}\multicolumn{12}{c}{\myfont \textbf{Story Image Method}} \\ 
        \midrule
         
        \rowcolor{teaserdatasetback}StoryGen~\cite{Liu_2024_CVPR_Storygen}~\autoregressive & SD1.5 & 0.66 & 2.75 & 0.85 & 1.12 & 1.35 & 0.65 & 2.62 & 0.84 & 1.10 & 1.30 \\ 
        \rowcolor{teaserdatasetback}StoryGen~\cite{Liu_2024_CVPR_Storygen}~\imageref & SD1.5 & 0.71 & 2.57 & 0.82 & 0.98 & 1.27 & 0.72 & 2.65 & 0.85 & 1.07 & 1.32 \\ 
        \rowcolor{teaserdatasetback}StoryGen~\cite{Liu_2024_CVPR_Storygen}~\autoregressive\imageref & SD1.5 & 0.48 & 2.73 & 0.74 & 1.18 & 1.28 & 0.56 & 2.68 & 0.83 & 1.14 & 1.30 \\

        \rowcolor{teaserdatasetback}TheaterGen~\cite{cheng2024theatergen} & SD1.5 & 2.28 & 2.32 & 0.55 & 0.73 & 1.47 & 2.16 & 2.26 & 0.49 & 0.67 & 1.39 \\
        \rowcolor{teaserdatasetback}StoryDiffusion~\cite{zhou2024storydiffusion}~\textref & SDXL & 2.73 & 3.52 & 1.23 & 1.38 & 2.21 & 2.77 & 3.48 & 1.40 & 1.45 & 2.27 \\
        \rowcolor{teaserdatasetback}StoryDiffusion~\cite{zhou2024storydiffusion}~\imageref & SDXL & 1.40 & 3.43 & 1.62 & 1.82 & 2.07 & 1.38 & 3.35 & 1.73 & 1.72 & 2.04 \\
        \rowcolor{teaserdatasetback}SEED-Story~\cite{yang2024seedstory} & SDXL & 1.62 & 2.79 & 0.32 & 0.44 & 1.29 & 1.64 & 2.57 & 0.54 & 0.51 & 1.32 \\
        \rowcolor{teaserdatasetback}Story-Adapter~\cite{mao2024story_adapter}~\imageref$^0$ & SD1.5 & 1.60 & 3.48 & 2.00 & \rankfifth{2.05} & 2.28 & 1.68 & 3.47 & 2.17 & \rankfifth{2.07} & 2.35 \\ 
        \rowcolor{teaserdatasetback}Story-Adapter~\cite{mao2024story_adapter}~\imageref$^5$ & SD1.5 & 1.65 & 3.48 & 1.83 & 1.74 & 2.17 & 1.67 & 3.50 & 1.99 & 1.86 & 2.26 \\
        \rowcolor{teaserdatasetback}Story-Adapter~\cite{mao2024story_adapter}~\textref$^0$ & SD1.5 & 1.81 & 3.54 & 2.35 & \rankfourth{2.12} & 2.45 & 1.83 & 3.51 & \rankfifth{2.28} & \rankthird{2.11} & 2.43 \\ 
        \rowcolor{teaserdatasetback}Story-Adapter~\cite{mao2024story_adapter}~\textref$^5$ & SD1.5 & 1.57 & 3.46 & 1.84 & 1.77 & 2.16 & 1.64 & 3.49 & 1.99 & 1.87 & 2.25 \\
        \rowcolor{teaserdatasetback}UNO~\cite{wu2025uno} & FLUX1 & \rankfourth{3.11} & 3.46 & \rankfifth{2.38} & 1.94 & \rankfifth{2.72} & \rankfourth{3.03} & \rankfourth{3.53} & 2.22 & 1.88 & \rankfifth{2.67} \\
        \rowcolor{teaserdatasetback}OmniGen2~\cite{wu2025omnigen2} & DiT & \rankthird{3.16} & \rankfourth{3.56} & \rankfourth{2.43} & 2.03 & \rankfourth{2.80} & \rankthird{3.09} & \ranksecond{3.59} & \rankthird{2.48} & \rankfourth{2.07} & \rankthird{2.81} \\
        \rowcolor{teaserdatasetback}CharaConsist~\cite{wang2025characonsist} & FLUX1 & 2.54 & \rankthird{3.59} & 1.56 & 1.40 & 2.27 & 2.50 & 3.53 & 1.61 & 1.39 & 2.26 \\
        \rowcolor{teaserdatasetback}QwenImageEdit-2509~\cite{qwen2025qwenimage} & DiT & \ranksecond{3.24} & \rankfifth{3.55} & \rankthird{2.64} & \ranksecond{2.22} & \rankthird{2.91} & \rankfifth{2.98} & \rankfifth{3.53} & \rankfourth{2.46} & 2.07 & \rankfourth{2.76} \\
    
        \midrule
        \rowcolor{blue!3}\multicolumn{12}{c}{\myfont \textbf{Story Video Method}} \\ 
        \midrule
        \rowcolor{blue!3}Vlogger~\cite{zhuang2024vlogger}~\textref  & SD1.4 & 1.41 & 3.31 & 2.12 & 1.95 & 2.20 & 1.36 & 3.27 & 2.10 & 1.88 & 2.15 \\
        \rowcolor{blue!3}Vlogger~\cite{zhuang2024vlogger}~\imageref & SD1.4 & 1.37 & 3.28 & 1.96 & 1.81 & 2.10 & 1.35 & 3.24 & 2.07 & 1.90 & 2.14 \\ 
        \rowcolor{blue!3}AnimDirector~\cite{li2024anim} & SD3 & \rankfirst{3.37} & \rankfirst{3.59} & \rankfirst{2.98} & \rankthird{2.17} & \rankfirst{3.03} & \rankfirst{3.32} & \rankfirst{3.61} & \rankfirst{2.97} & \ranksecond{2.28} & \rankfirst{3.04} \\ 
        \rowcolor{blue!3}MMStoryAgent~\cite{xu2024mmstoryagent} & SDXL & 2.64 & 3.19 & 1.10 & 1.25 & 2.04 & 2.55 & 3.30 & 1.34 & 1.24 & 2.11 \\ 
        \rowcolor{blue!3}MovieAgent~\cite{wu2025movieagent} & SD1.5 & 0.96 & 3.07 & 0.86 & 0.96 & 1.46 & 0.95 & 3.10 & 0.80 & 0.79 & 1.41 \\
        \rowcolor{blue!3}MovieAgent~\cite{wu2025movieagent} & SD3 & \rankfifth{3.09} & \ranksecond{3.59} & \ranksecond{2.90} & \rankfirst{2.31} & \ranksecond{2.97} & \ranksecond{3.14} & \rankthird{3.58} & \ranksecond{2.96} & \rankfirst{2.32} & \ranksecond{3.00} \\        
        
        \midrule
        \rowcolor{SIMback}\multicolumn{12}{c}{\myfont \textbf{Commercial Platform}} \\ 
        \midrule
        \rowcolor{SIMback}MOKI~\cite{moki2024} & - & 2.49 & 2.32 & 0.54 & 0.60 & 1.49 & - & - & - & - & - \\
        \rowcolor{SIMback}MorphicStudio~\cite{morphic2024studio} & - & 3.00 & 3.47 & 2.12 & 1.88 & 2.62 & - & - & - & - & - \\
        \rowcolor{SIMback}AIbrm~\cite{brmgo2025} & - & 2.92 & 3.43 & 1.73 & 1.64 & 2.43 & - & - & - & - & - \\
        \rowcolor{SIMback}ShenBi~\cite{shenbi2025} & - & 3.48 & 3.51 & 3.11 & 2.25 & 3.09 & - & - & - & - & - \\
        \rowcolor{SIMback}Typemovie~\cite{typemovie2024} & - & 2.34 & 3.51 & 1.94 & 1.82 & 2.40 & - & - & - & - & - \\
        \rowcolor{SIMback}Doubao~\cite{doubao2024} & - & \ranksecond{3.80} & \rankthird{3.60} & \rankthird{3.51} & \rankfourth{2.73} & \ranksecond{3.41} & - & - & - & - & - \\
        
        \midrule
        \rowcolor{red!5}\multicolumn{12}{c}{\myfont \textbf{Multi-modal Large Model (Language, Image and Video)}} \\ 
        \midrule
        \rowcolor{red!5}GPT-4o$^*$~\cite{hurst2024gpt4o} & - & \rankthird{3.68} & \rankfirst{3.79} & \ranksecond{3.55} & \ranksecond{2.95} & \rankfirst{3.49} & - & - & - & - & - \\
        \rowcolor{red!5}Gemini-2.0$^*$~\cite{gemini2.02025} & - & 3.08 & 3.55 & 2.68 & 2.07 & 2.84 & - & - & - & - & - \\
        \rowcolor{red!5}Gemini-2.5$^*$~\cite{google2025nanobanana} & - & \rankfourth{3.61} & 3.48 & \rankfifth{3.19} & \rankfifth{2.72} & \rankfifth{3.25} & - & - & - & - & - \\
        \rowcolor{red!5}Seedream-4.0~\cite{gong2025seedream} & - & \rankfifth{3.61} & \rankfourth{3.58} & \rankfourth{3.23} & \rankthird{2.75} & \rankfourth{3.29} & - & - & - & - & - \\
        \rowcolor{red!5}Sora2$^*$~\cite{2025sora2}~\imageref & - & 3.05 & \rankfifth{3.56} & 3.14 & 2.20 & 2.99 & - & - & - & - & - \\
        \rowcolor{red!5}Sora2$^*$~\cite{2025sora2}~\textref & - & 2.83 & \ranksecond{3.62} & 2.92 & 2.21 & 2.89 & - & - & - & - & - \\
        \bottomrule
        
    \end{tabular}
    }
    \vspace{-1em}
\end{table*}

\definecolor{CharConsSelf}{HTML}{dbe4ec}
\definecolor{CharConsCross}{HTML}{4b79a1}
\definecolor{StyleConsSelf}{HTML}{d8ebe3}
\definecolor{StyleConsCross}{HTML}{3a9d74}
\definecolor{PromptAlignScene}{HTML}{f4dcdc}
\definecolor{PromptAlignShot}{HTML}{ecc0c0}
\definecolor{PromptAlignOccm}{HTML}{e3a3a3}
\definecolor{PromptAlignGlobalChar}{HTML}{da8787}
\definecolor{PromptAlignSingleChar}{HTML}{d26b6b}
\definecolor{PromptAlignAvg}{HTML}{c94f4f}
\definecolor{Aesthetics}{HTML}{e08e3e}
\definecolor{InceptionScore}{HTML}{7e7d85}

\begin{table*}[h!]
    \captionsetup{type=table}
    \caption{\selectfont \small \textbf{Best Normalized Polar Visualization of SOTA Methods on ViStoryBench-Lite}. Anti-Clockwisely, \textbf{\underline{Character Similarity}}: \legendsquarebox{CharConsSelf} Self-similarity, \legendsquarebox{CharConsCross} Cross-similarity; \textbf{\underline{Style Similarity}}: \legendsquarebox{StyleConsSelf} Self-similarity, \legendsquarebox{StyleConsCross} Cross-similarity; \textbf{\underline{Prompt Alignment}}: \legendsquarebox{PromptAlignScene} Scene, \legendsquarebox{PromptAlignShot} Shot, \legendsquarebox{PromptAlignGlobalChar} Character Interaction~(CI), \legendsquarebox{PromptAlignSingleChar} Individual Action~(IA), \legendsquarebox{PromptAlignAvg} Average Score; \textbf{\underline{Generation Quality}}: \legendsquarebox{Aesthetics} Aesthetics; \textbf{\underline{Diversity}}: \legendsquarebox{InceptionScore} Inception Score.}
\centering
\resizebox{1\linewidth}{!}{
\begin{tabular}{ccccc}
\toprule
\rowcolor{blue!3}\multicolumn{5}{c}{\myfont \textbf{Story Image Method}}
\\
\midrule
\texttt{StoryGen (Multi-image)} & 
\texttt{TheaterGen} & 
\texttt{StoryDiffusion (Img-ref)} & 
\texttt{SeedStory} & 
\texttt{StoryAdapter (Img-ref, scale=0)} 
\\ 
\includegraphics[width=0.3\linewidth]{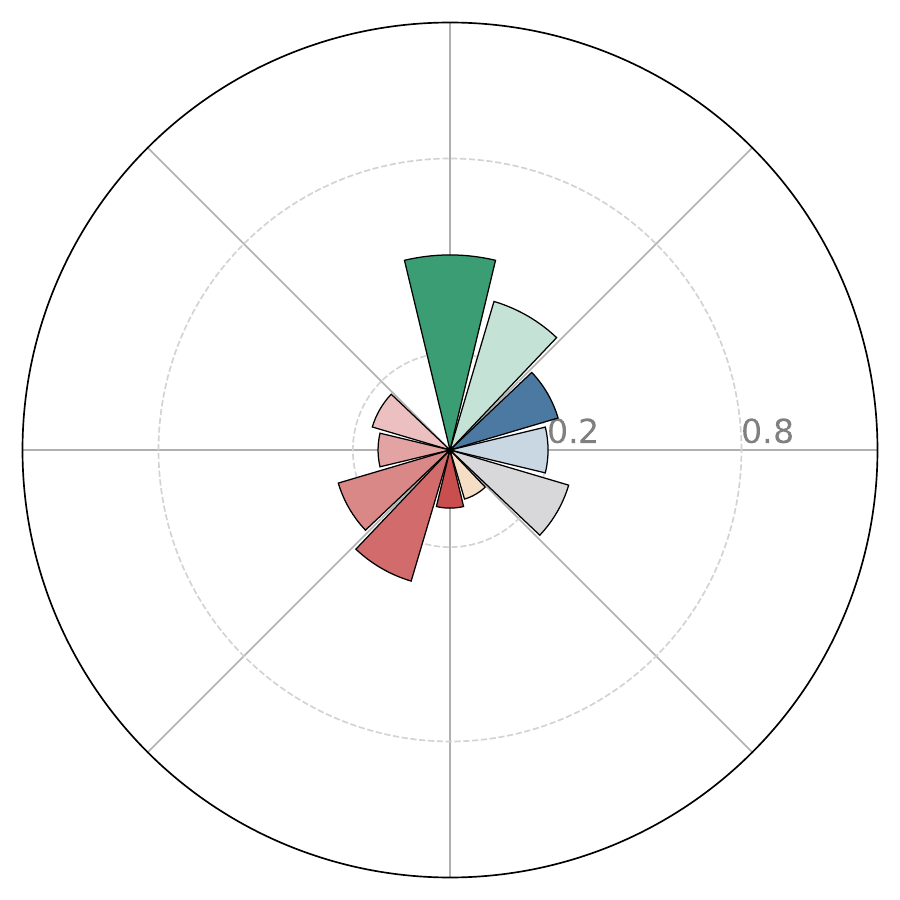} & 
\includegraphics[width=0.3\linewidth]{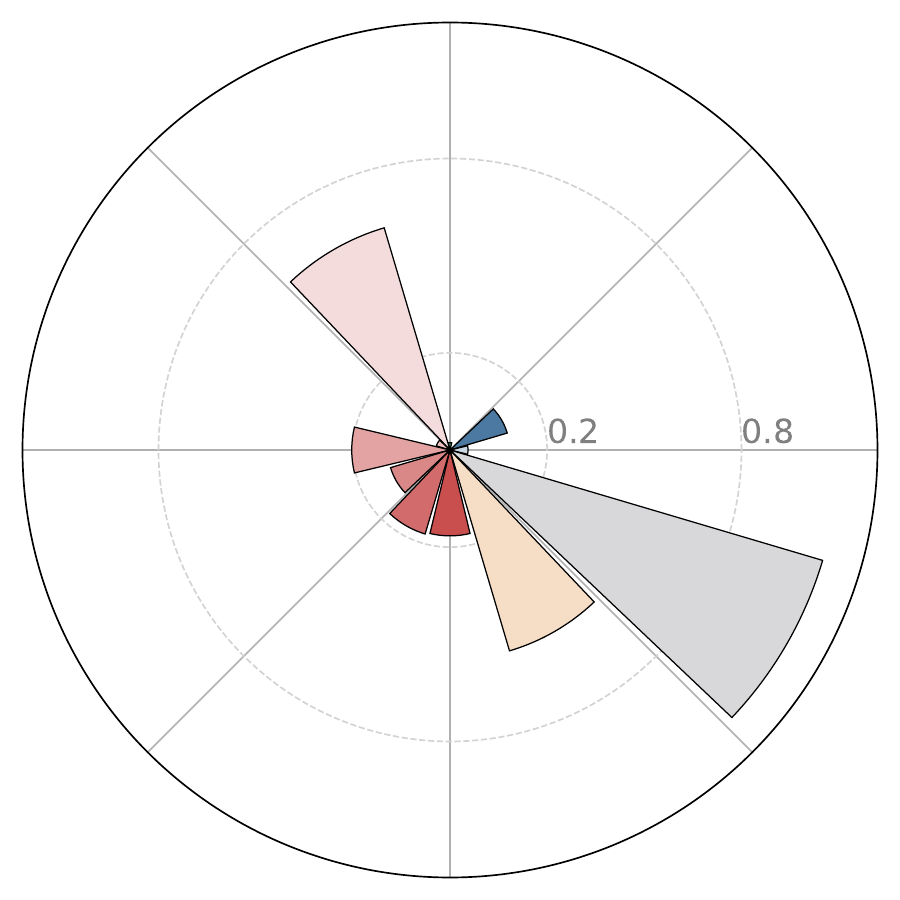} & 
\includegraphics[width=0.3\linewidth]{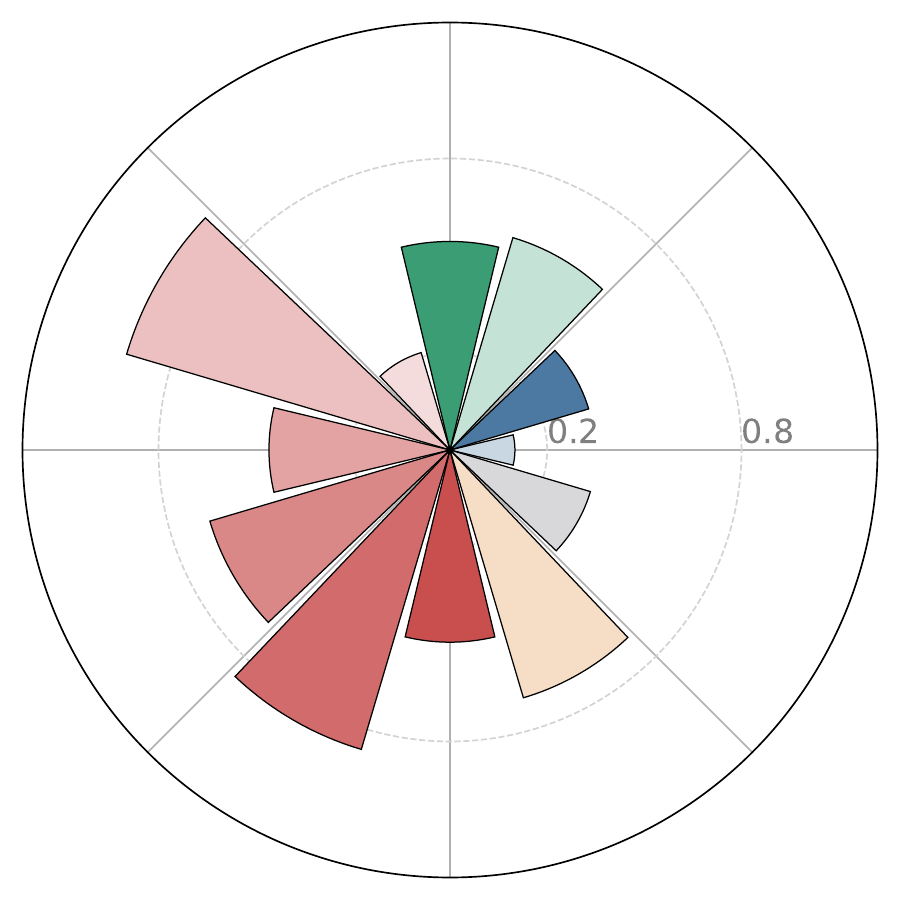} & 
\includegraphics[width=0.3\linewidth]{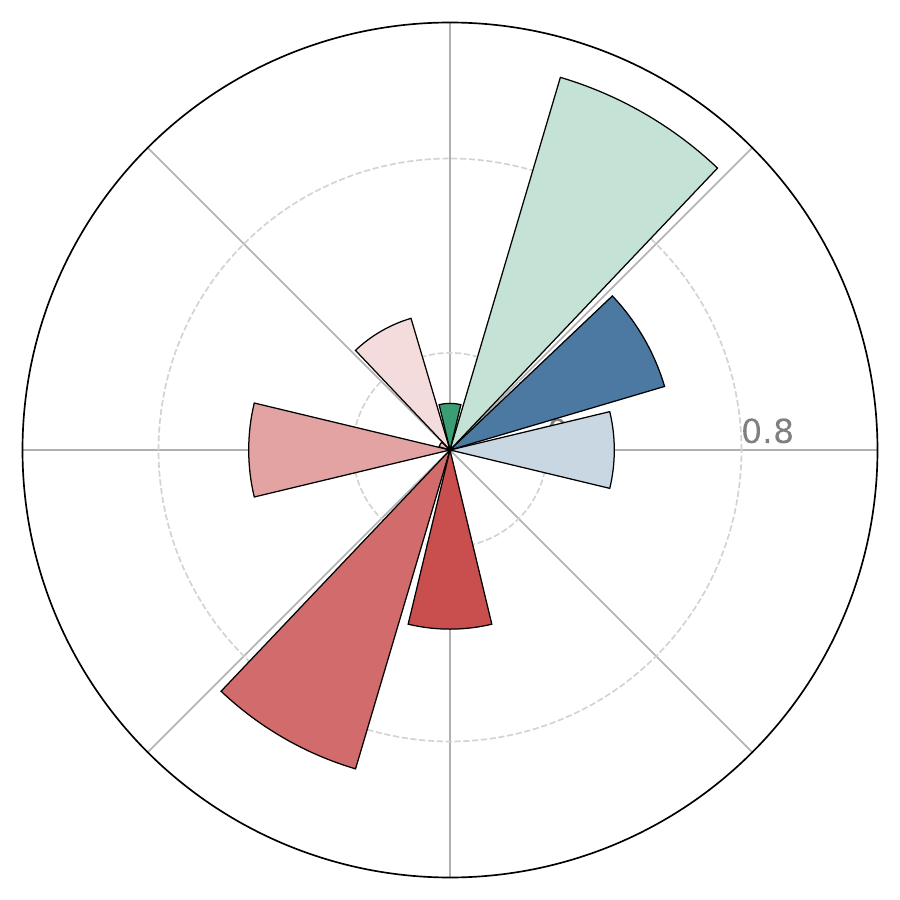} & 
\includegraphics[width=0.3\linewidth]{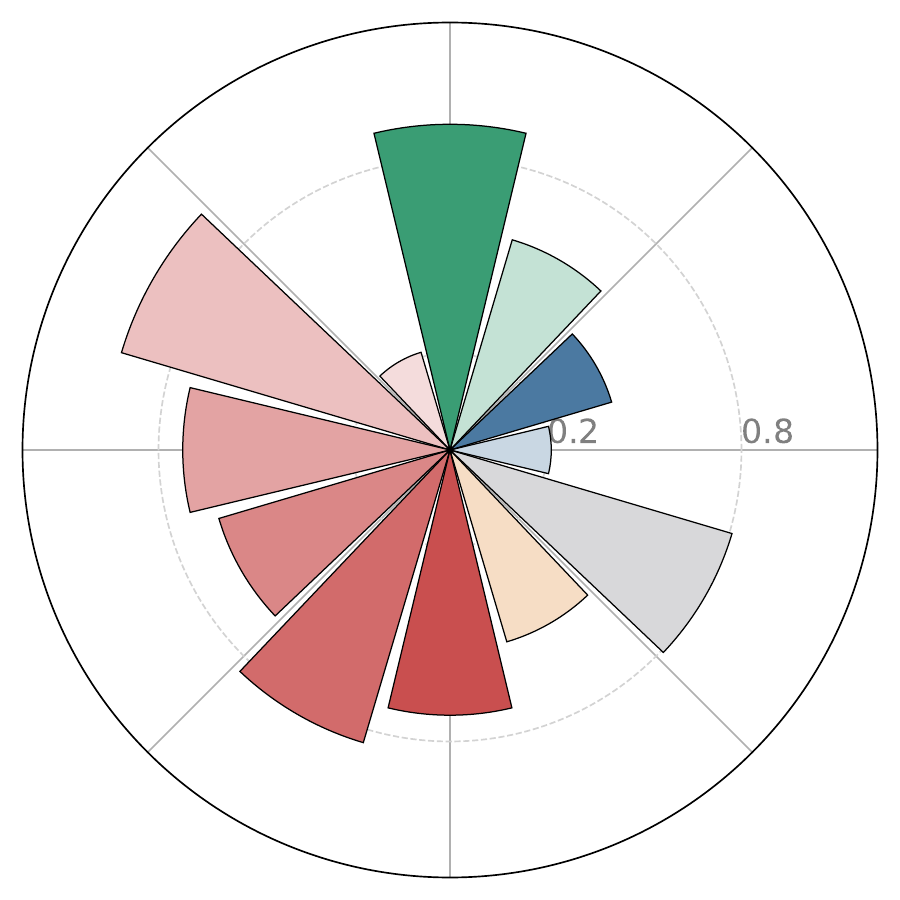} 
\\
\midrule
\texttt{UNO} & 
\texttt{OmniGen2} & 
\texttt{CharaConsist} & 
\texttt{QwenImageEdit-2509} &
\\ 
\includegraphics[width=0.3\linewidth]{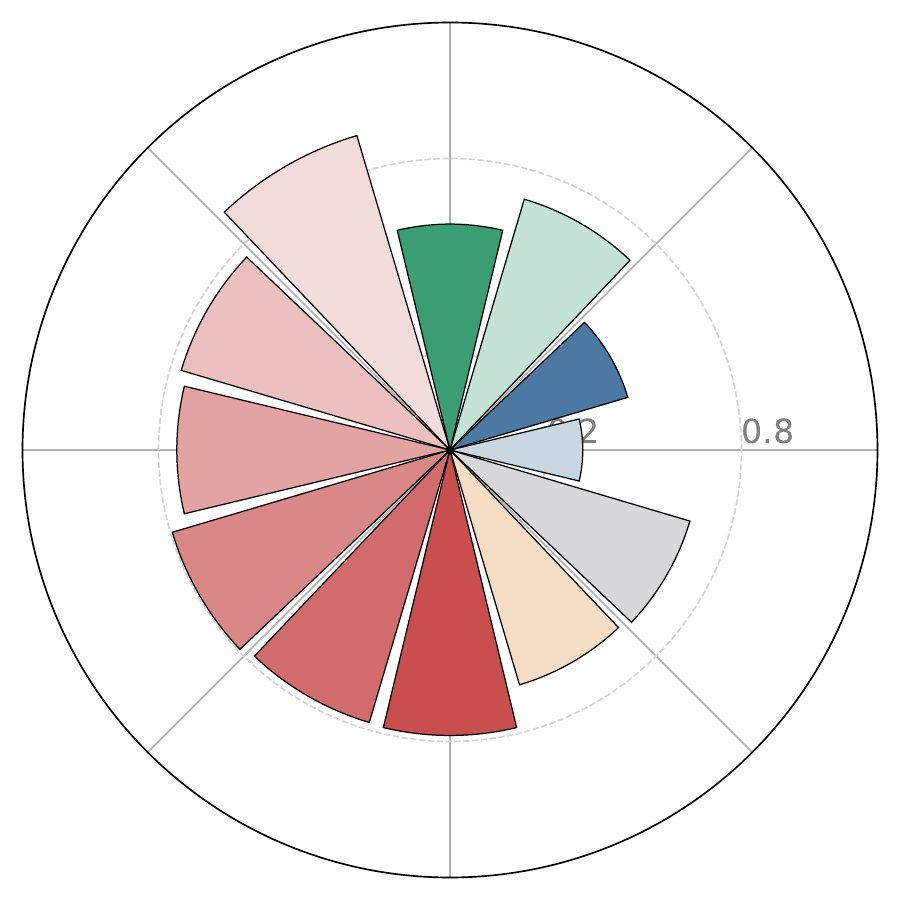} & 
\includegraphics[width=0.3\linewidth]{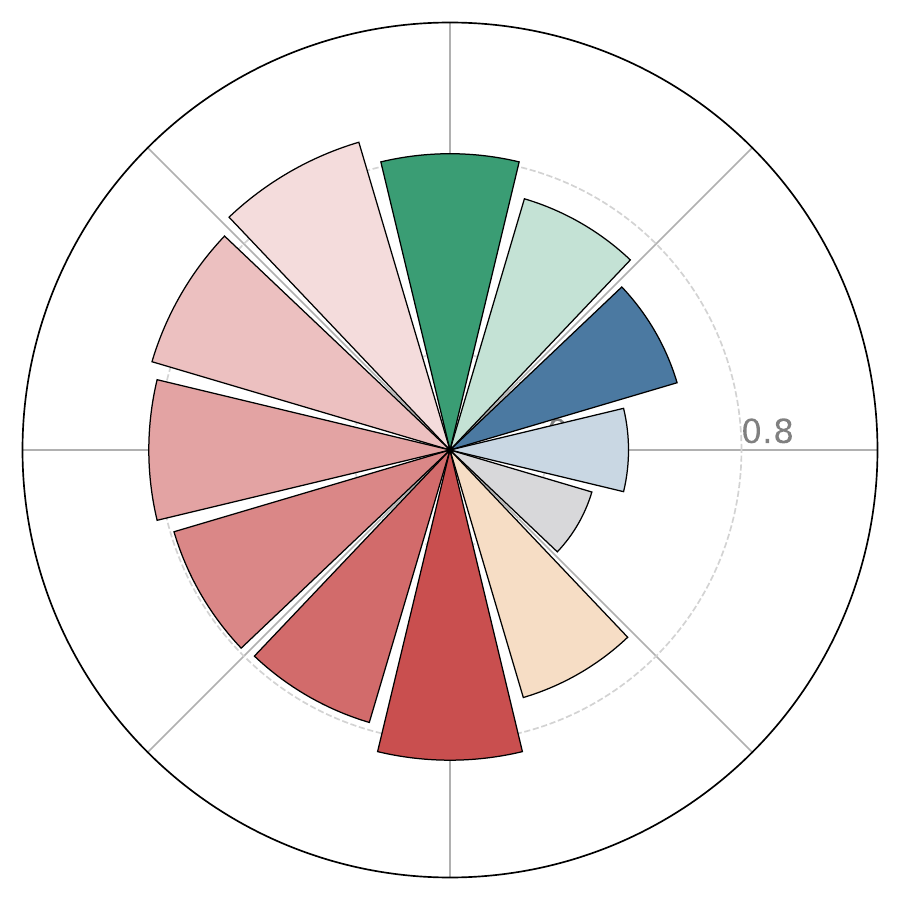} & 
\includegraphics[width=0.3\linewidth]{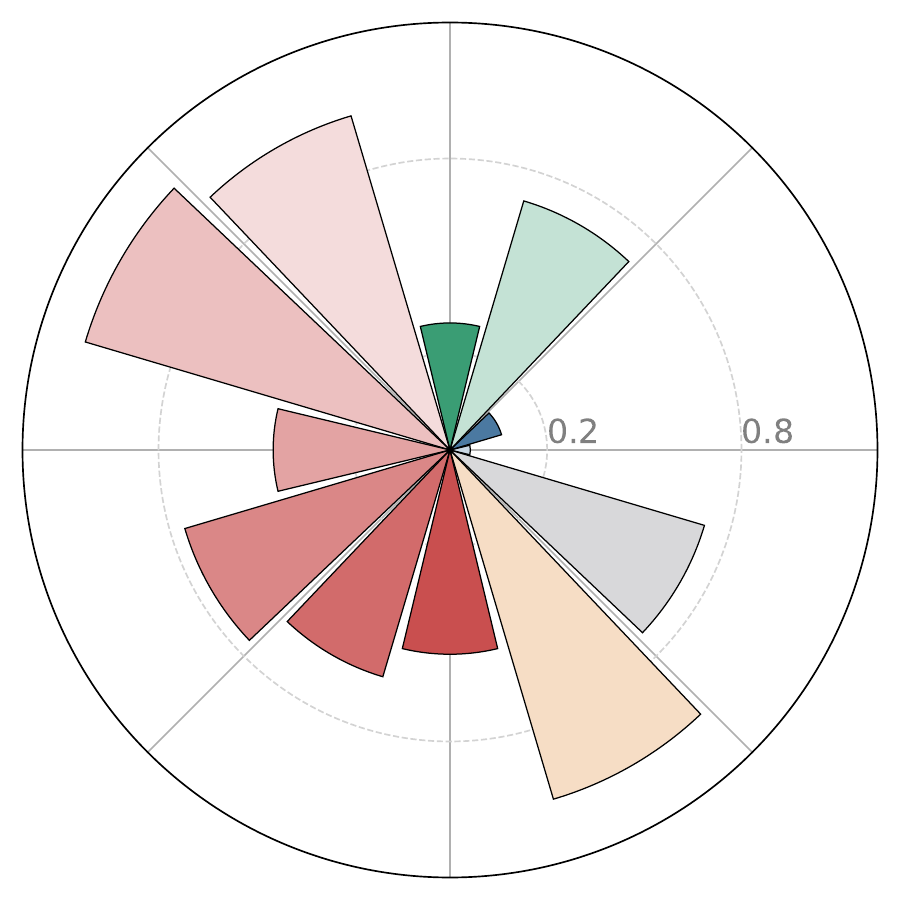} & 
\includegraphics[width=0.3\linewidth]{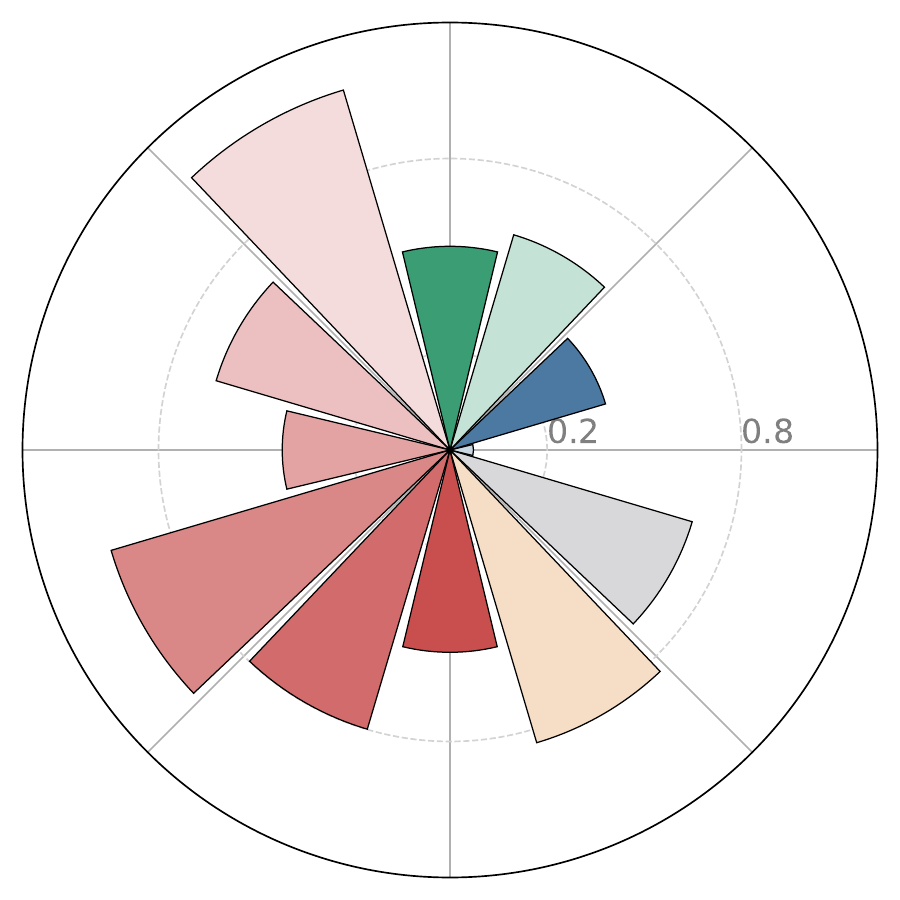} & 
\\
\midrule
\rowcolor{blue!3}\multicolumn{5}{c}{\myfont \textbf{Story Video Method \& Commercial Platform}}
\\
\midrule
\texttt{Vlogger (Text-only)} & 
\texttt{AnimDirector (SD3)} & 
\texttt{MMStoryAgent} & 
\texttt{MovieAgent (SD3)} & 
\texttt{MOKI} 
\\ 
\includegraphics[width=0.3\linewidth]{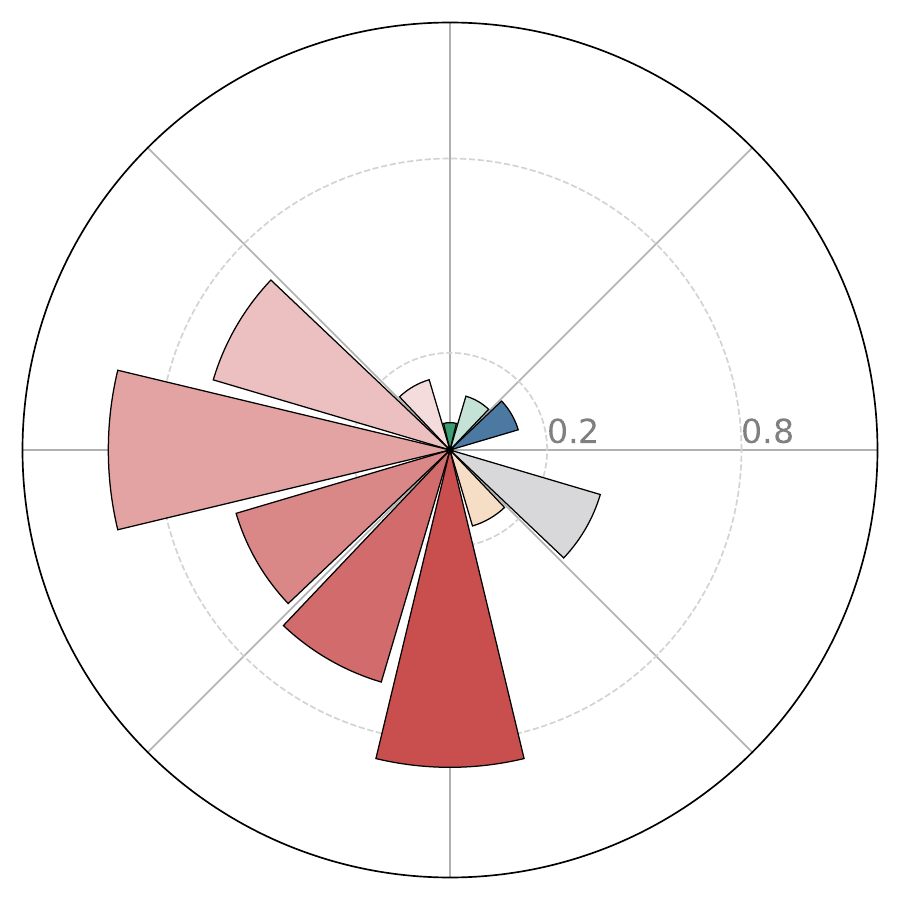} & 
\includegraphics[width=0.3\linewidth]{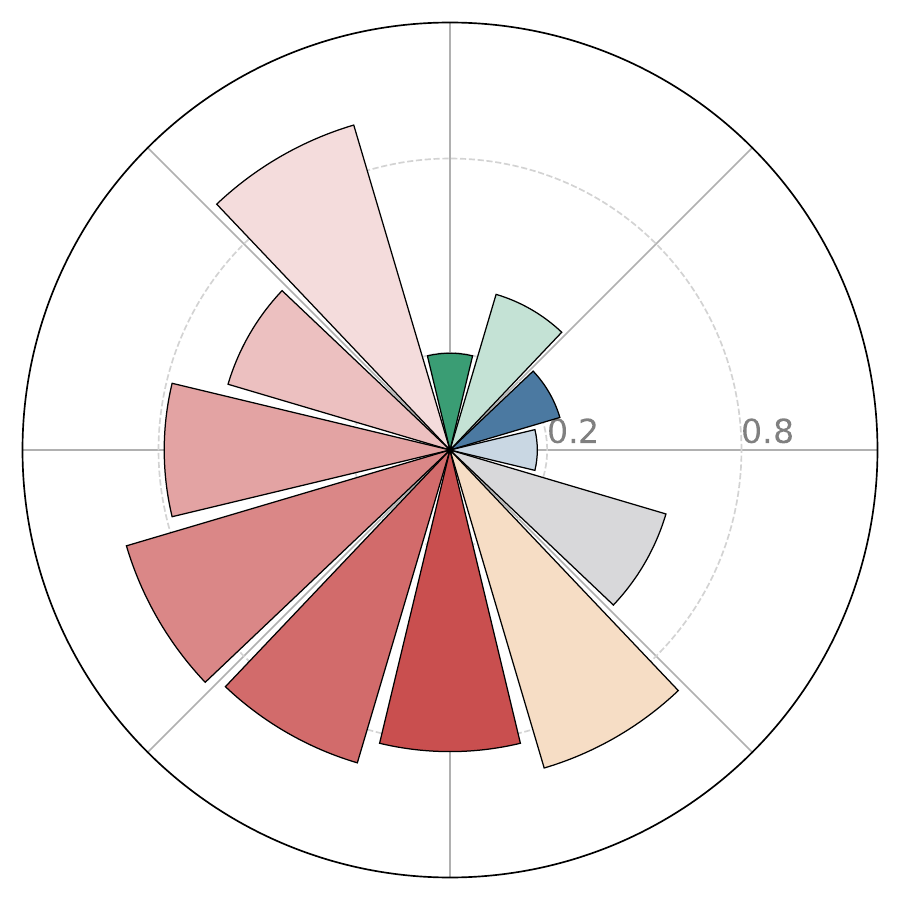} & 
\includegraphics[width=0.3\linewidth]{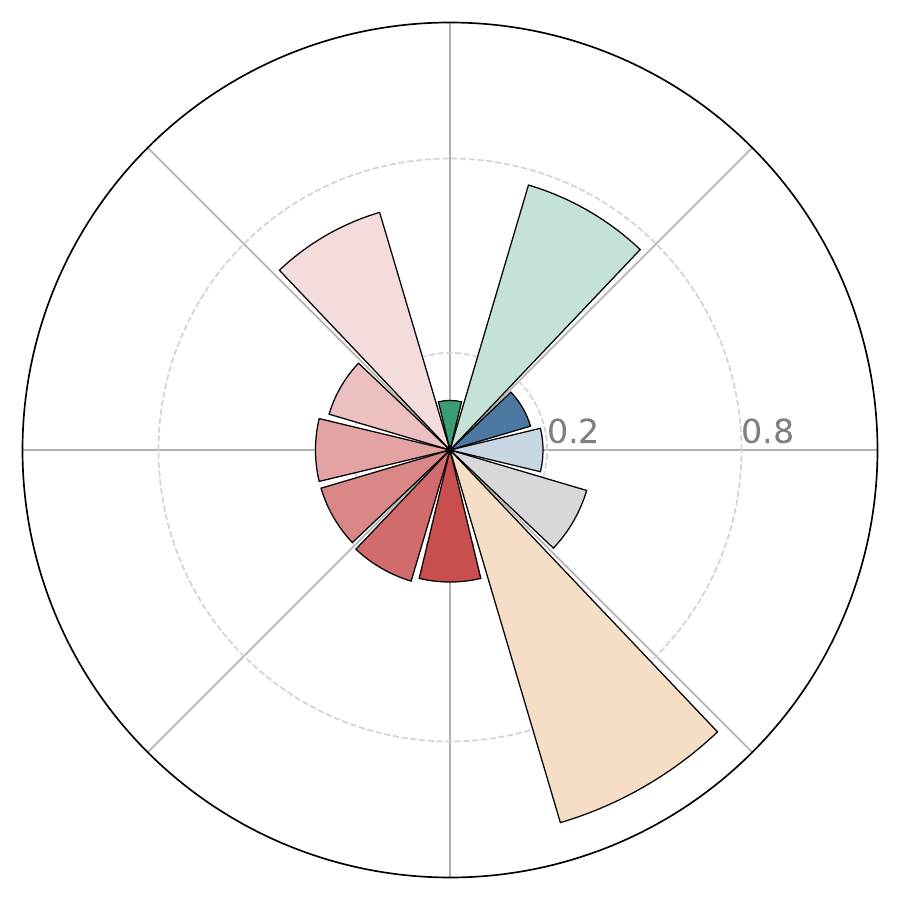} & 
\includegraphics[width=0.3\linewidth]{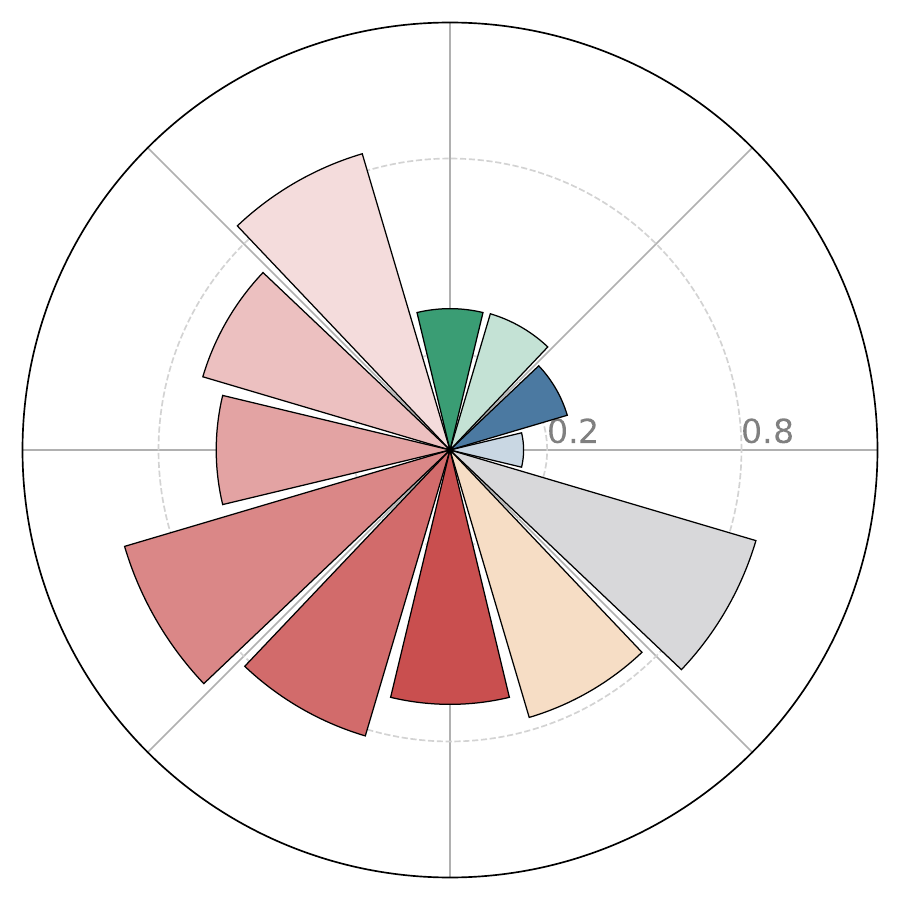} & 
\includegraphics[width=0.3\linewidth]{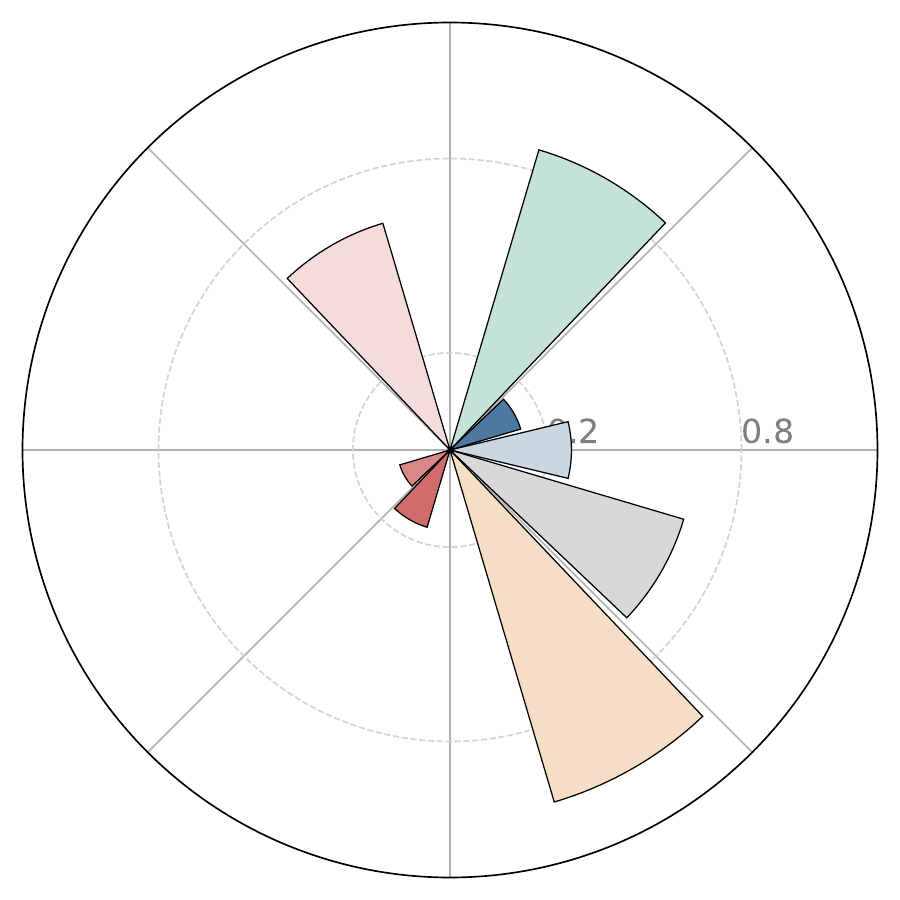} 
\\
\midrule
\texttt{MorphicStudio} & 
\texttt{AIbrm} & 
\texttt{ShenBi} & 
\texttt{TypeMovie} & 
\texttt{Doubao} 
\\ 
\includegraphics[width=0.3\linewidth]{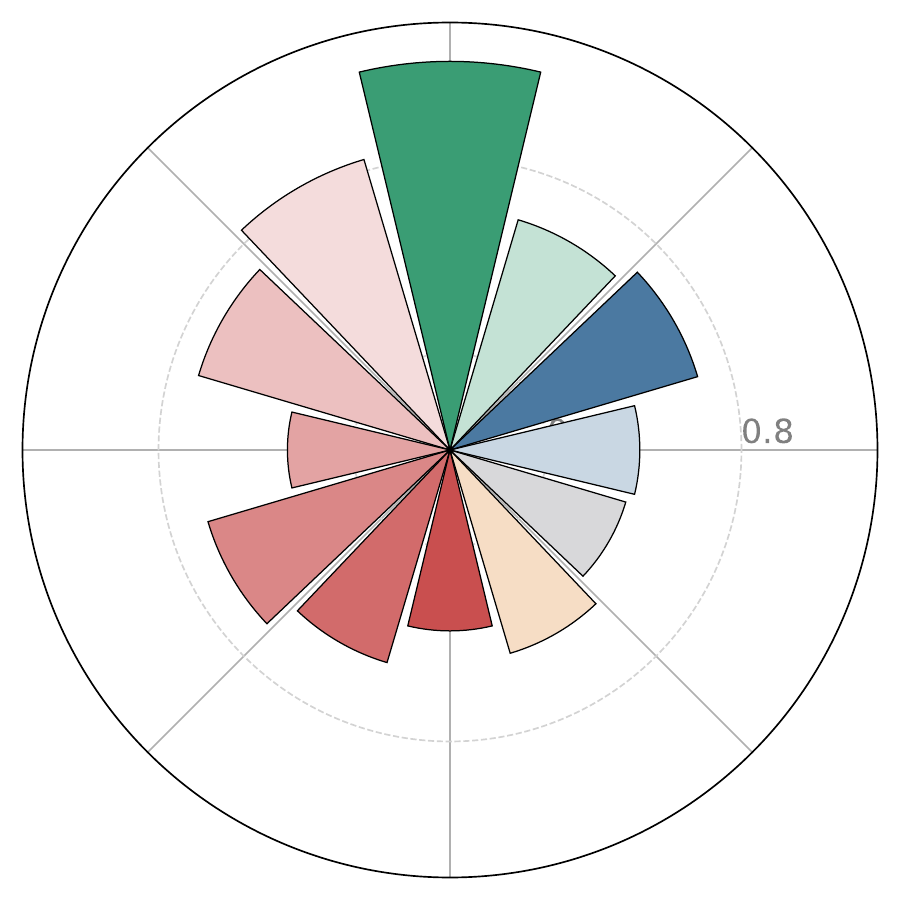} & 
\includegraphics[width=0.3\linewidth]{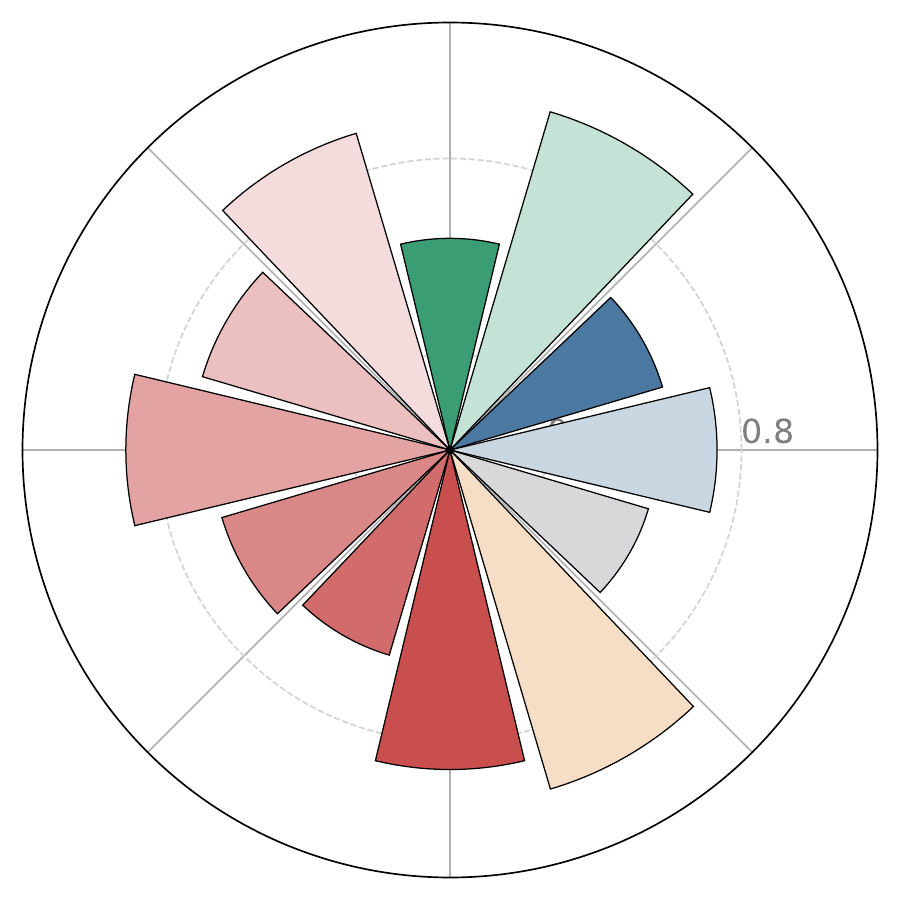} & 
\includegraphics[width=0.3\linewidth]{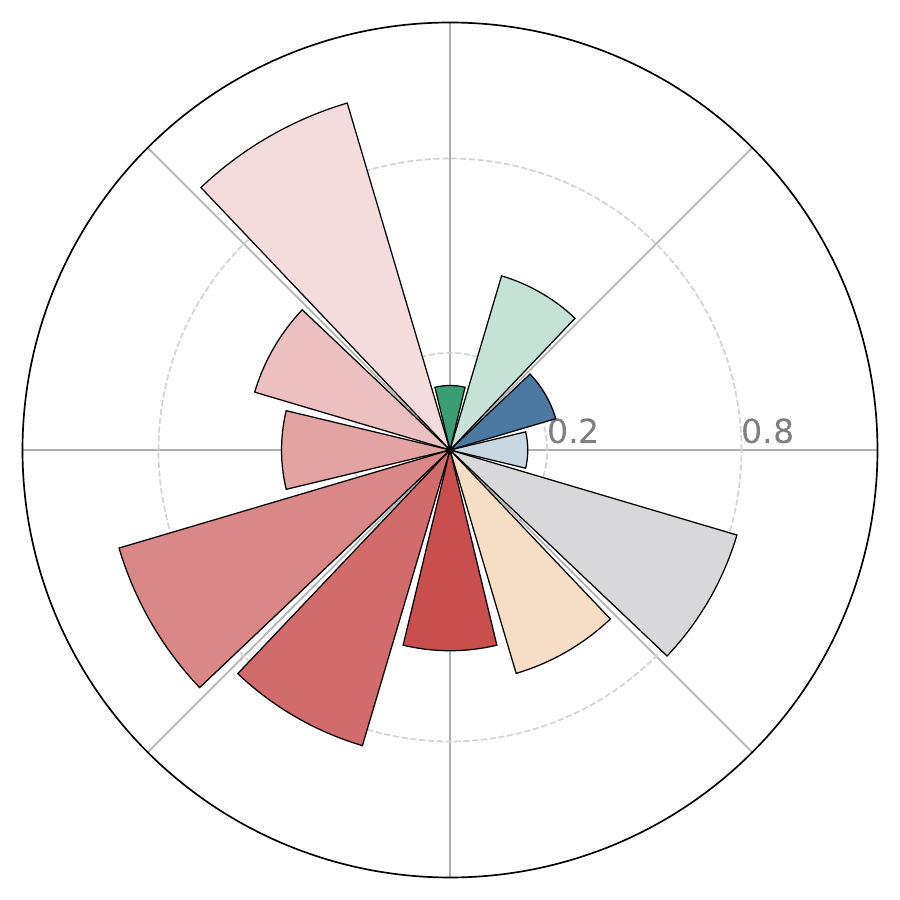} & 
\includegraphics[width=0.3\linewidth]{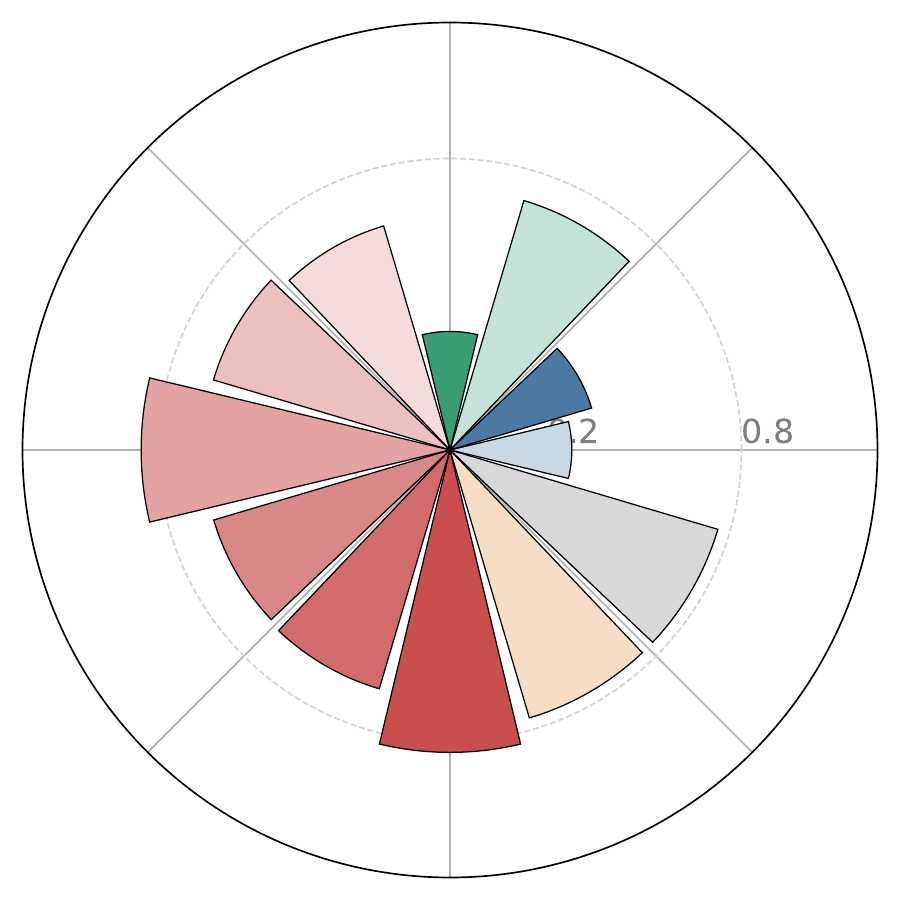} & 
\includegraphics[width=0.3\linewidth]{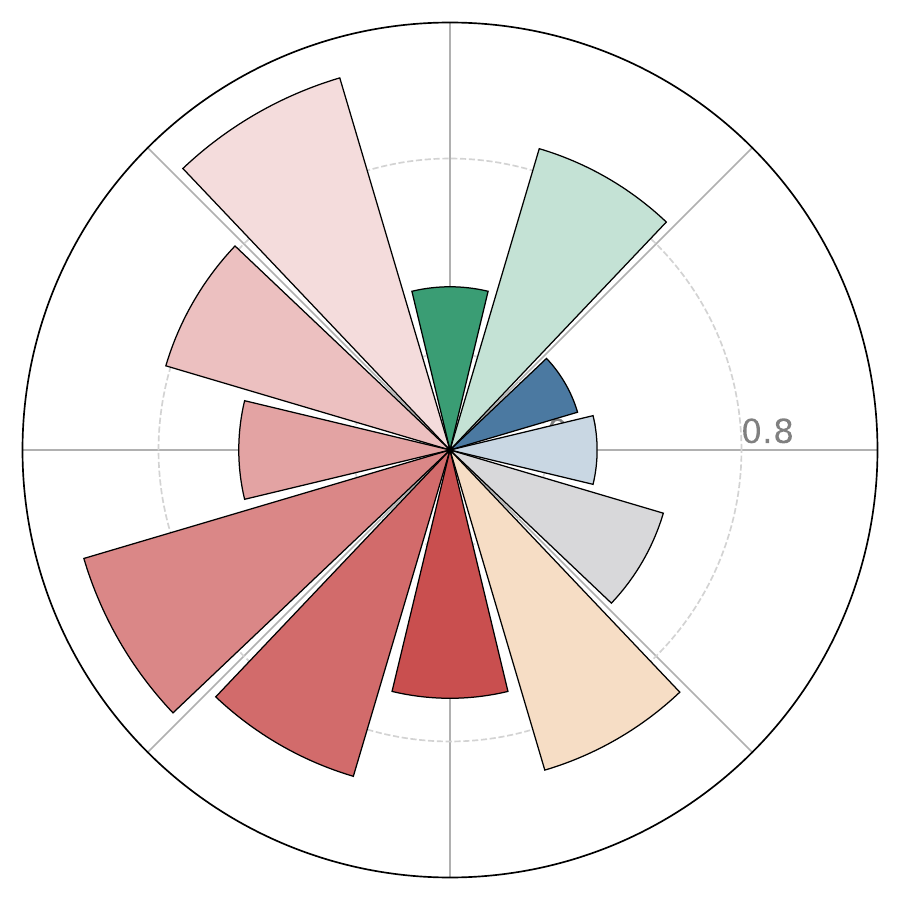} 
\\
\midrule
\rowcolor{blue!3}\multicolumn{5}{c}{\myfont \textbf{Multi-modal Large Model}}
\\     
\midrule
\texttt{GPT-4o} & 
\texttt{Gemini-2.0} & 
\texttt{Gemini-3 (NanoBananaPro)} & 
\texttt{Seedream-4.0} & 
\texttt{Sora2 (Img-ref)} 
\\
\includegraphics[width=0.3\linewidth]{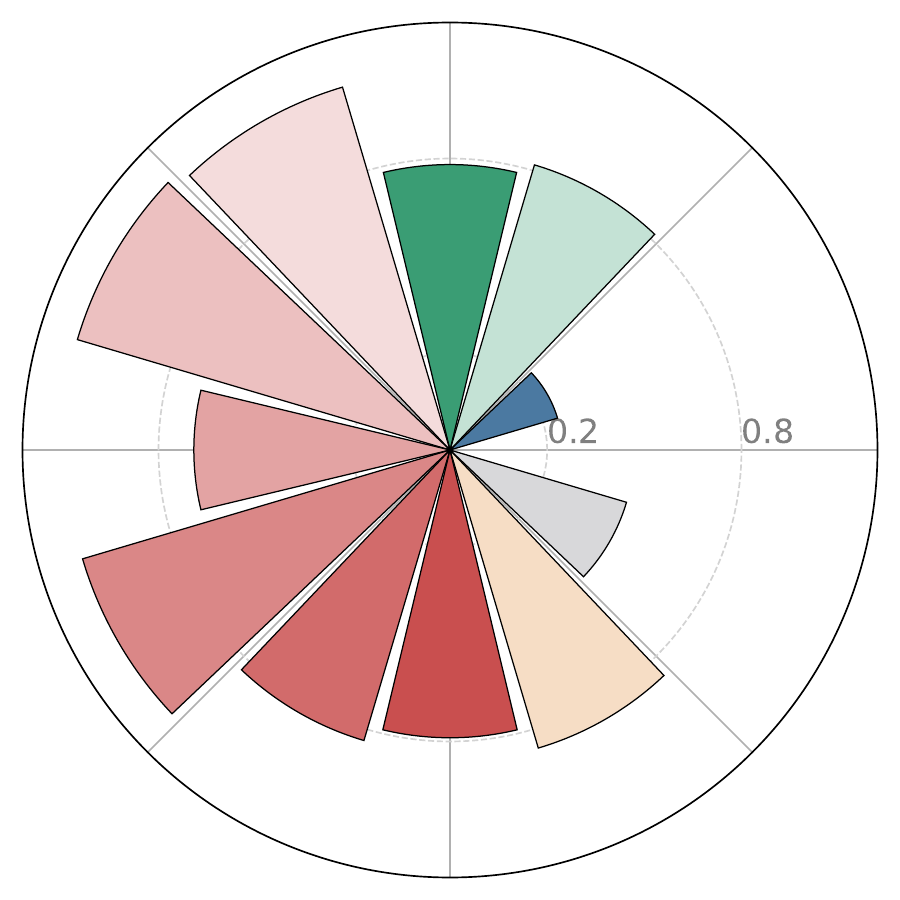} & 
\includegraphics[width=0.3\linewidth]{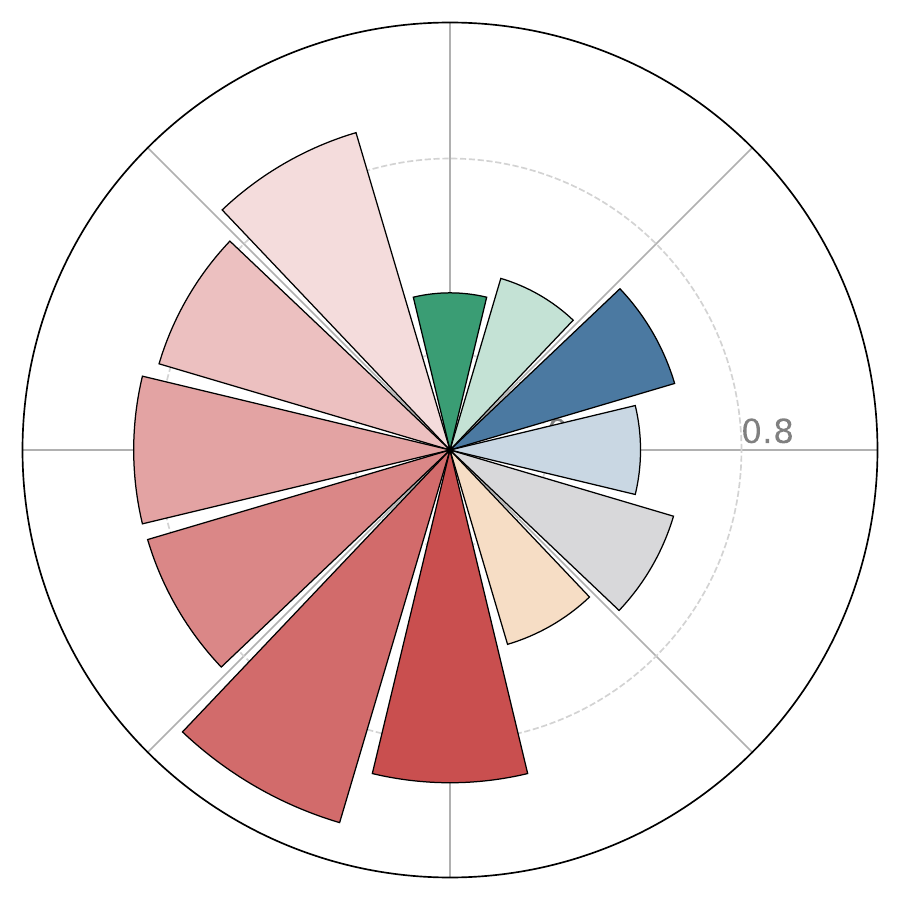} & 
\includegraphics[width=0.3\linewidth]{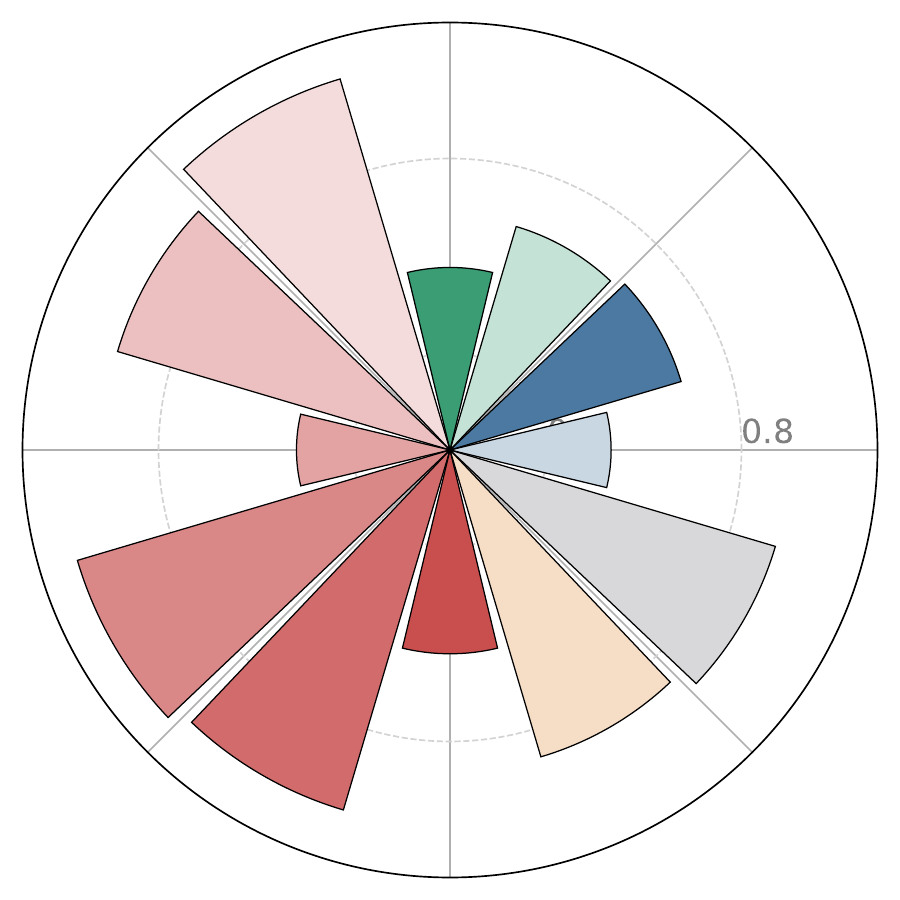} & 
\includegraphics[width=0.3\linewidth]{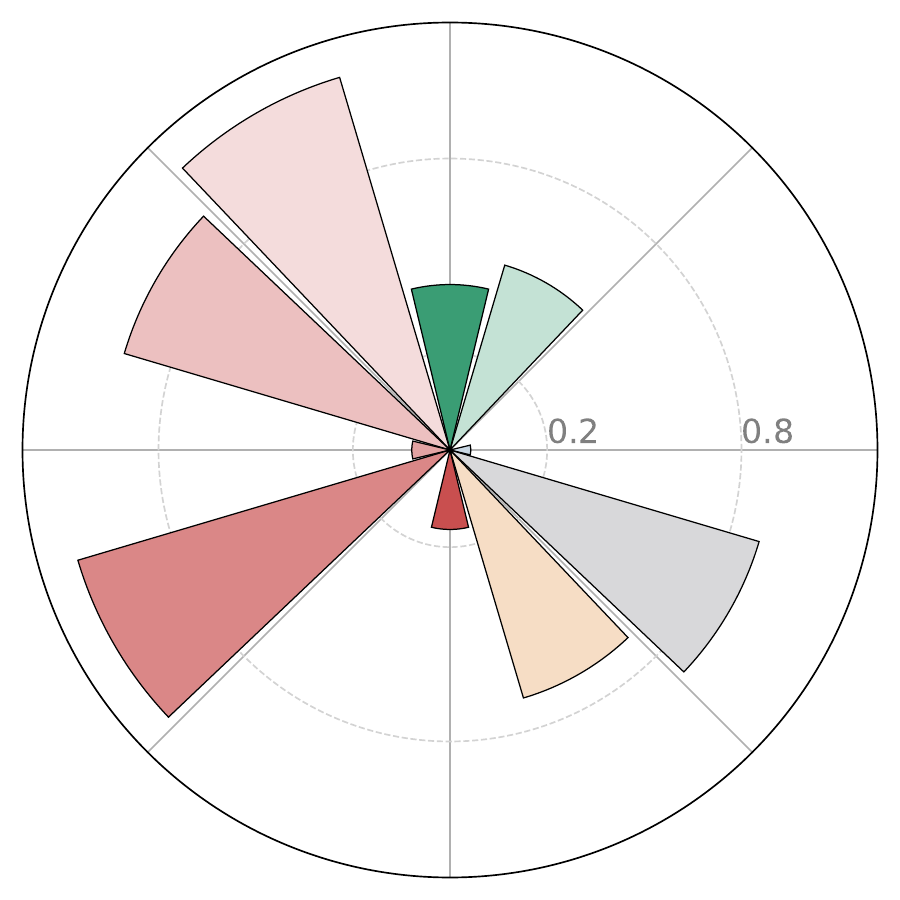} & 
\includegraphics[width=0.3\linewidth]{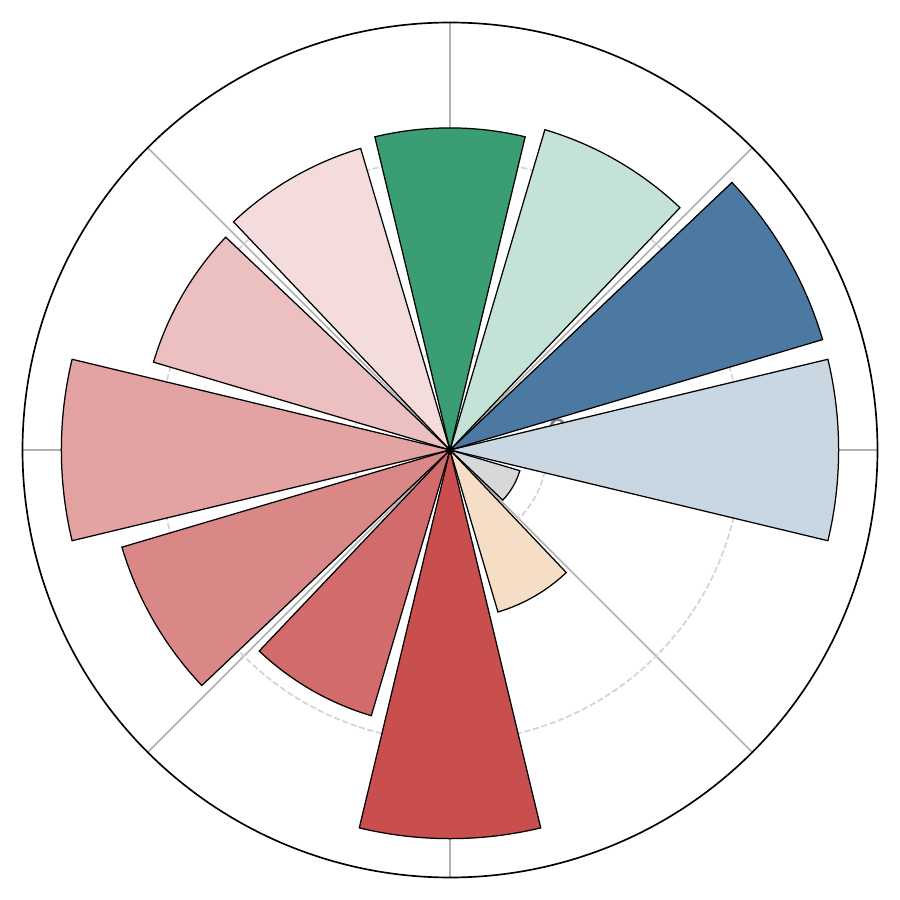}

\\
\bottomrule
\end{tabular}
}
\label{tab:polar}
\end{table*}

We present the results of several selected methods on the ViStoryBench-Lite benchmark in the main paper. To facilitate a more comprehensive comparison, we provide the complete evaluation results for all the methods on the lite set, as shown in Table~\ref{tab:comparison_lite}. These results serve as a thorough reference for assessing the performance of different approaches from multiple perspectives. Additionally, to ensure reproducibility, we provide the complete prompt alignment evaluation results using \textbf{Qwen3-VL-8B-Instruct}, deployed via vLLM offline with a fixed seed of 42, as shown in Table~\ref{tab:comparison_full_lite_pa_qwen}.

To further interpret the results, we visualize the normalized performance of each evaluated SOTA method across all twelve dimensions using polar coordinates in Table~\ref{tab:polar}. Each radius is normalized relative to the best score achieved among all models for the corresponding dimension. 

\paragraph{Performance Analysis of Various Methods.}

On ViStoryBench-Lite, we observe a wide spectrum of performance among the evaluated methods. 

As evidenced by the quantitative results in the tables, \textbf{GPT-4o}~\cite{hurst2024gpt4o} and \textbf{Gemini-2.5}~\cite{comanici2025gemini2.5} demonstrate superior prompt alignment capabilities, which can be attributed to their fine-grained comprehension abilities rooted in their LLM foundations. Meanwhile, \textbf{Sora2}~\cite{2025sora2} achieves the best balance across the four key metrics of character consistency (cross and self) and style consistency (cross and self), likely benefiting from its extensive training on visually coherent multi-shot data. The performance gap between earlier methods (e.g., \textbf{StoryGen}~\cite{Liu_2024_CVPR_Storygen}, \textbf{TheaterGen}~\cite{cheng2024theatergen}, \textbf{Vlogger}~\cite{zhuang2024vlogger}) and more recent approaches (e.g., \textbf{UNO}~\cite{wu2025uno}, \textbf{OmniGen2}~\cite{wu2025omnigen2}) reflects the natural progression and collective advancement within the research community. 

Models such as \textbf{Storydiffusion}~\cite{zhou2024storydiffusion} and \textbf{Story-Adapter}~\cite{mao2024story_adapter} exhibit strong prompt alignment, while maintaining balanced performance in generation quality and character similarity. \textbf{MovieAgent (SD-3)}~\cite{wu2025movieagent} and \textbf{AnimDirector}~\cite{li2024anim} achieve consistently high scores across most dimensions, with particularly notable strengths in aesthetics and diversity, indicating their advantages in generating visually appealing and varied outputs. Interestingly, \textbf{SEED-Story}~\cite{yang2024seedstory}, which focuses on story continuation tasks, exhibits excellent self-consistency in style but shows weaker performance in other metrics. This observation highlights a potential trade-off between maintaining visual fidelity and introducing meaningful variation in generated content.

Among the commercial and proprietary systems, we observe competitive and often leading performance.  \textbf{AIbrm}~\cite{brmgo2025} and \textbf{Doubao}~\cite{doubao2024}, two commercial Chinese platforms, show highly stable performance across almost all metrics. \textbf{Doubao}, in particular, demonstrates outstanding results in both prompt alignment and generation quality, with a strong balance between character rendering fidelity and stylistic coherence. These results suggest that closed-source commercial tools have made significant strides in multi-shot storytelling, although their internal pipelines remain opaque.

Overall, these results highlight the multidimensional nature of high-quality story visualization and suggest that no single model excels uniformly across all criteria.
These quantitative findings align consistently with our qualitative visual observations and common sense, thereby validating the credibility and reliability of our proposed evaluation metrics for story visualization tasks.


\section{Method Evaluation Detail on ViStoryBench}\label{sec:implementation_detail}
In this section, we report how we adapt each method to the ViStoryBench test. In general, we strive to implement reasonable inputs for reference character images and shot script prompts on each work as much as possible. We make efforts to standardize inputs, such as adjusting output resolution to a 16:9 aspect ratio whenever feasible. To guarantee reproducibility, we fix the random seed. Additionally, we implement mechanisms for inputting reference images and adapting lengthy shot script prompts. These adaptations enable methods to generate continuous image results of stories.

\begin{figure*}[t]
    \centering
    \includegraphics[width=1\linewidth]{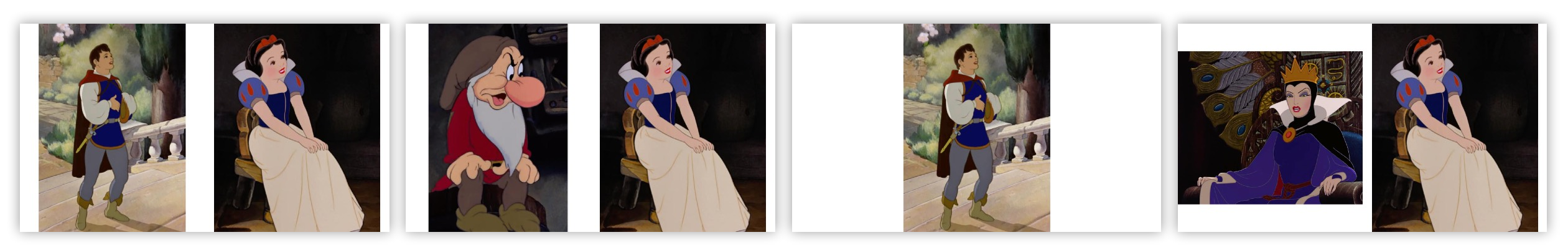}
    \caption{\small \textbf{Sample Results of Copy-Paste Baseline}.}
    \label{fig:copy_paste_baseline}
\end{figure*}
\subsection{Copy-Paste Baseline}
The Copy-Paste Baseline aims to verify the correctness of certain metrics by constructing the simplest possible generation method, which involves directly copying and pasting characters into images. For example, this baseline demonstrates that the metrics in ViStoryBench can perform an accurate calculation in areas such as CIDS, CSD, Copy-paste rate, and Matched Character Count calculation.

We obtain the image for the current shot by simply stitching together the images of the onstage characters in each shot, resulting in the image outcome for the current shot. A sample is shown in Figure~\ref{fig:copy_paste_baseline}.

\subsection{Story Image Methods}\label{sec:story_image_method}

\subsubsection{StoryDiffusion}
The original StoryDiffusion~\cite{zhou2024storydiffusion} is primarily designed for image generation using multiple reference images of different characters and multi-shot prompts, where each prompt includes a character name and description. We implement the following adaptations for ViStoryBench evaluation:
\begin{enumerate}[label=\textcircled{\scriptsize\arabic*}, leftmargin=1.4em, parsep=-0.05em]
    \item Since the native implementation can only handle short texts that cannot exceed the model's maximum sequence length limit (77 tokens), similarly to StoryGen adaption, we employ the grouped encoder sd\_embed~\cite{sd_embed_2024} to address this issue.
    \item Given StoryDiffusion does not support inputting reference images of multiple characters while generating one shot, we sort characters based on their appearance frequency in the full shot script, from highest to lowest, and only introduced the highest-priority characters in one shot.
    \item StoryDiffusion predefines various style templates, each containing detailed style descriptions, including quality words. We insert the character prompt into the specified position within the style description, i.e., between the style word and the quality word, to achieve richer style expression.
    \item During testing, we obtain two types of results: those based on image reference (img ref) and those based solely on text (text only). We report the metrics for both types of results separately.
\end{enumerate}

\subsubsection{Story-Adapter}
The original Story-Adapter~\cite{mao2024story_adapter} generates multiple images through multiple prompts and does not inherently support input image references. We implement the following adaptations for ViStoryBench evaluation:
\begin{enumerate}[label=\textcircled{\scriptsize\arabic*}, leftmargin=1.4em, parsep=-0.05em]
    
\item Incorporating Reference Images: Story-Adapter performs multiple rounds of iterative inference. In the first round, it generates all the storyboard images, which are then used as image references for regeneration in subsequent rounds. This multi-round process enhances both prompt alignment and style consistency of the images. We leverage this inherent image-referencing capability of Story-Adapter by inputting the image of the current onstage character into the pipeline during the first round of generation, thereby achieving image generation with reference images.

\item We test the results of different iteration rounds (scale 0 and scale 5) under two modes: text-only and image-Ref, and reported the results of all four methods.

\end{enumerate}

\subsubsection{UNO}
UNO~\cite{wu2025uno}'s original implementation supports various reasoning methods such as one2one, one2many, many2one, and many2many. Among these, the many2many approach involves inputting multiple images along with prompts that provide simple descriptions of the images. We exclusively evaluate the many2many method, which is capable of completing the ViStoryBench task. For each shot, we generate the corresponding image result by inputting the shot's prompt and the images of the onstage characters.

\subsubsection{OmniGen2}
As a general-purpose image generation model, OmniGen2 is evaluated for its visual consistency capability.  
Scene, plot, action, and shot design descriptions are concatenated into a prompt. The first reference image of every character appearing in the shot is supplied as a visual reference. Each shot is generated individually. The previous shot’s image is excluded from the context, as empirical tests indicated performance degradation when including such references.

\subsubsection{CharaConsist}
CharacterConsist enables training-free, cross-shot character identity ``locking''. We evaluate its performance on key metrics such as CIDS\_Self. 
\begin{enumerate}[label=\textcircled{\scriptsize\arabic*}, leftmargin=1.4em, parsep=-0.05em]
    \item ``Setting Description'' serves as the background prompt, ``Character Description'' as the foreground prompt, and ``Static Shot Description'' + ``Shot Perspective Design'' as the action prompt. Since CharacterConsist does not support explicit character references, no reference images are injected.  
    \item The most frequently appearing character in each story is selected as the protagonist, with ``prompt\_en'' used as the unified foreground character descriptor across all shots.  
    \item The first shot is generated as the identity reference. Subsequent frames reuse the ``id\_fg\_mask'' and ``id\_bg\_mask'' from the first frame. Each generation first performs a pre-run (``is\_pre\_run''=True) before formal generation. Spatial parameters (``spatial\_kwargs'') are propagated throughout to ensure character and background consistency across shots.
\end{enumerate}
    
\subsubsection{QwenImageEdit-2509}
This model is specifically designed for multi-image consistency and is evaluated for its visual coherence capability.  
Scene, plot, action, and shot design descriptions are concatenated as the prompt. The first reference image of each character in the shot is provided as a visual reference. Each shot is generated sequentially. The image from the previous shot is omitted from the input, as tests demonstrated that including it led to output degradation.

\subsubsection{StoryGen}
The original StoryGen~\cite{Liu_2024_CVPR_Storygen} supports the integration of previously generated image-text pairs as a context to construct image sequences that align with the input shot scripts. We implement the following adaptations for ViStoryBench evaluation:

\begin{enumerate}[label=\textcircled{\scriptsize\arabic*}, leftmargin=1.4em, parsep=-0.05em]
    \item Mix Inference Method: The mix method behaves similarly to the multi-image-condition method for the first frame, while subsequent frame generation follows the auto-regressive method.
    
    \item When image references are absent (e.g., no onstage characters), the first shot uses aggregated story-wide character descriptions and images for consistent initial representation. For subsequent shots, a dynamic sliding window blends historical generation results as input, maintaining temporal coherence and mitigating quality degradation from long-range dependencies.

    \item Grouped Encoder for Ultra Long Prompts: To handle ultra-long prompts and prevent information loss due to truncation in CLIP models, we employ a grouped encoder sd\_embed~\cite{sd_embed_2024} to address this issue.
    \item Resolution Adjustment: We modify the resolution settings to approximate a 9:16 aspect ratio, specifically $512\times912$, to align with the generation resolutions of other methods.
    \item Deterministic Random Seed Strategy: To enhance test controllability, we adopt a deterministic random seed strategy.
\end{enumerate}

\subsubsection{TheaterGen}
The original TheaterGen~\cite{cheng2024theatergen} only supports text input, where an LLM is used to parse the description of each character and the overall image of the text. Subsequently, each character's independent image is generated through the IP-Adapter, and then these images are placed in their corresponding positions using detection and segmentation models. Finally, the final image is generated under the guidance of the character images. 

Since the method does not open-source the code for invoking the LLM part, we supplement the corresponding code to primarily obtain the bounding box coordinates of each IP image required for model input. Afterward, following TheaterGen's setup, we place the character reference images in our dataset into the corresponding positions to generate final results.

\subsubsection{SEED-Story}
The original SEED-Story~\cite{yang2024seedstory} focus on generating long multi-modal stories and their visualization through an auto-regressive approach. SEED-Story requires the user to provide an initial image and text description, then proceeds with the story generation, using each output as the next input. Due to the story-continuation nature, it is different from other approaches (SEED-Story was only trained on the StoryStream dataset~\cite{yang2024seedstory}, which contains data from only three cartoon series, the model does not have generalization).

Input Handling: We only adapt our data in the visualization stage, and there is no need for the previous story generation. For the first input image (000start), select the first reference image of the first character. In the pre-visualization input data, add the prompt words of this role to the beginning of the prompt list to ensure the first shot's prompts are not compromised.

\subsection{Story Video Method}\label{sec:story_video_method}

\subsubsection{MovieAgent}
Similar to Vlogger, MovieAgent~\cite{wu2025movieagent} also receives a story and utilizes an LLM to generate a series of fine-grained descriptions. Additionally, ROIctrl includes the bounding boxes of characters. We directly map shot prompts onto Vlogger's fine-grained descriptions for generation. We only perform the step from prompt to image, without further generating a video, and use this image as the final result.

\subsubsection{AnimDirector}
AnimDirector~\cite{li2024anim} does not support input reference images. It utilizes LLM to supplement character and scene descriptions based on simple prompts provided by users, refining them into complete story sequences by scenes. Visual images are generated through the Stable Diffusion 3~\cite{esser2024scaling} model, filtered by VLM, and subsequently used to generate videos. We adopt the following strategies to complete the test on ViStoryBench:

\begin{enumerate}[label=\textcircled{\scriptsize\arabic*}, leftmargin=1.4em, parsep=-0.05em]
    \item We directly input the prompt of each shot from ViStoryBench as the story sequence prompt to generate images. After filtering by VLM, we obtain the results without proceeding with the subsequent video generation steps.
  
    \item To accommodate ultra-long prompts, we employ grouped encoder sd\_embed~\cite{sd_embed_2024} to address this issue.
  
    \item We modify the original resolution of $1024\times1024$ to a resolution close to 16:9, specifically $768\times1344$.
  
    \item We fix the random seed to ensure reproducibility.
\end{enumerate}

\subsubsection{MMStoryAgent}
MMStoryAgent~\cite{xu2024mmstoryagent} does not support the input of reference images. We only utilize the Image Agent mentioned in the code to generate images from prompts, without further generating a video. As MMStoryAgent is based on StoryDiffusion, we also employ grouped encoder sd\_embed~\cite{sd_embed_2024} to adapt to long prompts.

\subsubsection{Vlogger}
The original Vlogger~\cite{zhuang2024vlogger} takes a short story and utilizes an LLM to generate a series of fine-grained descriptions. These descriptions encompass character and object descriptions, video script descriptions (in both Chinese and English), characters and objects present, duration settings, etc. We directly map shot prompts onto Vlogger's fine-grained descriptions, selecting the first frame of each shot from the generated video as the result.
Similar to StoryDiffusion, vlogger does not support inputting reference images of multiple characters while generating one shot. We adopt a similar solution to that of StoryDiffusion.

\subsection{Commercial Software}\label{sec:commercial_method}
Given the absence of API or similar call methods for the following commercial softwares and their intricate interaction processes, we employ external annotators
to generate image results for all the commercial software mentioned below.
The annotators adhered to a predefined collection instruction and protocol, and all generated results subsequently underwent a rigorous quality check by the authors.
All the following methods were tested between May 1 and May 7, 2025.

\subsubsection{MOKI}
When generating shots on MOKI~\cite{moki2024}, it is necessary to select one of the provided painting style options. To best replicate the effects of real human usage, we instruct the annotators to choose the option that most closely aligns with the style of the character reference images for each script.

Given that MOKI restricts the maximum number of reference characters in a story to three, we sort the characters based on their appearance frequency among all shots. The three characters with the highest frequency were then added as reference characters to maximize the performance of the model.

We generate images of MOKI using Chinese version dataset. MOKI has a limit on the length of each shot's prompt, capped at 60 Chinese characters. Consequently, we make certain omissions in the prompt for each shot. For instance, we only used the Static Shot Description and Plot Correspondence sections as input. If the character count still exceeded the limit, we employ an LLM to abbreviate the text while preserving key information.

For each shot, the platform generated four images at a time. We select the first image as the final result.

\subsubsection{MorphicStudio}

MorphicStudio~\cite{morphic2024studio} restricts each shot to include only one reference character. Consequently, we rank the frequency of character appearances in the script and use this order as a priority to select the current onstage reference character for each shot.

MorphicStudio allows uploading 2 to 10 images for the same reference character. For characters with only one reference image, we upload an additional identical image to meet the minimum requirement of two images. For characters with multiple reference images, we upload all available reference images. Therefore, we do not provide the Copy-Paste Rate for MorphicStudio because all reference images were uploaded and used for generation, making it incalculable.

For each shot, the platform generates four images at a time. We select the first image as the final result.

\subsubsection{AIbrm}

AIbrm~\cite{brmgo2025} restricts each shot to a maximum of two reference characters. We employ the same sorting methodology as utilized in MorphicStudio.

The character creation process in AIbrm involves uploading a real human image, selecting a style from the provided options, and inputting a character prompt to generate the character. For the real human category in our dataset, we upload images and chose the "realistic" style. For virtual human and non-human categories, since image uploads were not feasible, we select the closest available style. Subsequently, we input the character prompts from our dataset across all styles to complete the character creation.

\subsubsection{ShenBi}

Firstly, when creating a project, ShenBi~\cite{shenbi2025} requires selecting a generation style from a list. We choose the style that is closest to the reference images in the dataset.

When creating characters, we upload the character images along with their corresponding prompts to generate new character images, which are subsequently utilized within the method to produce shot images.

ShenBi restricts each shot to a maximum of three reference characters. We employ the same sorting methodology as utilized in MorphicStudio.

The output of ShenBi is in the form of videos. We extract the first frame of each scene video as the final result.

\subsubsection{Typemovie}

Firstly, when creating a project, Typemovie~\cite{typemovie2024} requires selecting a generation style from a list. We choose the style that is closest to the reference images in the dataset.

The character creation process in Typemovie involves uploading a real human image, selecting a style from the provided options, and inputting a character prompt to generate the character. For the real human category in our dataset, we simply upload reference images. For virtual human and non-human categories, since image uploads were not feasible, we select the closest available style. Subsequently, we input the character prompts from our dataset across all styles to complete the character creation.

Typemovie restricts each shot to include only one reference character. We employ the same sorting methodology as utilized in MorphicStudio.

\subsubsection{Doubao}
We conduct our tests using the grayscale test version of the "Image Generation" model on the Doubao homepage~\cite{doubao2024}, dated April 27, 2025. 

Since this image generation model only supports uploading a single image, we employ a sorting method similar to that used in MorphicStudio to prioritize characters and selected the highest-priority character to upload as a single reference image. 

The prompt used during generation was (translated to English): "This is an image of the protagonist \textless character\_name\textgreater. Next, please generate storyboard scenes based on the protagonist's image and the script I provide. The script for the first scene is as follows: \textless shot\_prompt\textgreater." We perform multiple rounds of generation in one session to obtain the desired consistent image results for one story.

\subsection{Multi-modal Large Models}\label{sec:large_model}

To integrate Multi-modal Large Models into the ViStoryBench evaluation framework, we adopt the following key adaptation strategies:

\begin{enumerate}[label=\textcircled{\scriptsize\arabic*}, leftmargin=1.4em, parsep=-0.05em]
    \item \textbf{Atomic Shot Processing:} Each ``shot'' defined within ViStoryBench was treated as an independent generation request to the model, ensuring focused processing for individual narrative segments.
    \item \textbf{Comprehensive Prompt Engineering:} For every shot, a structured textual prompt was meticulously crafted. This prompt amalgamated all critical textual information provided by ViStoryBench, including plot details, scene descriptions, character portrayals, camera perspective guidelines, and the desired aspect ratio for the output image.
    \item \textbf{Direct Visual Referencing:} Character reference images, after undergoing a standardization pre-processing pipeline (e.g., resizing, color space conversion), were directly incorporated as visual inputs for each shot's generation request. This aimed to guide the model in rendering characters consistent with their specified appearances.
    \item \textbf{Conversational Context Continuation:} To foster narrative coherence across sequential shots, the model's inherent capability to process conversational history was leveraged. The most recent interaction cycles, encompassing the prompt for the preceding shot and the model's response (which includes the generated image), served as contextual information for the subsequent shot's generation task.
\end{enumerate}

\subsubsection{GPT-4o}
OpenAI's GPT-4o~\cite{hurst2024gpt4o} represents a general-purpose multimodal understanding and generation model. We evaluate whether its prompt alignment capability directly translates to high-quality visual storytelling.  

Scene, plot, action, and shot design descriptions are concatenated into a unified prompt. The image from the previous shot and the first reference image of all characters appearing in the current shot are used as visual references. Each shot is generated sequentially.

\subsubsection{Gemini-2.0}
Google's Gemini-2.0~\cite{gemini2.02025} represents a general-purpose multimodal understanding and generation model. We evaluate whether its prompt alignment capability directly translates to high-quality visual storytelling.  

Scene, plot, action, and shot design descriptions are concatenated into a unified prompt. The image from the previous shot and the first reference image of all characters appearing in the current shot are used as visual references. Each shot is generated sequentially.

\subsubsection{Gemini-2.5 (NanoBanana)} 
Gemini-2.5~\cite{comanici2025gemini2.5} represents a general-purpose multimodal understanding and generation model. We evaluate whether its prompt alignment capability directly translates to high-quality visual storytelling.  

Scene, plot, action, and shot design descriptions are concatenated into a unified prompt. The image from the previous shot and the first reference image of all characters appearing in the current shot are used as visual references. Each shot is generated sequentially.

\subsubsection{Seedream-4.0}
Seedream-4.0~\cite{gong2025seedream} represents a multimodal Image understanding and generation model. Scene, plot, action, and shot design descriptions are concatenated into a unified prompt. The image from the previous shot and the first reference image of all characters appearing in the current shot are used as visual references. Each shot is generated sequentially.

\subsubsection{Sora2}
Sora2~\cite{2025sora2} represents a long-video generation model with native multi-shot capability. We aim to evaluate its long-range narrative comprehension and multi-shot visual consistency.  
\begin{enumerate}[label=\textcircled{\scriptsize\arabic*}, leftmargin=1.4em, parsep=-0.05em]
    \item Scene, plot, action, and shot design descriptions for each shot are concatenated to form the input prompt.  

    \item In ``img\_ref'' mode, the first reference image of every character in the story is provided to define overall visual style and character baselines. In text\_only mode, image references are disabled.  
    
    \item TransNetV2~\cite{soucek2020transnetv2} is used for shot boundary detection and keyframe extraction, followed by manual verification to match keyframes to their corresponding shots.  
    
    \item Stories that fail to generate due to copyright restrictions or realistic portrait style limitations are excluded; only successfully generated stories are included in average metric calculations.
\end{enumerate}

\section{Details of User Study}\label{sec:user_study}

\begin{table*}[h!]
\centering
\resizebox{0.95\linewidth}{!}{
    
\begin{tcolorbox}[title=Scoring Criterial and Interface of User Study,
enhanced, 
skin first=enhanced,
skin middle=enhanced,
colback=yellow!10!white,
colframe=orange!80!black,
fonttitle=\bfseries]{}
    \textbf{Character Identification Consistency}: Based on the provided story visualization results, please assess the character id consistency of characters throughout the story and provide a score.
    
    Scoring Criteria:
    \begin{itemize}[itemsep=-1.4pt,leftmargin=2em]
        \item 0: There is a lack of fundamental ID consistency, with nearly every image featuring different characters, indicating an almost complete absence of images with matching characters.
        \item 1: In a smaller subset of images (about 10-30\%), the main characters demonstrate mutual consistency.
        \item 2: In a moderate number of images (around 30-60\%), the main characters can be treated as having mutual consistency.
        \item 3: In a substantial subset of images (approximately 60-80\%), the main characters exhibit mutual consistency. However, a minor portion of images still shows inconsistencies in character representation.
        \item 4: The main characters are consistently identifiable across the vast majority of images.
    \end{itemize}

    \textbf{Environment Consistency}: Based on the provided story visualization results, please assess the environment consistency throughout the story and provide a score.

    Scoring Criteria:
    \begin{itemize}[itemsep=-1.4pt,leftmargin=2em]
        \item 0: There is a lack of fundamental environmental consistency; under the same environmental description, the generated scenes exhibit neither consistent style nor content.
        \item 1: At a glance, the scenes appear to have some level of consistency, such as similar styles. However, upon closer inspection, the content is entirely different and lacks any coherence.
        \item 2: There is a certain level of consistency in the style and semantic information of the image scenes, such as the presence of similarly styled beds and windows. However, inconsistencies exist in either the style or specific content, for instance, while tables and desk lamps are present in both, the desk lamps themselves are not similar.
        \item 3: The majority of image have consistent semantic information, with style and specific content being largely uniform.
        \item 4: Nearly all image scenes exhibit strong consistency in both style and specific content, akin to the effect of video recording within the same scene over a continuous timeframe.
    \end{itemize}

    \textbf{Subjective Aesthetics Score}: Based on the provided results, please assess the aesthetics of the story and provide a score.

    \textbf{Scoring Criteria}:
    \begin{itemize}[itemsep=-1.4pt,leftmargin=2em]
        \item 0: Most characters have very obvious generation problems, such as distorted faces, extra/missing limbs, or the painting style is very uncomfortable for humans to watch. Or the image quality is extremely poor.
        \item 1: Characters have obvious generation problems, such as extra/missing limbs, distortion, etc., but there is no discomforting content. Or the image quality is poor.
        \item 2: Over 80\% of the characters have no obvious physical problems, and there is no obvious content that causes physical discomfort, but the visual experience is poor. Almost all the content of the images, such as character poses, is completely the same, lacking variation.
        \item 3: Over 80\% of the characters have no obvious physical problems. Mediocre picture books with ordinary visual experience, lacking variation and storytelling in images.
        \item 4: Over 80\% of the characters have no obvious physical problems. Excellent and beautiful picture books that can be commercialized, with rich content, beautiful details, diversity, and interest, and obvious storytelling. 
    \end{itemize}
    \centering
    \includegraphics[width=1\linewidth]{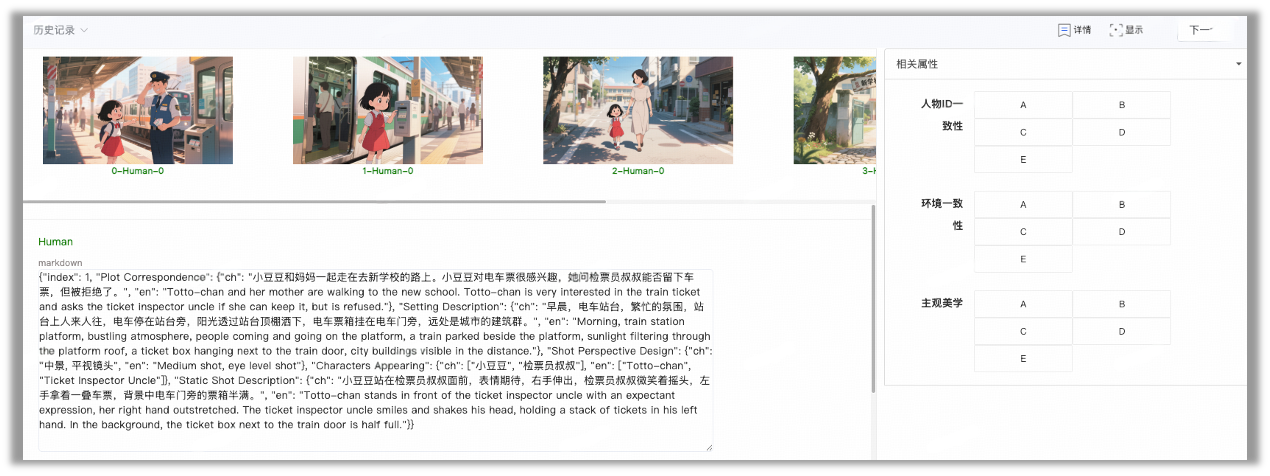}
\end{tcolorbox}
}
\end{table*}

\begin{table*}[h]
\caption{
    \small\textbf{Results of User Study}. For certain methods, we evaluate multiple inference configurations and report all corresponding results. \legendsquare{colorfirst} \legendsquare{colorsecond} \legendsquare{colorthird} \legendsquare{colorfourth} \legendsquare{colorfifth} indicate the first, second, third, fourth, and fifth performance, respectively. \imageref: With image reference; \textref: Only text input; \autoregressive: Auto-regressive mode; superscript $^k$ means scale=$k$.
    }
    \vspace{1mm}
    \centering
    \resizebox{0.7\textwidth}{!}{
    \begin{tabular}{l|c|ccc}
        \toprule

        \multirow{2}{*}{\textbf{Method}} &
        \multirow{2}{*}{\textbf{Model}} &
        \textbf{Character Identification}  & 
        \textbf{Environment}  & 
        \textbf{Subjective}  \\
        & & 
        \textbf{Consistency} $\uparrow$ & 
        \textbf{Consistency} $\uparrow$ & 
        \textbf{Aesthetics} $\uparrow$\\
        \midrule 

        \rowcolor{teaserdatasetback}\multicolumn{5}{c}{\myfont \textbf{Story Image Method}} \\ \midrule
        \rowcolor{teaserdatasetback}StoryGen~\cite{Liu_2024_CVPR_Storygen}~\autoregressive & SD1.5 & \textit{\underline{0.10}} & \textit{\underline{0.12}} & \textit{\underline{0.05}} \\ 
        \rowcolor{teaserdatasetback}StoryGen~\cite{Liu_2024_CVPR_Storygen}~\imageref & SD1.5 & 0.18 & 0.27 & 0.13 \\ \rowcolor{teaserdatasetback}StoryGen~\cite{Liu_2024_CVPR_Storygen}~\autoregressive\imageref & SD1.5 & 0.37 & 0.22 & 0.10 \\ 
        
        \rowcolor{teaserdatasetback}TheaterGen~\cite{cheng2024theatergen} & SD1.5 & 0.35 & 0.55& 0.30 \\
        \rowcolor{teaserdatasetback}StoryDiffusion~\cite{zhou2024storydiffusion}~\textref & SDXL & 2.72 & 2.73 & 2.45\\
        \rowcolor{teaserdatasetback}StoryDiffusion~\cite{zhou2024storydiffusion}~\imageref & SDXL & 2.62 & 2.33 & 2.30 \\
        
        \rowcolor{teaserdatasetback}SEED-Story~\cite{yang2024seedstory} & SDXL & 2.05 & 2.11 & 1.05 \\
        \rowcolor{teaserdatasetback}Story-Adapter~\cite{mao2024story_adapter}~\imageref$^0$ & SD1.5 & 2.55& 2.62&\rankfifth{2.80} \\ 
        \rowcolor{teaserdatasetback}Story-Adapter~\cite{mao2024story_adapter}~\imageref$^5$ & SD1.5 & 2.90& \rankfourth{2.98}& 2.68 \\
        \rowcolor{teaserdatasetback}Story-Adapter~\cite{mao2024story_adapter}~\textref$^0$ & SD1.5 & 2.33 & 2.23 & 2.50 \\ 
        \rowcolor{teaserdatasetback}Story-Adapter~\cite{mao2024story_adapter}~\textref$^5$ & SD1.5 & \rankfourth{3.10} & 2.67& 2.68 \\
        
        \rowcolor{teaserdatasetback}UNO~\cite{wu2025uno} & FLUX1 & \rankthird{3.20} & \rankfirst{\textbf{3.10}} & \rankfourth{3.02} \\

        \midrule
        \rowcolor{blue!3}\multicolumn{5}{c}{\myfont \textbf{Story Video Method}} \\ \midrule
        \rowcolor{blue!3}Vlogger~\cite{zhuang2024vlogger}~\textref  & SD1.4 & 0.87 & 1.07 & 0.67 \\
        \rowcolor{blue!3}Vlogger~\cite{zhuang2024vlogger}~\imageref & SD1.4 & 1.30 & 1.33 & 1.08 \\ 
        \rowcolor{blue!3}AnimDirector~\cite{li2024anim} & SD3 & 2.52 & 2.28 & 1.77 \\ 
        \rowcolor{blue!3}MMStoryAgent~\cite{xu2024mmstoryagent} & SDXL & 2.27 & 2.78 & 2.55 \\ 
        \rowcolor{blue!3}MovieAgent~\cite{wu2025movieagent} & SD1.5 & 1.90 & 1.95 & 1.55 \\
        \rowcolor{blue!3}MovieAgent~\cite{wu2025movieagent} & SD3 & 2.45 & 2.47 & 1.83 \\ 

        \midrule
        \rowcolor{SIMback}\multicolumn{5}{c}{\myfont \textbf{Commercial Platform}} \\ \midrule
        \rowcolor{SIMback}MOKI~\cite{moki2024} & - & 1.73&2.20&2.55 \\
        \rowcolor{SIMback}MorphicStudio~\cite{morphic2024studio} & - & 2.60&2.53&2.39 \\
        \rowcolor{SIMback}AIbrm~\cite{brmgo2025} & - & \ranksecond{3.42}&\rankfifth{2.97}&\rankthird{3.15} \\
        \rowcolor{SIMback}ShenBi~\cite{shenbi2025} & - & 2.74&2.89&2.48 \\
        \rowcolor{SIMback}Typemovie~\cite{typemovie2024} & - & 2.25&2.35&2.00 \\
        \rowcolor{SIMback}Doubao~\cite{doubao2024} & - & \rankfirst{\textbf{3.63}}&\rankthird{3.02}&\ranksecond{3.25} \\
        
        \midrule
        \rowcolor{red!5}\multicolumn{5}{c}{\myfont \textbf{Multi-modal Large Model (Language, Image and Video)}} \\ \midrule
        \rowcolor{red!5}GPT-4o$^*$~\cite{hurst2024gpt4o} & - & 3.08 & \ranksecond{3.06} & \rankfirst{\textbf{3.28}} \\
        \rowcolor{red!5}Gemini-2.0$^*$~\cite{gemini2.02025} & - & 2.84 & 2.84 & 2.26 \\

        \bottomrule
    \end{tabular}
    }
    \label{tab:comparison_user_study}
    \vspace{-1em}
\end{table*}

The user study component of our research was time-limited, concluding in mid-May 2025 due to project constraints. Its main purpose was to validate our automated evaluation metrics. As the user feedback showed a strong correlation with the automated scores, we confirmed the reliability of the automated method. Therefore, subsequent model evaluations were predominantly carried out using this approach.

In correlation analysis, we exclude results from StoryGen~\cite{Liu_2024_CVPR_Storygen}, as its limited generation quality caused human evaluators to penalize character consistency scores, This is due to the generated character deviating significantly from the distribution of typical characters, leading to mismatched human expectations. 

To comprehensively assess the visual quality and consistency of story visualization results, we conducted a structured human evaluation on the ViStoryBench-Lite benchmark. This benchmark contains a diverse subset of stories across multiple genres and character settings, making it suitable for evaluating both identity consistency and visual storytelling fidelity.

\paragraph{Evaluation Dimensions.}
Each story visualization result was evaluated on three key dimensions:
\begin{itemize}
  \item \textbf{Character Identification Consistency:} Measures whether the main characters remain visually consistent and recognizable across different shots in the story. 
  \item \textbf{Environment Consistency:} Assesses the consistency of environmental elements—such as furniture, architecture, or background settings—across the sequence of generated images.
  \item \textbf{Subjective Aesthetics Score:} Evaluates the overall visual appeal of the story, including character quality, composition, artistic style, and storytelling clarity.
\end{itemize}

Each dimension was scored on a 5-point Likert scale (from 0 to 4), based on clear qualitative criteria provided to the annotators as shown in the scoring interface below. For example, a score of 0 in Character Consistency indicates a complete lack of identifiable characters across shots, whereas a score of 4 indicates nearly perfect character continuity.

\paragraph{Annotation Interface and Process.}
We developed a web-based annotation interface below that displays:
\begin{itemize}
  \item The full set of generated images for a story.
  \item Relevant textual annotations, such as the story prompt and shot descriptions.
  \item A structured table for scoring each of the three criteria.
\end{itemize}

The interface was designed to facilitate efficient and focused annotation, allowing annotators to toggle between image sequences and text prompts while assigning scores.

\paragraph{Annotator Pool and Assignment Strategy.}
We recruited \textbf{20 human annotators} with prior experience in visual content evaluation, including graduate students and crowd workers trained on our scoring rubric. Given the large number of models, stories, and dimensions to evaluate, we adopted a \textbf{balanced partial assignment} strategy: each annotator was assigned only a subset of the full evaluation set, but we ensured that:
\begin{itemize}
  \item \textbf{Every model-story pair} received at least \textbf{10 independent ratings per dimension}.
  \item Annotators were randomly assigned different stories and methods to avoid bias.
  \item Each task contained only a manageable number of stories (typically 6–8), to reduce fatigue.
\end{itemize}

This design helped scale the annotation process while maintaining evaluation reliability.

\paragraph{Aggregation and Analysis.}
For each model and story, we aggregated the scores by taking the mean across annotators. We also report standard deviation across annotators to reflect inter-rater variability. The collected annotations form the human evaluation benchmark for comparing model performance on ViStoryBench-Lite in terms of character coherence, environmental stability, and visual storytelling quality.

\section{Details of Prompt Alignment Evaluation}\label{sec:prompt_adherance}

To evaluate how well the generated images align with the input shot prompts, we employ GPT-4.1~\cite{GPT4.1} as an automatic evaluator. The LLM is prompted to assign a Likert-scale rating (0–4) for each image-prompt pair across several semantic dimensions, including scene correctness, camera composition, and character actions. The average of these subtask scores yields the final Alignment Score used in Table~\ref{tab:comparison_lite}.

To better analyze the capabilities and limitations of each method, we further break down the Alignment Score into four interpretable sub-scores: \textbf{Scene Score}, \textbf{Shot Score}, \textbf{Character Interaction}, and \textbf{Individual Action}, with the final score computed as an equally weighted average of these components. Each dimension focus on a specific aspect of visual-textual alignment. For example, the Scene Score evaluates the match between background or scene attributes and the prompt, while the Camera Score assesses adherence to cinematographic framing (e.g., close-up, long shot). The action scores reflect whether the actions of characters (either collectively or individually) are faithful to the described narrative.

From the results, we observe that recent video generation methods adapted for story visualization, such as \textbf{AnimDirector} and \textbf{MovieAgent (SD3)}, achieve the highest alignment scores across most categories. AnimDirector leads in \textbf{Scene Score} (3.61) and \textbf{Character Interaction} (3.24), while MovieAgent (SD3) excels in \textbf{Individual Action} (2.50). In contrast, conventional image-generation methods like \textbf{StoryGen} or \textbf{SEED-Story} generally perform poorly, with significantly lower scores across all metrics.

This detailed analysis reveals that high-quality story visualization requires coherent handling of both low-level visual elements (like camera and scene) and high-level semantics (like character intent and interaction), underscoring the necessity of specialized multi-modal reasoning for prompt alignment.

\subsection{Character Interaction}

To assess fine-grained alignment between visual content and textual prompts—particularly focusing on interactions between characters—we introduce a semantic consistency evaluation protocol targeting \textit{Character Interaction}. This task evaluates whether the generated image accurately captures the described relational dynamics between two or more characters, such as hugging, fighting, handing over an object, or sitting together. Such interactions are crucial for evaluating story-level coherence and the model’s ability to capture nuanced inter-character behavior.

\subsection{Shooting Method~(Shot)}

To assess the framing and compositional accuracy of generated images from a cinematic perspective, we propose a \textit{shot-type alignment} evaluation. This task examines how well the image conforms to the specified camera distance (e.g., close-up, medium shot, wide shot) and camera angle (e.g., eye-level, high-angle, low-angle) provided in the textual prompt. Accurate shot framing is essential for conveying narrative focus, emotional tone, and spatial arrangement—hallmarks of professional visual storytelling.

\subsection{Static Shot Description~(Scene)}

To evaluate the fidelity of background and environment rendering, we introduce a \textit{static shot grounding} task. Unlike the character-centric evaluations, this task focus on non-character elements—including environmental context, background objects, spatial layout, and overall ambient mood. It measures whether these visual elements align semantically with the scene descriptions in the prompt, such as "a classroom with a blackboard and wooden desks" or "a cozy bedroom with warm lighting and starry wallpaper." This evaluation is critical for assessing the model’s holistic scene understanding.

\subsection{Individual Action}

To further probe character-level grounding and behavioral fidelity, we propose an \textit{individual action consistency} evaluation. This task isolates the action of a specific named character described in the prompt—such as "Tom is raising his right hand"—and checks whether the generated image faithfully represents this behavior. We crop the character from the image using a detected bounding box and perform feature-based comparisons for evaluation. This task offers a focused lens on the model’s ability to correctly associate and render discrete physical actions with the correct character identity.

\subsection{Effectiveness of Prompt Alignment Metric}
\begin{figure*}[h!]
    \centering
    \includegraphics[width=\linewidth]{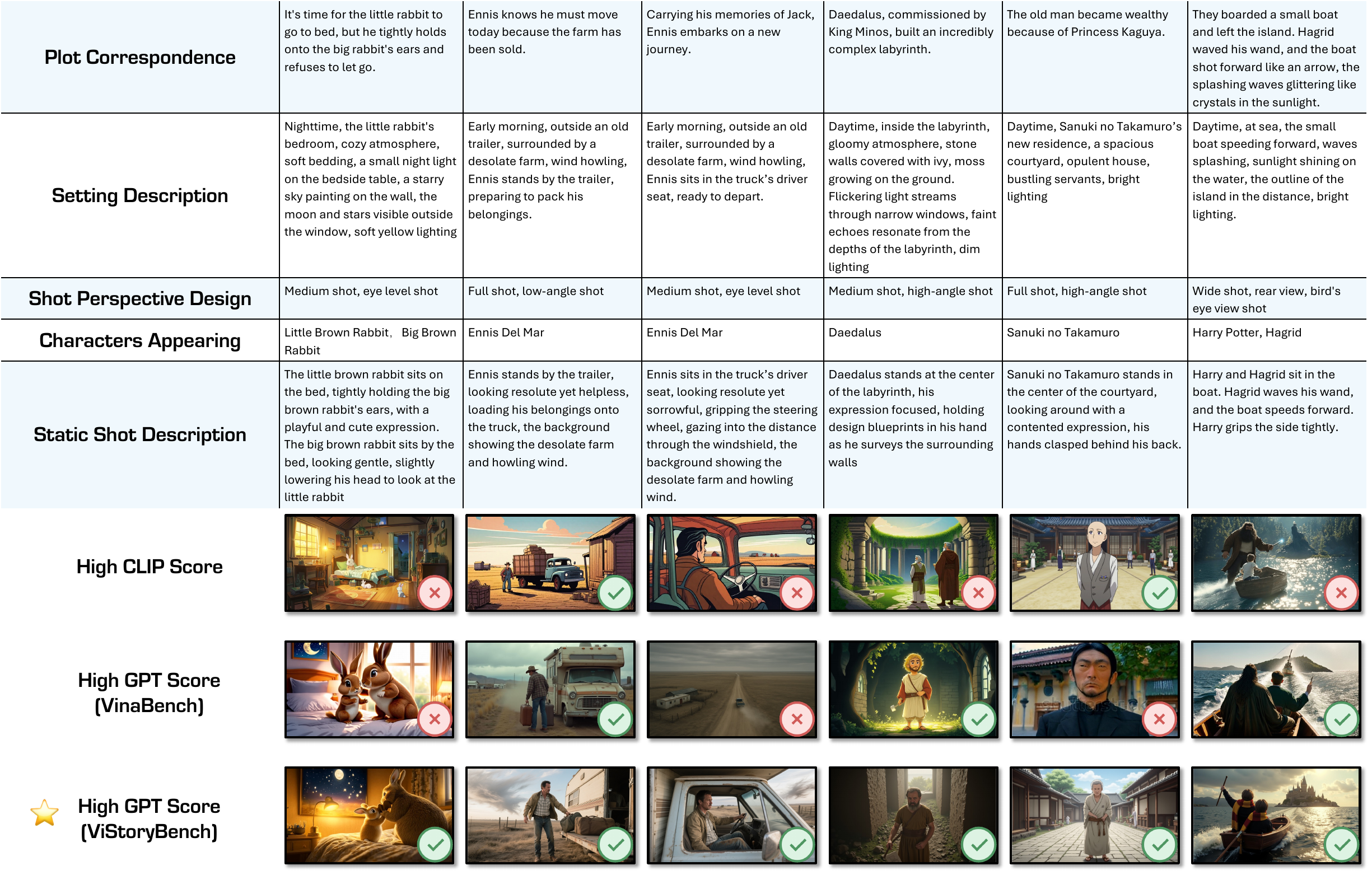}
    \caption{\small\textbf{Case Study of Prompt Alignment.} We compare the alignment between generated visualizations and the detailed prompt components—including plot correspondence, setting, shot perspective, characters, and static shot description—across six different shots. Each row provides expert annotations on prompt elements, while the bottom three rows indicate whether the generated image achieves a high CLIP score, a high GPT score according to VinaBench, or a high GPT score under the proposed ViStoryBench evaluation. ViStoryBench consistently shows better alignment with visual-semantic fidelity and shot design, as reflected by the green checks.}\label{fig:prompt_alignment}
\end{figure*}
To ensure the evaluation protocol accurately captures the nuanced alignment between generated visualizations and textual prompts, we conducted an in-depth case study analyzing prompt alignment across a diverse set of generated shots. As illustrated in Figure~\ref{fig:prompt_alignment}, we dissected the prompt into multiple semantically rich components—such as plot correspondence, setting, camera perspective, character presence, and static shot descriptions—and compared how different evaluation methods respond to these elements.

For each shot, we collected expert annotations to indicate the relevance and completeness of the generated content with respect to these prompt components. We then compared three types of automatic scoring signals: CLIP-based similarity, GPT-based scoring from VinaBench~\cite{gao2025vinabench}, and the proposed ViStoryBench evaluation protocol. The case study revealed that while CLIP and VinaBench scores often misalign with semantic and stylistic fidelity, our ViStoryBench protocol consistently correlates better with expert judgments, particularly in terms of narrative consistency and cinematographic correctness (as highlighted by the green checkmarks).

This iterative analysis informed the design of ViStoryBench’s evaluation prompts and scoring rubric, enabling a more reliable and interpretable assessment of visual story generation quality.

\subsection{Stability of VLM-based Automatic Evaluation}\label{sec:appendix_stability}

A common concern with evaluations that leverage Large Language Models (LLMs), such as GPT-4.1~\cite{GPT4.1}, is the potential for variability in their outputs, which could affect the reliability of the results. To rigorously validate the stability and robustness of our automated evaluation framework for prompt alignment, we conducted a detailed stability analysis.

\paragraph{Experimental Setup for Stability Test}
We performed 3 to 5 independent evaluation runs for each prompt-adherence metric (\textit{Scene Score}, \textit{Character Interaction}, \textit{Individual Action}, and \textit{Camera Score}). This analysis was conducted on a representative subset of our benchmark, comprising 5 diverse stories (01, 09, 27, 41, 53) and the results from three key methods: UNO, StoryDiffusion, and Story-Adapter. By repeatedly evaluating the same set of generated images, we can precisely quantify the variance attributable to the LLM evaluator itself.

\paragraph{Results and Analysis}
The results demonstrate exceptionally low variance across all evaluation runs. The standard deviation for each metric was consistently an order of magnitude smaller than the performance gaps observed between different models in our main experiments. For instance, after aggregating the scores across the selected methods and stories, we observed the following mean scores and standard deviations:
\begin{itemize}
    \item \textit{Scene Score}: $2.82 \pm 0.03$
    \item \textit{Character Interaction}: $2.57 \pm 0.04$
    \item \textit{Individual Action}: $2.40 \pm 0.08$
    \item \textit{Camera Score}: $3.23 \pm 0.03$
\end{itemize}
These minimal standard deviations confirm that the random fluctuations in the GPT-4.1~\cite{GPT4.1} evaluator are negligible. This high level of consistency ensures that the performance differences we report in our benchmark are meaningful reflections of model capabilities, rather than artifacts of evaluation noise. This finding strongly supports the reliability of using our VLM-based protocol for large-scale, automated story visualization assessment.

\subsection{Correlation Analysis of Qwen-base and GPT-based Evaluation}\label{sec:qwen_gpt_correlation}

Based on the Prompt Alignment scoring results from the ViStoryBench-Lite dataset, we conducted a correlation analysis between the scores provided by \textbf{GPT-4.1}~\cite{GPT4.1} and \textbf{Qwen3-VL-8B-Instruct}~\cite{yang2025qwen3}. This analysis aimed to reveal the consistency between these two VLMs in evaluating story visualization methods. We focused on the Prompt Alignment Score (PA) average (Avg.) as it represents the comprehensive indicator of overall performance. The PA score includes sub-items: Scene (scene description), Shot (shot perspective), CI (character interaction), IA (individual action), and Avg. (average). Below is a detailed analysis report.

We extracted the PA Avg. scores (GPT-based and Qwen-based) for each method from Table~\ref{tab:comparison_lite} and Table~\ref{tab:comparison_full_lite_pa_qwen}, forming a paired dataset with scores ranging from 0 to 4 points. Statistical correlation analysis was performed, calculating the \textbf{Pearson correlation coefficient} (measuring linear correlation) and the \textbf{Spearman rank correlation coefficient} (measuring monotonic correlation based on rankings). The calculations and analyses are as follows:

\paragraph{Pearson Correlation Analysis for PA Average.}
The Pearson correlation coefficient is calculated using the formula:
\[
r = \frac{\sum (X_i - \bar{X})(Y_i - \bar{Y})}{\sqrt{\sum (X_i - \bar{X})^2 \sum (Y_i - \bar{Y})^2}},
\]
where X represents GPT-based scores and Y represents Qwen-based scores.

Based on 34 data points, the Pearson correlation coefficient is $r \approx 0.93$. This value close to 1 indicates a strong positive linear correlation between GPT-based and Qwen-based PA Avg. ratings. That is, when GPT scores are higher, Qwen scores tend to be higher as well, and vice versa.

\paragraph{Spearman Correlation Analysis for PA Average.}
The Spearman correlation coefficient is calculated using the formula:
\[
\rho = 1 - \frac{6 \sum d_i^2}{n(n^2 - 1)},
\]
where $d_i$ is the difference in ranks, and n is the sample size.

After ranking the scores (higher scores receive higher ranks) and computing the rank differences, the Spearman correlation coefficient is $\rho \approx 0.89$. This value also close to 1 indicates high consistency in the ranking order between the two ratings, demonstrating significant monotonic correlation. This implies that both models similarly evaluate the relative performance of different methods.

\paragraph{Correlation Analysis for PA Sub-items.}
For comprehensiveness, we also calculated the correlation coefficients for the PA sub-items (Scene, Shot, CI, IA). The average Pearson correlation coefficient ranges from approximately $0.85$ to $0.90$, and the average Spearman coefficient ranges from $0.82$ to $0.88$, both indicating moderate to strong correlations. 

\paragraph{Conclusion of Correlation Analysis.}
Through correlation analysis of the Prompt Alignment scores on the ViStoryBench-Lite dataset, we found that the ratings from GPT-4.1 and Qwen3-VL are highly correlated (Pearson $r \approx 0.93$, Spearman $\rho \approx 0.89$). Both GPT-4.1 and Qwen3-VL are advanced VLMs with powerful multimodal understanding capabilities. They are trained on similar datasets, resulting in similar evaluation criteria for prompt alignment. The strong correlation indicates that in story visualization tasks, both models perform consistently when evaluating prompt alignment capability, demonstrating interchangeability between them.

\section{Details of Character Identification Similarity}\label{sec:cids}

\subsection{Calculation}

Figure~\ref{fig:CIDS} illustrates the computation pipeline of our Character Identification Similarity (CIDS) metric. CIDS quantifies both \textit{self-similarity} (within generated images) and \textit{cross-similarity} (between generated and reference images) through a four-stage computational pipeline:  

\begin{enumerate}[label=\textcircled{\scriptsize\arabic*}, leftmargin=1.4em, parsep=-0.05em]
    \item \textbf{Character Detection.}  
    Grounding DINO localizes character regions using text prompts.  
    
    ◦ \textit{Reference images}: Crops character regions with edge trimming (minor boundary adjustments).  

    ◦ \textit{Generated images}: May fail to detect characters (returns empty result), indicating identity inconsistency.  

    \item \textbf{Feature Extraction.}  
    Extracts 512D embeddings from cropped regions:  
    
    ◦ \textit{Realistic characters}: ArcFace/AdaFace/FaceNet tri-model ensemble (robust facial features).  

    ◦ \textit{Stylized characters}: CLIP ViT-L/14 (semantic alignment).  

    \item \textbf{Bipartite Matching.}  
    Solves optimal character correspondence via Hungarian algorithm:  
    
    ◦ Computes cosine similarity matrix between reference/generated features.  

    ◦ Matches characters maximizing global similarity (excludes failed detections).  

    \item \textbf{Scoring.}  
    Final metric: $\text{CIDS} = \frac{1}{N}\sum_{i=1}^{N} \cos(\mathbf{v}_{\text{ref}}^{(i)}, \mathbf{v}_{\text{gen}}^{(i)})$
    where N = number of matched pairs, $\mathbf{v}$ = feature vectors.  
\end{enumerate}  

\subsection{Impact of Reference Image Selection on Cross-CIDS Metric}
\begin{table}[h!]
  \centering
  \caption{\small \textbf{Cross-CIDS Metric with Different Reference Image}. "\textbf{Dataset Reference}" refers to results calculated with reference images in ViStoryBench dataset, "\textbf{Generated Reference}" refers to results calculated with generated reference images of methods. All results below are obtained on ViStoryBench-Lite.}
  \vspace{0.2cm}
  \resizebox{\linewidth}{!}{
  \begin{tabular}{l|cc}
    \toprule
    \textbf{Method} & \textbf{Dataset Reference} & \textbf{Generated Reference} \\
    \midrule
    MOKI~\cite{moki2024} & 0.292 & 0.338 \\
    \rowcolor{teaserdatasetback}AIbrm~\cite{brmgo2025} & 0.559 & 0.683 \\
    ShenBi~\cite{shenbi2025} & 0.347 & 0.389 \\
    \bottomrule
  \end{tabular}
  }
  \label{tab:comparison_CIDS}
\end{table}

In certain methods, the character reference images or features used for generation are not directly sourced from our dataset but are instead synthesized through an additional generation stage. A common scenario involves converting real-person reference images from our dataset into stylized versions—such as anime-style characters—resulting in visual appearances that may differ substantially from the original subjects. These discrepancies in reference image selection can significantly impact the Cross-CIDS metric. In the main results tables, we report scores based on the original reference images provided in the dataset. For a more comprehensive comparison, we additionally report Cross-CIDS scores computed using the synthesized reference characters, as shown in Table~\ref{tab:comparison_CIDS}.

\section{Details of Style Similarity Calculation}\label{sec:csd}
\begin{figure*}[h]
    \centering
    \includegraphics[width=0.8\linewidth]{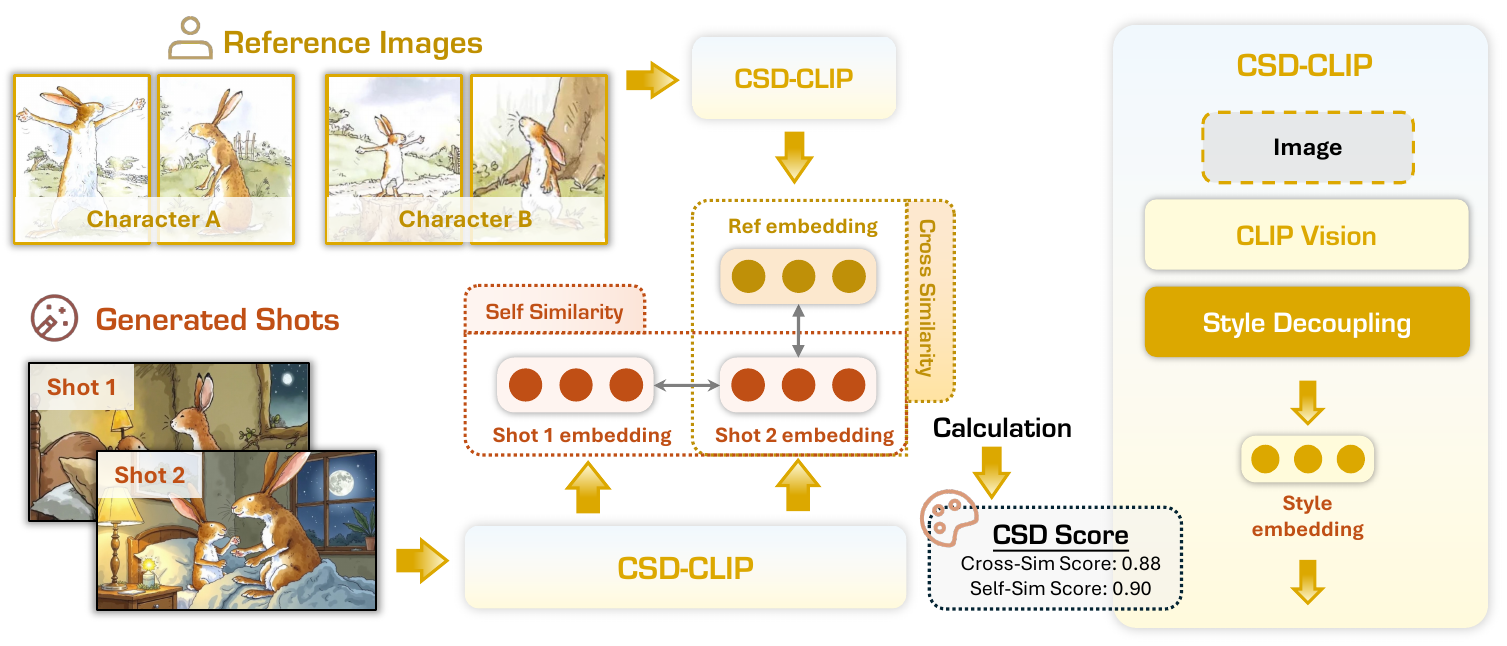}
    \caption{\small \textbf{Style Similarity Calculation Pipeline}. Evaluating both cross-similarity and self-consistency by computing cosine similarity between the style features of two images.}
    \label{fig:stylesim}
\end{figure*}
Figure~\ref{fig:stylesim} illustrates the computation pipeline of our Style Similarity metric. Adapted from CSD~\cite{somepalli2024measuring,zhuang2025styleme3d}, this metric captures both \textit{self-similarity} (within generated images) and \textit{cross-similarity} (between generated and reference images) by analyzing style-specific features extracted via CSD-CLIP~\cite{somepalli2024measuring}.

The computation consists of three key steps:  
\begin{enumerate}[label=\textcircled{\scriptsize\arabic*}, leftmargin=1.4em, parsep=-0.05em] 
    \item Each image is encoded into visual embeddings using a CLIP~\cite{radford2021learning} vision encoder pre-trained on large-scale style datasets;  
    \item The extracted features are passed through CSD layers to disentangle content and style representations, retaining only the style components;  
    \item Pairwise cosine similarity is computed between the resulting style embeddings to measure stylistic alignment.
\end{enumerate}

This design enables fine-grained comparison of artistic and stylistic consistency, independent of content semantics.

\section{Details of OCCM}\label{sec:occm}
Onstage Character Count Matching (OCCM) metric relies on an upstream character detector to obtain the detected character count ($D$). Consequently, the accuracy of the OCCM score is inherently bounded by the detector's performance, which may be affected by factors like heavy occlusion or extreme artistic styles. We opted for expert models as the upstream detector. While expert models can misjudge, VLMs perform worse in hallucination counting.
The design of OCCM formula is based on two core principles:
    \begin{enumerate}
        \item \textbf{Scale Normalization:} By dividing the absolute error $|D - E|$ by the expected count $E$, we convert the error into a relative percentage. This allows the metric to fairly evaluate scenes of varying scales, ensuring that the same relative error receives a similar penalty, whether in a single-character scene ($E=1$) or a multi-character scene ($E=10$).
        \item \textbf{Non-linear Penalty:} We employ an exponential function to penalize the relative error. Unlike a linear penalty, the exponential function leads to a gentle score decay for small errors but a sharp drop for larger ones. This characteristic better aligns with human perception: missing one out of ten characters is a minor flaw, whereas missing five constitutes a critical failure.
    \end{enumerate}

\section{Details of Copy-Paste Detection}\label{sec:copy_paste}

To rigorously evaluate whether the generated image is merely a replication of a specific reference image (denoted as the anchor or target reference $\mathbf{r}_0$) rather than a generalized synthesis from the provided character concept, we employ a Softmax-based Copy-Paste Score. 

Let $\mathbf{g}$ be the unit-normalized feature vector of the generated image, and $\mathcal{R} = \{\mathbf{r}_0, \mathbf{r}_1, \dots, \mathbf{r}_N\}$ be the set of unit-normalized feature vectors for the input reference images, where $\mathbf{r}_0$ represents the primary reference subject to copy-paste detection, and $\{\mathbf{r}_1, \dots, \mathbf{r}_N\}$ serves as the set of auxiliary references for the same character.

We first calculate the cosine similarity between the generated image and each reference image in the set $\mathcal{R}$. To quantify the exclusivity of the match between $\mathbf{g}$ and the target $\mathbf{r}_0$ relative to other references, we formulate the score as a probability distribution using a temperature-scaled Softmax function:

\begin{equation}
    \label{eq:copy_rate}
    \text{CopyRate}(\mathbf{g} | \mathcal{R}) = \frac{\exp(\mathbf{g}^\top \mathbf{r}_0 / \tau)}{\sum_{k=0}^{N} \exp(\mathbf{g}^\top \mathbf{r}_k / \tau)}
\end{equation}

where $\tau$ is a temperature hyperparameter set to $0.01$. This low temperature value sharpens the distribution, making the metric highly sensitive to the nearest neighbor in the feature space. 

The resulting \textbf{Copy-Paste Rate} ranges from $0$ to $1$. A score approaching $1$ indicates that the generated image $\mathbf{g}$ is significantly more similar to the specific reference $\mathbf{r}_0$ than to any other provided references, suggesting a "copy-paste" overfitting behavior. Conversely, a lower score implies that the generated features are either distributed among multiple references or have successfully generalized beyond the specific appearance of $\mathbf{r}_0$. The final metric is averaged across all generated samples for each character.

\section{Benchmark Evaluation Efficiency}\label{sec:efficiency}
\begin{table}[h!]
  \centering
  \caption{\small \textbf{Computational Efficiency of Evaluation Metrics}.}
  \label{tab:metric_timing}
  \vspace{0.2cm}
  \resizebox{\linewidth}{!}{
  \begin{tabular}{l|ll|l}
    \toprule
    \textbf{Metric} & \textbf{Scope} & \textbf{Time} & \textbf{Notes} \\
    \midrule
    Aesthetics Score & Single image & 0.026s & per generated image\\
    \rowcolor{teaserdatasetback}Style Similarity & Pair images & 0.046s & cross or self\\
    Character Similarity & Pair images & 0.450s & cross or self\\
    \rowcolor{teaserdatasetback}Inception Score & Total data & 8.057s & full dataset\\
    Prompt Alignment\textsuperscript{*} & Single image & 25.173s & per generated image\\
    \bottomrule
  \end{tabular}
  }
  
  \vspace{0.5em}
  \footnotesize
  * Longest computation due to LLM inference constraints
\end{table}

To ensure the reproducibility and transparency of our computational experiments, we provide detailed information regarding the hardware setup and evaluation runtime. All experiments are conducted using high-performance GPU accelerators (e.g., NVIDIA H800 with 80GB memory), and we ensure consistent measurement conditions across metrics.

Table~\ref{tab:metric_timing} reports the average computational cost associated with each evaluation metric. These metrics differ significantly in scope and computational complexity. For example, aesthetic scoring is performed on individual images and is highly efficient, averaging only 0.026 seconds per image. Style and character similarity metrics, which operate on image pairs, are slightly more demanding, especially character similarity (0.450 seconds per pair), likely due to the use of deep feature extractors.

In contrast, metrics such as the Inception Score are computed over the entire dataset and thus have a higher runtime (8.057 seconds), though only once per dataset. The most computationally intensive metric is the Prompt Alignment score, requiring over 25 seconds per image. This is primarily due to the involvement of large language model (LLM) inference, which introduces latency constraints.

\onecolumn

\begin{tcolorbox}[title=Prompt for Character Interactions,
enhanced, 
skin first=enhanced,
skin middle=enhanced,
skin last=enhanced,
colback=yellow!10!white,
colframe=orange!80!black,
fonttitle=\bfseries]{}
\textbf{Task Definition}\\
You will be provided with an image and a text prompt describing the main character's action. As an experienced evaluator, your task is to evaluate the semantic consistency between the image and the text prompt, according to the scoring criteria. This evaluation focus specifically on whether the action of the main character in the image aligns with the action described in the text.\\

\textbf{Scoring Criteria}\\
When evaluating the semantic consistency between an image and its corresponding text prompt, the following aspects are crucial:
\begin{itemize}[itemsep=-1.4pt,leftmargin=2em]
    \item \textbf{Relevance}: Does the image show the main character performing the action or behavior mentioned in the text? The action in the image should match the core description provided in the text.
    \item \textbf{Accuracy}: Does the image depict the action correctly according to the text prompt? Any specific details related to the action, such as gestures, posture, or environment, should align with the description.
    \item \textbf{Completeness}: Does the image show the main character completing the entire action as described in the text? The image should not omit important parts of the action or behavior.
\end{itemize}

\textbf{Scoring Range}\\
Based on these criteria, you will assign a score from 0 to 4 that reflects the degree of semantic consistency between the image and the text prompt:
\begin{itemize}[itemsep=-1.4pt,leftmargin=2em]
    \item \textbf{Very Poor (0)}: No correlation. The image does not reflect any aspect of the action described in the text prompt.
    \item \textbf{Poor (1)}: Weak correlation. The image addresses the text in a very general way but misses most details and accuracy of the action.
    \item \textbf{Fair (2)}: Moderate correlation. The image depicts the action to some extent, but there are several inaccuracies or missing details.
    \item \textbf{Good (3)}: Strong correlation. The image accurately portrays most elements of the action with minor inaccuracies or omissions.
    \item \textbf{Excellent (4)}: Near-perfect correlation. The image closely aligns with the text prompt and portrays the main character’s action with high accuracy and precision.
\end{itemize}

\textbf{Input format}\\

Every time you will receive a text prompt and an image.

Please carefully review the image and text prompt. Before giving a score, please provide a brief analysis of the above evaluation criteria, which should be very concise and accurate.\\

\textbf{Output Format}\\
Analysis: \textless \textcolor{blue}{Your analysis}\textgreater

Score: \textless \textcolor{blue}{Your Score}\textgreater

\end{tcolorbox}

\centering
\begin{tcolorbox}[title=Prompt for Shooting Evaluation,
enhanced, 
skin first=enhanced,
skin middle=enhanced,
skin last=enhanced,
colback=yellow!10!white,
colframe=orange!80!black,
fonttitle=\bfseries]{}
    \textbf{Task Definition}

You will be provided with an image and a text prompt describing the shot type of the image. As an experienced evaluator, your task is to assess whether the generated image meets the specified shot requirements based on the evaluation criteria.\\

\textbf{Additional Material}

Instruction: You are a professor evaluator. Below is information about different shot types and shot distances. Please evaluate whether the generated image meets the requested shot type.\\

\textbf{Shot Distance Descriptions}
\begin{itemize}[itemsep=-1.4pt,leftmargin=2em]
\item \textbf{Long Shot}: Shows the relationship between characters and their environment, typically used to display the scene or environment.
\item \textbf{Full Shot}: Shows the full body of a character, commonly used to display movement or the full scene.
\item \textbf{Medium Long Shot}: Starts from above the character’s knees, capturing part of the environment.
\item \textbf{Medium Shot}: Captures the character from the waist up.
\item \textbf{Close-Up}: Captures the character from the chest up.
\item \textbf{Extreme Close-Up}: focus on the character’s head or face, with the background and environment typically blurred or not visible.
\end{itemize}

\textbf{Angle Descriptions}
\begin{itemize}[itemsep=-1.4pt,leftmargin=2em]
\item \textbf{Eye Level Shot}: The camera is positioned at the subject's eye level.
\item \textbf{Low Angle Shot}: The camera is positioned below eye level, shooting upward, emphasizing the character’s power or size.
\item \textbf{High Angle Shot}: The camera is positioned above eye level, shooting downward, often minimizing the subject's significance.
\item \textbf{Bird’s Eye View}: Camera shot taken from directly above, providing an overview of the scene.
\item \textbf{Tilted Shot}: The camera is intentionally tilted to create a sense of imbalance or tension.
\item \textbf{Perspective Compression}: A technique that emphasizes depth and the relationship between foreground and background through perspective.
\end{itemize}

\textbf{Scoring Range}

A score between 0 and 4 will be assigned based on how well the shot type aligns with the content described in the prompt:
\begin{itemize}[itemsep=-1.4pt,leftmargin=2em]
\item \textbf{Very Poor (0)}: The image does not meet any shot or angle requirements.
\item \textbf{Poor (1)}: The image meets some but not most of the shot or angle requirements.
\item \textbf{Fair (2)}: The image partially meets the shot or angle requirements, but some elements are off.
\item \textbf{Good (3)}: The image meets most of the shot or angle requirements.
\item \textbf{Excellent (4)}: The image fully meets all of the shot and angle requirements.
\end{itemize}

\textbf{Input Format}

You will receive a text prompt and an image. Please carefully review the image and text prompt. Provide an analysis followed by a score.\\

\textbf{Output Format}

Analysis: \textless \textcolor{blue}{Your analysis} \textgreater \\
Score: \textless \textcolor{blue}{Your score}\textgreater

\end{tcolorbox}

\centering

\begin{tcolorbox}[title=Prompt for Static Shot Evaluation,
enhanced, 
skin first=enhanced,
skin middle=enhanced,
skin last=enhanced,
colback=yellow!10!white,
colframe=orange!80!black,
fonttitle=\bfseries]{}
    \textbf{Task Definition}\\
You will be provided with an image and a text prompt that describes the \textbf{background}, objects, and mood of the scene (excluding characters). Your task is to evaluate the consistency between the \textbf{background and objects} described in the prompt and what is visually represented in the image.\\

\textbf{Evaluation Criteria}

When assessing the semantic consistency between the image and the text prompt, focus on how well the \textbf{background and non-character elements} in the image match the description provided in the text. The evaluation should be based on the following aspects:
\begin{itemize}[itemsep=-1.4pt,leftmargin=2em]
    
\item \textbf{Relevance}: The image should clearly relate to the primary background elements and objects described in the text. It should reflect the main setting and environment described, without introducing irrelevant or unrelated features.

\item \textbf{Accuracy}: Check if the specific details mentioned in the text are correctly represented in the image. This includes any mentioned objects, scenery, environmental conditions (e.g., weather, lighting), and relevant background elements.

\item \textbf{Completeness}: Evaluate whether the image accurately includes all critical background elements described in the text. The image should reflect the key details and setting, not leaving out essential aspects of the described background or scene.

\item \textbf{Context}: The image should maintain the context of the description. If the text describes a specific environment or atmosphere, the image must capture that context appropriately, considering the described mood and setting elements.
\end{itemize}

\textbf{Scoring Criteria}

Based on these factors, the image will be assigned a score from \textbf{0 to 4}, indicating the degree of consistency between the image and the description in the text:
\begin{itemize}[itemsep=-1.4pt,leftmargin=2em]
\item \textbf{Very Poor (0)}: No correlation. The image completely fails to reflect the background or objects described in the text.
\item \textbf{Poor (1)}: Weak correlation. The image touches on the background or objects in a very general sense but misses most of the important details or has significant inaccuracies.
\item \textbf{Fair (2)}: Moderate correlation. The image contains some relevant background and objects, but several important details are missing or inaccurately represented.
\item \textbf{Good (3)}: Strong correlation. The image accurately represents most of the described background and objects with minor omissions or inaccuracies.
\item \textbf{Excellent (4)}: Near-perfect correlation. The image perfectly captures the background and objects as described in the text, leaving no significant details missing or inaccurate.

\end{itemize}

\textbf{Input Format}

You will receive a text prompt and an image. Please carefully review the image and text prompt. Provide an analysis followed by a score.\\

\textbf{Output Format}\\
Analysis: \textless \textcolor{blue}{Your analysis}\textgreater\\
Score: \textless \textcolor{blue}{Your Score}\textgreater
\end{tcolorbox}

\begin{tcolorbox}[title=Prompt for Individual Action Evaluation,
enhanced, 
skin first=enhanced,
skin middle=enhanced,
skin last=enhanced,
colback=yellow!10!white,
colframe=orange!80!black,
fonttitle=\bfseries]{}
    \textbf{Task Definition}\\
For each evaluation, you will receive a text prompt, an image, and a character name. Your task is to \textbf{first extract the individual action or behavior of the specified character from the text prompt, then determine whether the image accurately reflects this description for that character}, and finally assign a score based on the criteria.\\

\textbf{Evaluation Process}

\begin{itemize}[itemsep=-1.4pt,leftmargin=2em]
\item \textbf{Extract Action Information}: Carefully extract the specific action or behavior described for the given character (character name) from the text prompt.
\item \textbf{Image Comparison}: Examine the image to determine whether the specified character’s action matches the extracted description.
\item \textbf{Analyze and Score}: Analyze the match according to the scoring criteria and assign a score.

\end{itemize}

\textbf{Scoring Criteria}

Focus on the following aspects when evaluating:
\begin{itemize}[itemsep=-1.4pt,leftmargin=2em]
\item \textbf{Relevance}: Does the image show the specified character performing the action or behavior described in the text?
\item \textbf{Accuracy}: Are the details of the character’s action in the image (such as posture, gestures, environment) consistent with the text description?
\item \textbf{Completeness}: Does the image fully depict the character completing the entire action as described, without omitting important parts?
\end{itemize}

Assign a score from 0 to 4 based on the degree of semantic consistency:

\begin{itemize}[itemsep=-1.4pt,leftmargin=2em]
\item \textbf{0 (Very Poor)}: No correlation. The image does not reflect any aspect of the described action.
\item \textbf{1 (Poor)}: Weak correlation. The image only generally addresses the text, missing most details and accuracy.
\item \textbf{2 (Fair)}: Moderate correlation. The image depicts the action to some extent but with several inaccuracies or missing details.
\item \textbf{3 (Good)}: Strong correlation. The image accurately portrays most elements of the action, with only minor inaccuracies or omissions.
\item \textbf{4 (Excellent)}: Near-perfect correlation. The image closely aligns with the text prompt and depicts the character’s action with high accuracy and completeness.
\end{itemize}

\textbf{Output Format}\\
Analysis: \textless \textcolor{blue}{Your analysis}\textgreater \\
Score: \textless \textcolor{blue}{Your Score}\textgreater
\end{tcolorbox}

\twocolumn
\clearpage

{
    \small
    \bibliographystyle{ieeenat_fullname}
    \bibliography{main}
}

\end{document}